\documentclass[sigconf]{acmart}

\setcopyright{none}
\renewcommand\footnotetextcopyrightpermission[1]{}

\usepackage{nicefrac}
\usepackage{float}
\usepackage{subcaption}
\usepackage{multirow}
\usepackage{adjustbox}
\usepackage[linesnumbered,ruled,vlined]{algorithm2e}
\usepackage{soul}
\usepackage{marvosym}
\usepackage{amsfonts}
\usepackage{xcolor}
\DeclareMathOperator*{\argmin}{arg\,min}
\newcommand{\norm}[1]{\left\lVert#1\right\rVert}

\AtBeginDocument{%
  \providecommand\BibTeX{{%
    \normalfont B\kern-0.5em{\scshape i\kern-0.25em b}\kern-0.8em\TeX}}}

\begin{document}

\title{Robust Few-Shot Ensemble Learning \\
with Focal Diversity-Based Pruning}

\author{Selim Furkan Tekin}
\email{stekin6@gatech.edu}
\affiliation{%
  \institution{Georgia Institute of Technology}
  \city{Atlanta}
  \state{Georgia}
  \country{USA}
}

\author{Fatih Ilhan}
\email{filhan@gatech.edu}
\affiliation{%
  \institution{Georgia Institute of Technology}
  \city{Atlanta}
  \state{Georgia}
  \country{USA}
}

\author{Tiansheng Huang}
\email{thuang@gatech.edu}
\affiliation{%
  \institution{Georgia Institute of Technology}
  \city{Atlanta}
  \state{Georgia}
  \country{USA}
}

\author{Sihao Hu}
\email{sihaohu@gatech.edu}
\affiliation{%
  \institution{Georgia Institute of Technology}
  \city{Atlanta}
  \state{Georgia}
  \country{USA}
}

\author{Ka-Ho Chow}
\email{khchow@gatech.edu}
\affiliation{%
  \institution{Georgia Institute of Technology}
  \city{Atlanta}
  \state{Georgia}
  \country{USA}
}

\author{Margaret L. Loper}
\email{margaret.loper@gtri.gatech.edu}
\affiliation{%
  \institution{Georgia Institute of Technology}
  \city{Atlanta}
  \state{Georgia}
  \country{USA}
}

\author{Ling Liu}
\email{ling.liu@cc.gatech.edu}
\affiliation{%
  \institution{Georgia Institute of Technology}
  \city{Atlanta}
  \state{Georgia}
  \country{USA}
}

\renewcommand{\shortauthors}{Trovato and Tobin, et al.}

\begin{abstract}
{
This paper presents {\sc FusionShot}, a focal diversity optimized few-shot ensemble learning approach for boosting the robustness and generalization performance of pre-trained few-shot models. The paper makes three original contributions. First, we explore the unique characteristics of few-shot learning to ensemble multiple few-shot (FS) models by creating three alternative fusion channels. Second, we introduce the concept of focal error diversity to learn the most efficient ensemble teaming strategy, rather than assuming that an ensemble of a larger number of base models will outperform those sub-ensembles of smaller size. We develop a focal-diversity ensemble pruning method to effectively prune out the candidate ensembles with low ensemble error diversity and recommend top-$K$ FS ensembles with the highest focal error diversity. Finally, we capture the complex non-linear patterns of ensemble few-shot predictions by designing the learn-to-combine algorithm, which can learn the diverse weight assignments for robust ensemble fusion over different member models.  
Extensive experiments on representative few-shot benchmarks show that the top-K ensembles recommended by {\sc FusionShot} can outperform the representative SOTA few-shot models on novel tasks (different distributions and unknown at training), and can prevail over existing few-shot learners in both cross-domain settings and adversarial settings. For reproducibility purposes, {\sc FusionShot} trained models, results, and code are made available at \url{https://github.com/sftekin/fusionshot}.
}
\end{abstract}





\maketitle
\pagestyle{empty}

\section{Introduction}

%

A few-shot image classifier combines deep embedding learning with metric space distance learning to extract and predict latent features of entities of interest.
It maps examples of similar features in the real world (input data space) to the latent neural embedding representations (latent feature space) with dual properties: (i) the latent feature vectors learned from the examples of similar features in the real world will be closer, and the examples with high dissimilarity in the real world will remain distant in the latent feature embedding space; and (ii) a well-trained few-shot model can extract the latent features from a novel data with different distributions, which are closer in the latent space to those entities that are more relevant while distant to those that have no or less relevance in the real world. Given that both deep embedding learning and metric space latent distance learning are critical to the learning efficiency of few-shot models, we make three arguments. First,  different DNN backbone algorithms for deep embedding learning may complement one another in more efficient learning of high-quality of deep embedding for latent feature extractions. Second, different metric space latent distance functions may contribute complementary wisdom in harmonizing the inconsistency among their distance computation. Finally, a hybrid fusion of both diverse deep embeddings and diverse latent distance computations may further stabilize the few-shot learning results. 


To this end, we introduce a focal diversity optimized few-shot ensemble learning approach, coined as {\sc FusionShot}, which intelligently integrates the complimentary wisdom of multiple independently trained few-shot models.
Given a pool of $N$ pre-trained base few-shot models, {\sc FusionShot} can select the best-performing ensembles that can outperform the ensemble of all $N$ models on novel tasks (with different distribution and unknown at training). Also, the top-$K$ ensembles chosen by {\sc FusionShot} can outperform the best-performing individual FS model in the ensemble team, even when a majority of the member models fail. {\sc FusionShot} by design is not only effective to novelty data in cross-domain settings or concept drift scenarios but also robust in adversarial settings.



{\bf Contributions and Scope.\/} 
This paper makes three original contributions. First, we ensemble multiple few-shot (FS) models by exploring all three alternative fusion channels:
(i) the fusion of various latent distance methods (FusionShot$^{dist}$), (ii) the fusion of multiple DNN embedding backbone algorithms (FusionShot$^{bb}$), and (iii) the hybrid fusion by combining different latent distance functions with different deep embedding backbones (FusionShot$^{hybrid}$). 
Second, to effectively capture the complex non-linear disagreement relations among multiple independently trained FS models, we introduce the concept of focal error diversity to learn the most efficient ensemble teaming strategy. 
Based on the error diversity scores, we show that there exist few-shot ensemble teams of small size (w.r.t. \# of member models), which can outperform the FS ensembles of large size because some FS model makes the same mistakes as other ensemble members and fails to contribute to the ensemble prediction due to lacking complimentary wisdom. We develop a lightweight focal-diversity ensemble pruning method to effectively prune out those candidate ensemble teams with low-error diversity and recommend top-$K$ FS ensembles with the highest focal error diversity. Finally, unlike conventional ensemble consensus methods such as simple means, majority or plural voting, we capture the complex non-linear patterns of ensemble few-shot predictions by designing the learn-to-combine algorithm, which can learn the diverse weight assignments to different member models of an FS ensemble for robust fusion.  
%

We evaluate the robustness of {\sc FusionShot} from three perspectives. In terms of performance, we compare it with both existing individual SOTA few-shot models and the representative few-shot ensemble methods on popular benchmarks.
Our results show that given a pool of $N$ base few-shot models, our FusionShot can select the best few-shot sub-ensembles, which offer better generalization performance on novel tasks even when the strongest models fail. In terms of adversarial resilience, we test under adversarial scenarios and show that {\sc FusionShot} can create an ensemble defense team to protect a victim FS model against the Projected Gradient Attack (PGD)~\cite{madry2017towards}.
In terms of generalizability, we show that FusionShot is more stable and adaptive under concept shifts and cross-domain settings, and can quickly adjust itself to the switched domain.

\vspace{-5pt}
\section{Related Work}
\vspace{-2pt}
The most relevant few-shot algorithms include the recent SOTA individual few-shot algorithms, e.g., SimpleShot~\cite{wang2019simpleshot}, DeepEMD~\cite{zhang2020deepemd}, feature adaptations~\cite{bateni2020improved, ye2020few}, and 
the recent few-shot ensemble methods, which are based on joint training of multiple feature extractors and normalization techniques, e.g., Robust-20~\cite{dvornik2019diversity} and Bendou~\cite{bendou2022easy}. These proposals improve the top-1 or top-5 performance over the selection of models including some traditional few-shot learners, e.g., ProtoNet~\cite{snell2017prototypical}, MatchingNet~\cite{vinyals2016matching}, and RelationNet~\cite{sung2018learning}. The key distinguishing property of these recent improvements can be characterized by two observations. First, recent studies suggest using a more complex latent distance function, such as Earth Mover's Distance~\cite{zhang2020deepemd}, Mahalanobis ~\cite{bateni2020improved} for learning latent feature similarity. For example, DeepEMD \cite{zhang2020deepemd} shows that under the same backbone DNN feature extractor, Earth Mover's Distance (EMD) metric performs better than other non-parametric distance metrics, e.g., cosine, Euclidean, and some popular parametric distance metrics, e.g., the RelationNet~\cite{sung2018learning}. Second, SimpleNet~\cite{wang2019simpleshot} shows that using k-nearest-neighbor and k-means clustering with cosine distance can be more effective for learning metric-space distance than traditional few-shot algorithms, such as MatchingNet, ProtoNet, RelationNet. 

Several efforts on enhancing few shot learning, especially few shot classification model, through optimizations on relevant feature selections or relevant latent distance function selection~\cite{chen2019closer,dvornik2020selecting} or optimiztions on universal templates for multiple datasets~\cite{ triantafillou2019meta,  triantafillou2021learning}. Our approach to few-shot ensemble learning differs from existing proposals in two perspectives. First, many existing methods rely on end-to-end re-training. In contrast, FusionShot leverages pre-trained FS models to construct a robust few-shot ensemble learner by selecting the ensemble fusion channel, pruning the largest ensemble to recommend the top-$k$ sub-ensemble teams of high focal ensemble diversity, and then boosting the ensemble performance by exploiting the learn-to-combine weight assignment function. Second, some recent efforts show that the few-shot models are highly susceptible to adversarial attacks, such as PGD \cite{madry2017towards}. Compared to \cite{goldblum2020adversarially, li2020adversarial}, the FusionShot approach creates a proactive defense mechanism with strong adversarial robustness.


\vspace{-4pt}
\section{Few-Shot Problem Definition}
\vspace{-2pt}
Let $\mathcal{D}=\{(\mathbf{I}_{i}, y_i)\}_{i=1}^{L}$ denote a dataset with each sample $\mathbf{I}_i$ paired with a corresponding class label $y_i$, where $y_i\in C$ and $C$ is the categories present in the dataset. The dataset is partitioned into $\mathcal{D}^{\mathrm{train}}$, $\mathcal{D}^{\mathrm{val}}$, and $\mathcal{D}^{\mathrm{novel}}$, denoting training set, validation set, and novel test set respectively. In contrast to classical supervised learning, the classes in each set are also disjoint, where the goal is to create a class-independent model by learning how to compare images and decide whether they are in the same class or not. The standard learning approach is episodic \cite{vinyals2016matching}, where randomly sampled images from the corresponding dataset form an episode.

For $K$-\textit{way} $J$-\textit{shot} learning, an episode consists of a query $\mathbf{Q}$ and a support set $\mathcal{S}$ of $K$ classes with $J$ samples per class. Namely, $\mathcal{S}=\{\{\mathbf{I}_{1}^{c_{1}}, \dots, \mathbf{I}_{J}^{c_{1}}\}, \dots, \{\mathbf{I}_{1}^{c_{K}}, \dots, \mathbf{I}_{J}^{c_{K}}\}\}$, and $|\mathcal{S}|=K\times J$, and let $\mathbf{I}_{j}^{c_{i}}$ denote an input sample $j$ belonging to class $c_i$. The query, $\mathbf{Q}$, should refer to a sample that does not exist in the support set $\mathcal{S}$ but belongs to one of the $K$ classes used in $\mathcal{S}$, i.e., $\mathbf{Q}\notin\mathcal{S}, \mathbf{Q}=\mathbf{I}^{c_i}$ $c_{i} \in \{c_1, \dots, c_K\}$. The goal is to match the query image with the representative images of the classes in the support set.


\begin{figure*}[hbt!]
    \centering
    \includegraphics[width=1\textwidth]{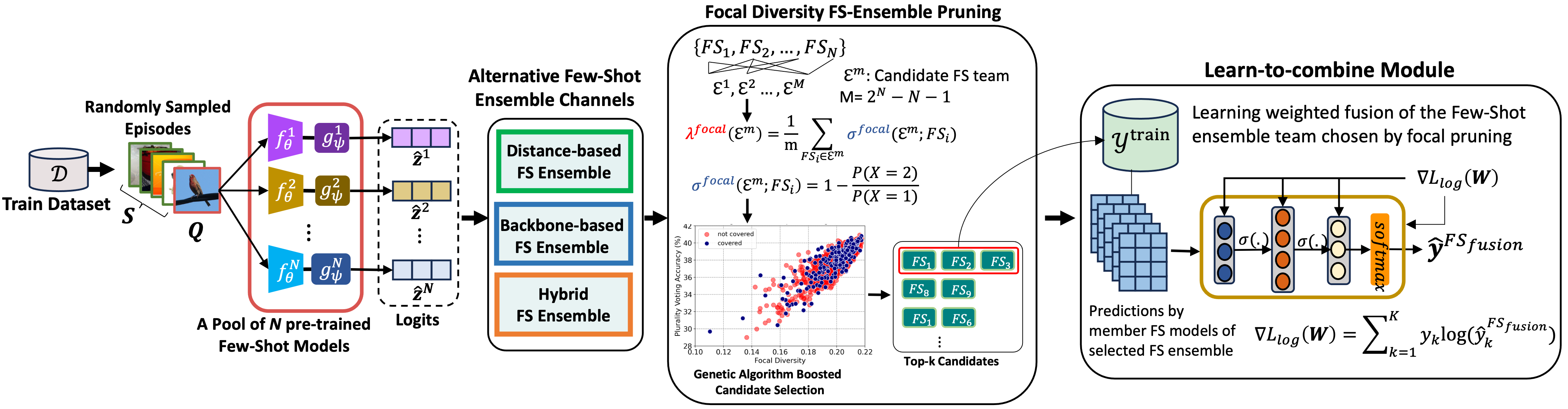}
    \vspace{-15pt}
    \caption{An overview of FusionShot approach to learning ensemble fusion.}
    \label{fig:ensemble}
\end{figure*}

A standard few-shot learner consists of two components, an embedding function $f_{\theta}(.)$ and a latent distance function $g_{\psi}(.)$. For each iteration, an episode $(\mathcal{S},\mathbf{Q})$ pass to the embedding function $f_{\theta}(.)$ to generate latent embedding representations of the support set $\{\mathbf{h}^{c_{1}}_{1}, \dots, \mathbf{h}^{c_{K}}_{J}\}$ and the query image $\mathbf{h}_{q}$. Second, the latent distance function $g_{\psi}(.)$ takes all $J$ latent embeddings of the same class $c_i$ to obtain a class representation per $K$ classes in the support set, denoted by $\{\mathbf{e}_{{c}_{1}},\dots,\mathbf{e}_{c_{K}}\}$. Then, it employs a latent-distance-method $\mathbf{z}=d_{\psi}(\mathbf{h}_q, \mathbf{e}_{c_i})$ to compute the distance between query embedding $\mathbf{h}_q$ and each of the $K$ class embedding $\mathbf{e}_{c_i}$. The distance vector $\mathbf{z}$ becomes the logits which are normalized by the \textit{softmax}. Thus, the final representation is the probability of the query belonging to one of the classes in the support set, i.e., $softmax(\mathbf{z}) = p_{\phi}(\mathbf{y}'|\mathcal{S}, \mathbf{Q})$ where $\phi = (\theta, \psi)$ is the parameters of few-shot learner and $\mathbf{y}'\in \mathbb{R}^{K}$ representing one-hot vector with K dimensions. We provide a visual explanation of the process in Figure \ref{fig:episode} in the Appendix.

Overall, given a dataset $\mathcal{D}$, the objective is to find the model parameters that maximize the likelihood:
\vspace{-3pt}
\begin{equation}
\phi_{ML} = \underset{\phi} {\arg\max} \mathop{\mathbb{E}_{(\mathcal{S}, \mathbf{Q}) \sim \mathcal{D}}} [\log p_{\phi}(\mathbf{y}'|\mathcal{S}, \mathbf{Q})].
\label{eq:mle}
\vspace{-4pt}
\end{equation}
The equation \ref{eq:mle} can be reduced to minimize the \textit{cross-entropy} loss on the model parameters $\mathbf{\phi}$ and probability over the support set of $K$ classes in each episode sampled from train dataset. In each iteration, the parameters are updated by 
SGD. There are alternatives to the cross-entropy loss in the literature, e.g., MSE \cite{sung2018learning}.


\section{Design Overview of FusionShot}
\vspace{-2pt}
For a given pool of $N$ few-shot (FS) models, assume that we want to construct the few-shot ensemble with all $N$ base models as its members. Consider each of the $N$ models is trained independently on both deep embedding for extracting latent features and metric space latent distance learning. Hence, we have $N_{bb}$ different feature extraction embedding methods and $N_{dist}$ different latent distance methods, which gives us $N_{bb}\times N_{dist}$ number of combinations to create few-shot ensembles through three alternative ensemble channels. First, we can choose to use the same DNN backbone architecture for latent feature extraction through deep embedding learning and then leverage multiple (up to $N$) latent space distance functions to ensemble $N$ member models and learn the ensemble prediction through a few shot fusion by using the learn to combine algorithm. We call this channel the metric space latent distance-based ensemble learning (Fusion$^{dist}$). Second, we can alternatively choose to leverage the different DNN backbone architectures for deep embedding fusion and leverage a unifying latent distance function to produce the few-shot ensemble prediction. We call this type of ensemble channel the backbone-based few-shot ensemble learning (Fusion$^{bb}$).
Finally, we may also choose to create few-shot ensembles by a hybrid method, which performs fusion at both the backbone deep embedding learning phase and the metric space latent distance computation phase. We call this third channel the hybrid ensemble learning (Fusion$^{hybrid}$).

Figure~\ref{fig:ensemble} shows a sketch of the generic FusionShot architecture applicable to all three types of few-shot ensemble channels. The sampled episode data is sent to $N$ pre-trained base models. We collect the logits produced by each of the $N$ few-shot base models. Users can choose one of the three types of FS ensemble channels, the FusionShot system will first perform focal diversity-based ensemble pruning to identify the top-$K$ subensemble teams from the pool of $N$ base models, which has a high probability of outperforming the largest ensemble of $N$ models. Next, for a chosen few-shot ensemble team of size $m$ ($m\leq N$), the FusionShot system will need to produce its ensemble prediction for each few-shot query through the learn-to-combine module, which learns the weight assignment function to combine the $m$ possibly conflicting few-shot predictions to generate the FusionShot prediction result. We will discuss the focal diversity-based ensemble pruning and the FusionShot learn to combine in the subsequent sections. 


\vspace{-8pt}
\subsection{FS Ensemble Pruning with Focal Diversity}
\vspace{-2pt}
Consider a pool of $N$ base FS models, the total number of ensemble teams with size $m$ ($2\leq m \leq N$) can be computed by $2^{N}-N-1$~\cite{wu2021boosting}. Let $M$ denote the total number of possible sub-ensemble teams from a pool of $N$ base models. For $N=5$, we have $M=26$ and for $N=10$, we have $M=1013$. As $N$ gets larger, $M$ grows exponentially.
A key question is how to effectively perform ensemble pruning. 

\textbf{Focal Negative Correlation and Focal Diversity.\/}
We introduce two episode-based disagreement metrics: the focal negative correlation metric, $\sigma^{focal}$, and the focal diversity metric $\lambda^{focal}$. The former is used to quantify the level of error diversity among the component models of an ensemble concerning each model within the ensemble. The latter is used to quantify the general error diversity of the ensemble by taking into account all focal negative correlation scores of an ensemble. As we show in Figure \ref{fig:ensemble}, consider a few-shot ensemble $\mathcal{E}^m$, composed of $m$ models: $\{{FS}_1, \dots, {FS}_i,\dots, {FS}_m\}$, we choose one of the $m$ base models each time as the focal model to compute the focal negative correlation score of this ensemble, denoted as $\sigma^{focal}(\mathcal{E}_m; {FS}_i)$. We define the focal diversity of this ensemble team by the average of the $m$ focal negative correlation scores.
The procedure of computing the focal negative correlation score of $\sigma^{focal}$ is as follows: 
(i) select a base model among the set of $m$ base models as the \textit{focal} model, (ii) take all the validation episodes that the focal model has failed and calculate the focal negative correlation score, (3) repeat the previous steps until all $m$ focal negative correlation scores are obtained. $\{\sigma^{focal}_1, \dots, \sigma^{focal}_m\}$, and (4) compute the average over the scores to obtain the focal diversity of ensemble $\mathcal{E}^m$, denoted by $\lambda^{focal}(\mathcal{E}^m)$:
\begin{equation}
\begin{split}
\lambda^{focal}(\mathcal{E}^m)=\nicefrac{1}{m}\times\sum_{\mathcal{E}^{m} \in {FS}_i} \sigma^{focal}(\mathcal{E}^m; {FS}_i)\\
\sigma^{focal}(\mathcal{E}^m; {FS}_i) = 1 - \nicefrac{\sum_{j=1}^{M}\frac{j(j-1)}{m(m-1)}p_j}{\sum_{j=1}^{M}\frac{j}{M}p_j}
\end{split}
\vspace{-8pt}
\end{equation}
Here $p_i$ is the probability that $i$ number of models fail together on a randomly chosen episode. We calculate as $p_i={n_i}/{L^{val}}$ where $n_i$ is the total number of episodes that $i$ number of models failed together on the $\mathcal{Y}^{\mathrm{val}}$ and $L^{val}$ is the total number of validation episodes. The nominator in $\sigma^{focal}$ represents the probability of two randomly chosen models simultaneously failing on an episode, while the denominator represents one randomly chosen model failing on an episode. The terms beneath $p_j$ values are the probability of the chosen model being one of the failures. For example, when $M=3$, there are three cases of model failures; one, two, or three models can fail simultaneously. If one model fails, the chance of selecting the failed model is $1/3$. Similarly, for two models, it is $2/3$, and for three models, it is $1$.
In the case of minimum diversity, the probability of two randomly chosen models failing together comes down to the probability of one of them failing, which makes the fraction term equal to 1 and $\sigma^{focal} = 0$. Similarly, in the case of maximum diversity, there are no simultaneous failures. Hence, the nominator equals 0 and $\sigma^{focal} = 1$. 

\textbf{Ensemble Pruning Strategy and Optimization.\/}
Our goal is to select an ensemble whose member models do not make the same error in their predictions and yet through our FusionShot learn to combine modules, we can effectively resolve the inconsistency and produce correct predictions. 
Figure~\ref{fig:heat_map} shows the computation of focal diversity score for a given pool of $N=10$ base FS models (see Appendix B.7 for the 10 few-shot models trained on {\it mini}-ImageNet). For $M=1013$ possible few-shot ensemble teams from the pool of $N=10$ base models, we plot them by their focal diversity scores and their ensemble prediction accuracy using plurality consensus voting, with the colors representing the size of the ensemble teams and the black dot-line representing the best-performing model accuracy. 
We make two interesting observations: (i) there are more than 10 sub-ensemble teams of size 2-4 that outperform the largest ensemble of size $10$, and (ii) a majority of these smaller ensemble teams also outperform the best-performing individual model in the base model pool (in our case DeepEMD~\cite{zhang2020deepemd}). This indicates that the focal error diversity is a good diversity metric and our focal diversity-optimized ensemble pruning can find high-performance ensembles of low complexity. Concretely, we create a focal pruning score metric by taking the convex combination of the diversity and validation accuracy of each ensemble set, i.e., $s_i=w_1 a_{i}+w_2 \lambda_{i}$. While $a_{i}, \lambda_{i}\in[0, 1]$ represent accuracy and focal diversity scores for ensemble set $\mathcal{E}_{i}$. The weights $w_1 + w_2 = 1$ represent the importance that one can put on the metrics to calculate the pruning scores. This allows us to create an ensemble selection strategy that focuses more on diversity and less on accuracy and vice versa.





\vspace{-8pt}
\begin{figure}[hbt!]
  \centering
    \includegraphics[width=0.25\textwidth]{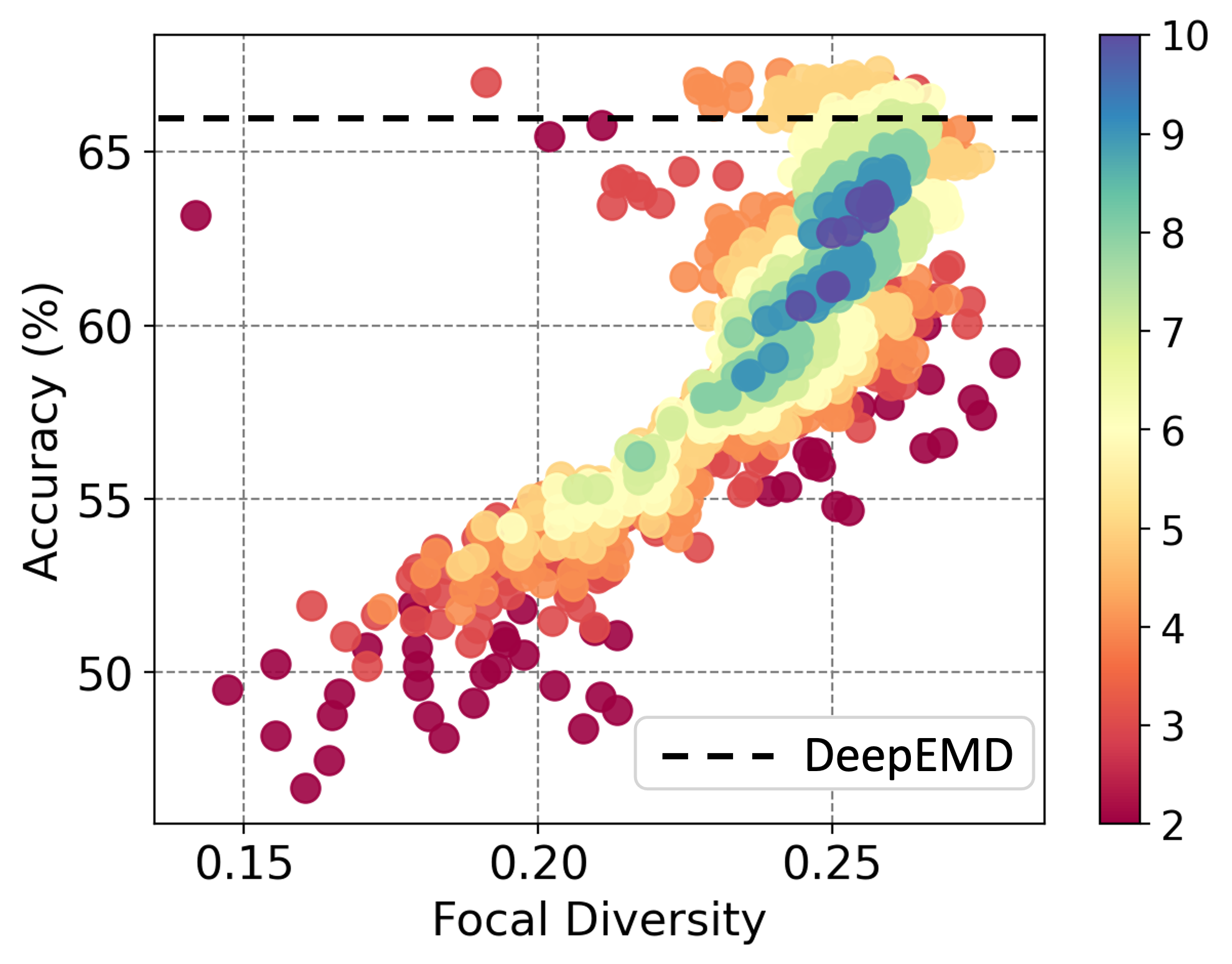}
  \vspace{-13pt}
  \caption{{\small All $M=1013$ ensemble teams from $N=10$ base model pool plotted with their focal diversity scores and their validation accuracy. The colors represent the size of each team, and the dotted line represents the best-performing individual model in the pool.}}
  \vspace{-4pt}
  \label{fig:heat_map}
\end{figure}

\vspace{-15pt}
\begin{table}[hbt!]
    \begin{adjustbox}{width=0.25\textwidth, center}
   \centering
    \small
    \begin{tabular}{p{1.1cm} p{1.1cm} p{0.7cm} p{0.8cm}}
      \hline
      \multirow{2}{1cm}{\# Base Models} & \multicolumn{3}{c}{Time(s)} \\
      \cline{2-4}
      & BF & GA & Gain\% \\
      \hline
      5 & 9.4 & 9.9 & -5.50 \\
      10 & 228.2 & 24.5 & 828 \\
      15 & 508.99 & 41.8 & 1116 \\
      20 & 16201.51 & 54.9 & 29402 \\
      \hline
    \end{tabular}
    \end{adjustbox}
\caption{{\small Comparison of Brute Force (BF) and GA implementation.}}
    \label{fig:speed}
  \vspace{-12pt}
\end{table}


\vspace{-5pt}
\noindent
{\bf Speeding up ensemble pruning with genetic algorithm.\/} To perform focal diversity-based ensemble pruning, we need to compute the focal diversity scores for all $M=2^{N}-N-1$ sub-ensemble teams when given a pool of $N$ base models. For $N=20$, we need to compute the focal diversity score for all $1,048,555$ candidate ensemble teams. 
The brute force (BF) approach is computing the focal diversity and the ensemble accuracy for each candidate ensemble of size $m$ ($2\leq m\leq N$). To speed up this process, we leverage the Genetic Algorithm (GA) \cite{mirjalili2019genetic}, which takes significantly less time to reach the best combination. Table \ref{fig:speed} shows a comparison. 
The Genetic algorithm requires (i) the representation of a candidate solution and (ii) a fitness function to evaluate the solutions. We use the binary vector, where each index represents the presence of the base model in the ensemble set, to represent a solution $\mathbf{\alpha}_i$. We use our pruning score calculation as the fitness function, i.e., $r(\mathbf{\alpha}_i) = w_1a_{i}+w_2\lambda_{i}$.  The initial population contains randomly created candidate solutions. During selection, the most fitted solutions survive to the next population. As the last step, we reproduce new solutions by performing a cross-over among the best-fitted solutions. The procedure is repeated until we reach a plateau or a predetermined fitness function value. As shown in Table \ref{fig:speed}, for a pool of $N=20$ pre-trained base models, we complete the focal diversity-based ensemble pruning in under a minute, achieving 5 orders of magnitude speed up (see Appendix C.5 for further illustration and details).

\textbf{The top-k selection and randomization.\/}
Users may configure FusionShot to get top-$k$ sub-ensembles by the highest error diversity based on their focal diversity scores. To select the top-$k$ teams, we record ensemble teams of size $m$ ($2\leq m\leq N$) with their pruning scores while traversing the candidate space with GA. This allows us to return top-$k$ candidate ensemble teams when the algorithm converges.
Then we can either select the top ensemble or make a random selection among the top-$k$ recommended ensembles. Randomization allows us to create another layer of security compared to a fixed ensemble defense team against 
adversaries~\cite{wenqiWei-TDSC}.

\vspace{-4pt}
\subsection{FusionShot Prediction: Learn-to-Combine}
\vspace{-2pt}
Given an episode ($\mathbf{S}, \mathcal{Q}$), a few-shot ensemble learner of $m$ component models will send the episode to all $m$ individual component models. Let each model $j$ produce the output $\mathbf{z}^{j} = g_{\psi}(f_{\theta}(\mathcal{S}, \mathbf{Q}))$, where $\mathbf{z}^{j}\in \mathbb{R}^{K}$ is the logits, and $1\leq j\leq m$. For $m$ component models there are $m$ different logits against the query $\mathbf{Q}$ over the support set $\mathcal{S}$, corresponding to $m$ different and possibly conflicting predictions. Thus, we have $\mathbf{z}^{1}, \dots, \mathbf{z}^{m}$. The goal of the FusionShot ensemble learner is to learn the most robust way to combine the $m$ different logits to generate the ensemble output against the query $\mathbf{Q}$ for each episode sampled from a dataset $\mathcal{D}$. Specifically, for an episode, the objective is to maximize $p(y=k|\mathbf{z}^{1}, \dots, \mathbf{z}^{m})$ for mapping the query $\mathbf{Q}$ to one of the support categories, say $c_{k}$. 
We parameterize the likelihood with a Multi-Layer Perception (MLP), $p_{\gamma}(y=k|\mathbf{z}^{1}, \dots, \mathbf{z}^{m})$, to approximate the probability, where $\gamma$ denotes the learnable model parameters for the MLP. The MLP contains multiple layers of fully connected weights with sigmoid activation functions. At the final layer, the model performs softmax to produce the probability for each of the $K$ categories
in the support set. The ensemble learner will output the ensemble fusion optimized prediction based on the logits layer of the MLP:
\vspace{-4pt}
\begin{align}
    \Tilde{\mathbf{y}} &= \mathrm{softmax}(\mathbf{W}_{\ell} (\dots \sigma(\mathbf{W}_{1}[\hat{\mathbf{z}}^{1}, \dots,\hat{\mathbf{z}}^{m}])\dots)) \label{eq:fusion},
    \vspace{-6pt}
\end{align}
where $\ell$ is the number of layers and $\Tilde{\mathbf{y}}$ is the final prediction. The first layer takes the concatenation of the logits as the input, i.e., $\mathbf{W}_1\in\mathbb{R}^{(mK)\times d}$ where $d$ is the input dimension of the second layer. We want to find the best parameters $\gamma = (\mathbf{W}_{1},\dots,\mathbf{W}_{\ell})$ to maximize the likelihood, which can be reduced to minimize the cross-entropy loss on a dataset which is the collection of logits for each component model. Therefore, we sample episodes from incoming data $\mathcal{D}^{\mathrm{train}}$ to feed them to the $m$ component models. As shown in the right side of Figure \ref{fig:ensemble}, we collect all the logits produced by the component models and create the fusion training set $\mathcal{Y}^{\mathrm{train}}=\{\mathbf{z}_{i}^{j}\}_{i=1,j=1}^{m, E^{train}}$. Let $E^{train}$ be the number of episodes we sample. Similarly, we also sample logits to create the validation set $\mathcal{Y}^{\mathrm{val}}$ for early-stopping.

\begin{table*}[t]
    \begin{adjustbox}{width=0.65\textwidth, center}
    \centering
    \small
    \begin{tabular}{p{1.5cm} c p{1.2cm} p{1.2cm} p{1.2cm} p{1.2cm} p{1.2cm} | p{1.2cm} c}
        \hline
        Method & Dist. & Conv4 & Conv6 & ResNet10 & ResNet18 & ResNet34 & $\mathrm{Fusion}^{\mathrm{bb}}$ & Gain \\
        \hline
        Matching & Cosine & $51.73_{0.75}$  & $47.95_{0.79}$  & $50.85_{0.84}$  & $50.89_{0.78}$  & $50.90_{0.84}$  & \hl{$\mathbf{56.79}_{0.45}$}  & 9.78\% \\
        Prototypical & L2 & $48.42_{0.79}$  & $49.17_{0.79}$  & $53.01_{0.78}$  & $51.72_{0.81}$  & $53.16_{0.84}$  & \hl{$\mathbf{57.39}_{0.45}$}  & 7.96\% \\
        MAML & MLP & $45.51_{0.77}$  & $47.13_{0.84}$  & $50.83_{0.84}$  & $47.57_{0.84}$  & $48.92_{0.83}$  & \hl{$\mathbf{54.74}_{0.44}$}  & 7.69\% \\
        Relation & CNN & $48.57_{0.82}$  & $48.86_{0.81}$  & $49.26_{0.85}$  & $47.07_{0.77}$  & $48.30_{0.77}$  & \hl{$\mathbf{53.75}_{0.47}$}  & 9.11\% \\
        Simpleshot & KNN & $48.90_{0.73}$  & $50.26_{0.75}$  & $61.38_{0.81}$  & $62.61_{0.80}$  & $61.96_{0.77}$  & \hl{$\mathbf{65.09}_{0.45}$}  & 3.96\% \\
        \hline
        $\mathrm{Fusion}^{\mathrm{dist}}$ & Cos, L2, MLP, CNN, KNN & \colorbox{pink}{$\mathbf{53.28}_{0.48}$} & \colorbox{pink}{$\mathbf{52.70}_{0.45}$} & \colorbox{pink}{$\mathbf{62.36}_{0.44}$} & \colorbox{pink}{$\mathbf{64.26}_{0.44}$} & \colorbox{pink}{$\mathbf{64.46}_{0.41}$} & & \\
        Gain & & 3.00\%  & 4.85\%  & 1.60\% & 2.64\%  & 4.03\% & & \\
        \hline
    \end{tabular}
    \end{adjustbox}
    \caption{{\small FusionShot ensemble performance for the fusion of 5 backbone algorithms (columns) and the fusion of 5 distance functions (rows).}}
    \vspace{-15pt}
    \label{table:ens_results}
\end{table*}

\begin{figure*}[t]
\centering
    \begin{subfigure}{0.3\textwidth}
        \centering
        \includegraphics[width=\textwidth]{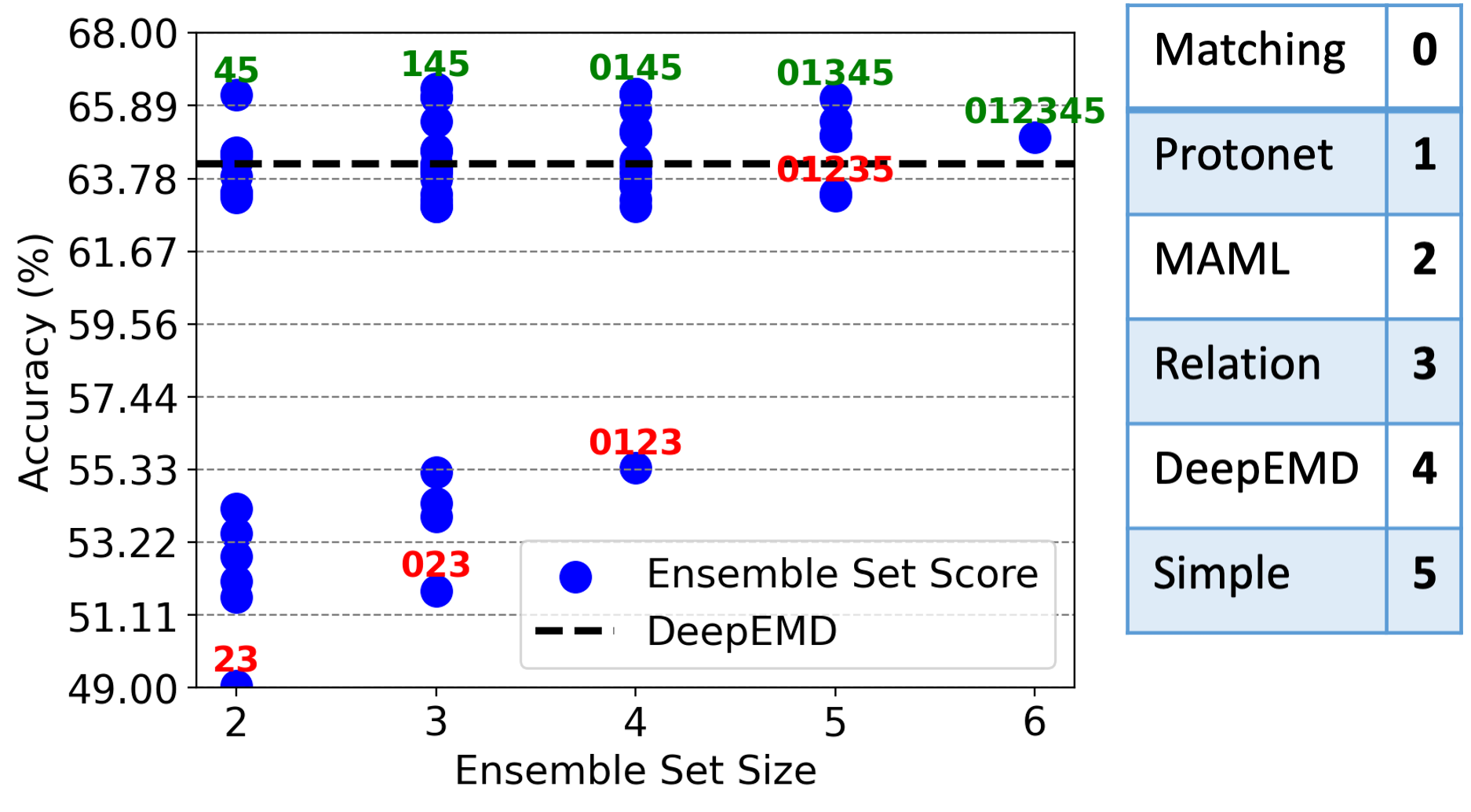}
        \vspace{-15pt}
        \caption{}
        \label{fig:scatter_method}
    \end{subfigure}
    \begin{subfigure}{0.3\textwidth}
        \centering
        \includegraphics[width=\textwidth]{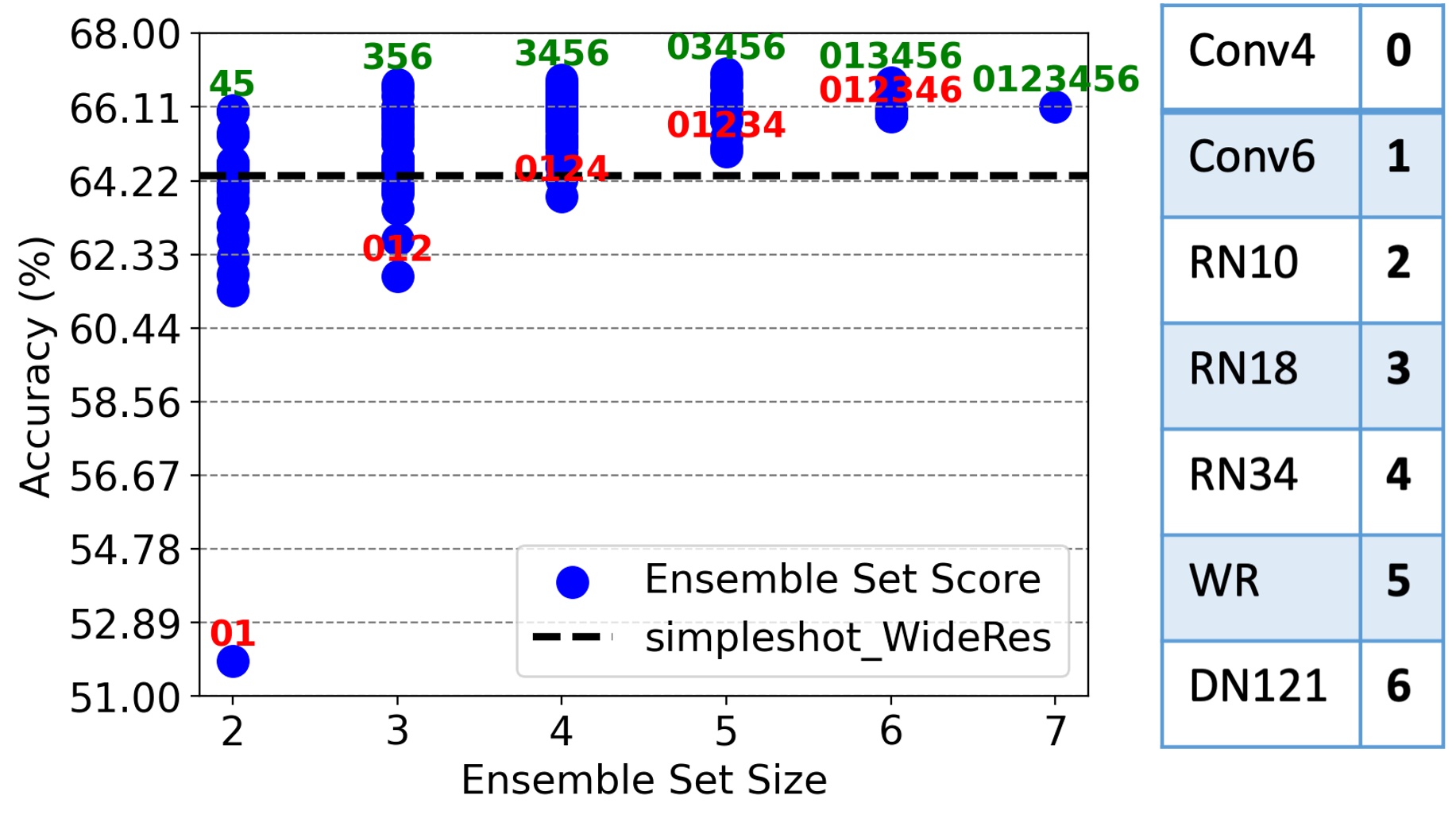}
        \vspace{-15pt}
        \caption{}
        \label{fig:scatter_backbone}
    \end{subfigure}
    \begin{subfigure}{0.35\textwidth}
        \centering
        \includegraphics[width=\textwidth]{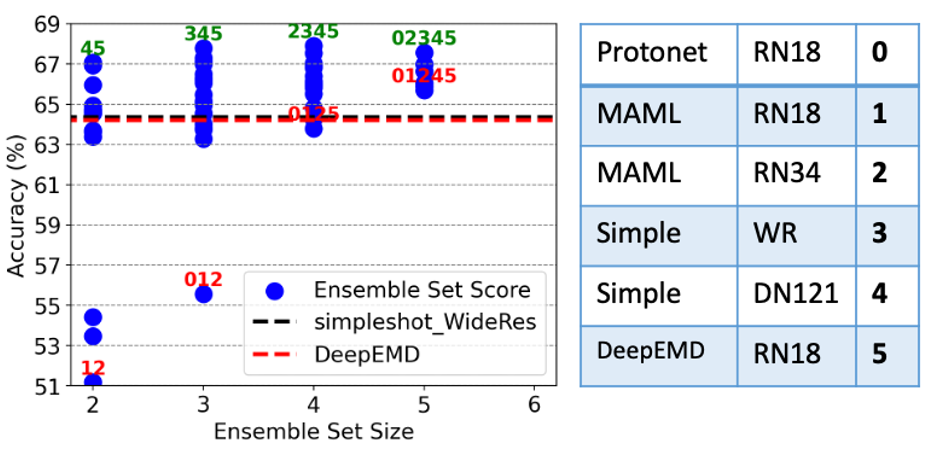}
        \vspace{-15pt}
        \caption{}
        \label{fig:scatter_mix}
    \end{subfigure}
    \vspace{-12pt}
    \caption{{\small Performance of 1-shot 5-way ensembles (\textit{mini}-Imagenet) produced by distance function fusion (a), by backbone fusion (b), and hybrid of both distance function and backbone fusion (c). The green and red texts represent the best and worst performing sets among the candidate ensemble sets of the same team size. The horizontal line is the performance of the best base model.}}
    \label{fig:scatter}
    \vspace{-8pt}
\end{figure*}

\section{Experiments}
\vspace{-2pt}
\textbf{Datasets and Scenarios.\/} 
Following the literature \cite{chen2019closer}, we use three benchmarks: For object recognition, we use \textit{mini}-ImageNet which is the common dataset to evaluate Few-shot models. The dataset contains 100 classes from ImageNet, 600 samples per class. For ease of comparison, we follow the standard partition suggested in ~\cite{ravi2016optimization}. For fine-grained image classification, we use Caltech-UCSD Birds-200-2011 (CUB), containing 11,788 samples belonging to 200 classes and split it into 100 training, 50 validation, and 50 novel classes following~\cite{hilliard2018few}. 
For the cross-domain scenario, we follow the standard approach, which trains the model on one dataset and evaluates it on another, e.g., \textit{mini}-ImageNet$\rightarrow$CUB. The goal is to discover the effects of domain shifts to the few-shot models. 

\textbf{Evaluation.\/} In the performance analysis, we follow the standard settings in few-shot learning, which are 5-ways 1-shot and 5-ways 5-shot classification $(J=\{1, 5\}, K=5)$. Since few-shot learners perform episodic learning where each episode chooses its samples randomly, we report the mean accuracy (\%) and 95\% confidence interval of the novel set, $\mathcal{Y}^{\mathrm{novel}}$, with 600 episodes.

\textbf{Model Pool.\/} We analyze the effect of our ensemble channel by using seven backbone architectures: Conv4-6 \cite{vinyals2016matching}, ResNet10, ResNet12, ResNet18, ResNet34, WideResnet, and DenseNet121, and we employ six alternative parametric/non-parametric methods, such as  ProtoNet, MatchingNet, RelationNet, Model-Agnostic Meta-Learning (MAML), DeepEMD, and Simpleshot. Note that, all the candidate models are trained on base and validation sets and ready to be combined by our fusion model. The details of candidate model training and their hyper-parameters are given in Appendix B.1-4. 


\textbf{Focal Diversity based Ensemble Pruning.\/} We selected $w_1=0.6$ and $w_2=0.4$ while scoring a candidate ensemble set to give more importance to the diversity. The genetic algorithm stops when the fitness function does not change for 100 consecutive generations. 



\textbf{Fusion Model.\/} The model contains two fully connected hidden layers with 100 neurons, with sigmoid activations between the layers. The last layer follows a softmax operation for the final prediction. The model optimizes its parameters using the Adam optimizer for 150 epochs, employing a learning rate of 0.001.



\vspace{-5pt}
\subsection{Performance of FusionShot}
\vspace{-2pt}
\textbf{The Effect of Backbone and Distance Function.\/}
Table \ref{table:ens_results} shows the performance of the fusion model on \textit{mini}-Imagenet dataset using five backbone algorithms (see columns) and five metric-space distance functions (see rows) without performing ensemble pruning.
Row-wise, we show that using the fusion channel for feature extraction ensemble, the fusion model achieves the  top-1 performance improvement by 7.69 - 9.78\% over the classical few-shot models, and 4\% for Simpleshot. In column-wise, we show that using the fusion channel for latent distance ensemble, we get up to 4\% top-1 performance improvements. The results indicate that the performance increases with the diversity of either backbone or distance function. In the next set of experiments, we analyze the sub-ensembles for each type of diversity.

We zoom in and take a closer look at the ResNet18 column in Table~\ref{table:ens_results} to analyze the performance impact of ensemble fusion by employing multiple distance functions. We add DeepEMD to the collection to get a total of six metric-space distance methods and keep the backbone architecture the same, ResNet18. Figure \ref{fig:scatter_method} shows the performance measurement of all the trained fusion models for each sub-ensemble set over \textit{mini}-Imagenet dataset in the 1-shot 5-way few-shot setting. Figure \ref{fig:scatter_method} also shows the top and worst performing sub-ensembles of varying team sizes. We observe that multiple ensemble teams outperform DeepEMD, the best individual model in the pool of six latent distance models.
The ensemble of Simpleshot and DeepEMD further improves the top-1 prediction over DeepEMD. By adding Protonet, the ensemble of the three models (145) outperforms most of the other ensembles, including the 2-model ensemble of Simplenet and DeepEMD, and the 6-model ensemble (012345).

We perform a similar experiment to measure the effectiveness of ensemble fusion by combining multiple backbone algorithms for feature extraction under one fixed latent distance comparison method. We chose Simpleshot in this set of experiments and paired it with one of the 7 backbone DNN architectures: Conv4, Conv6, ResNet10, ResNet18, ResNet34, WideResNet, and DenseNet121. From Figure \ref{fig:scatter_backbone}, we make two observations. (1) A large number of the ensemble teams show higher performance than the best-performing base model, which is SimpleShot with WideResNet. (2) The best-performing ensemble teams, highlighted in green, have high focal diversity and the most diverse backbone architectures, e.g., ResNet18, DenseNet121, and WideResNet form the best-performing 3-model ensemble \{356\}.
Both of these results show the importance of focal diversity-optimized ensemble pruning and the learn-to-combine parametric approach, which is the fusion model. Secondly, either backbone or distance method fusion improves the diversity of the ensemble set and consequently enhances the performance of the fusion model. Thus, we obtain the best results when both latent distance metric and feature extraction selections are mixed as shown in Figure \ref{fig:scatter_mix} reaching 67.79\% with more than 3\% improvement over the best performing base model.

\begin{table*}[t]
  \begin{adjustbox}{width=1\textwidth, center}
  \small
    \centering
    \begin{tabular}{l c c c c c c c}\\
        \hline
        \multirow{2}{0.5cm}{Method} & \multirow{2}{0.5cm}{Backbone} & \multirow{2}{1.2cm}{Dist. Func.} & \multirow{2}{3cm}{Backbone Embed. Dim.} & \multicolumn{2}{c}{\textit{mini}-Imagenet} & \multicolumn{2}{c}{CUB} \\
        \cline{5-8}
        & & & & 1-shot & 5-shot & 1-shot & 5-shot \\
        \hline
        Matching \cite{vinyals2016matching} & ResNet18 & Cosine & 512 & ${50.85}_{0.78}$ & ${68.46}_{0.67}$ & ${73.49_{0.89}}$ & ${83.64_{0.60}}$ \\
        Prototypical \cite{snell2017prototypical} & ResNet18 & L2 & 512 & ${51.72}_{0.81}$ & ${73.88_{0.65}}$ & ${72.07_{0.93}}$ & ${85.01_{0.60}}$ \\
        Relation \cite{sung2018learning} & ResNet18 & CNN & 512 & ${47.07}_{0.77}$ & ${69.03}_{0.70}$ & ${68.58_{0.94}}$ &  ${82.75_{0.58}}$ \\
        MAML \cite{finn2017model} & ResNet18 & MLP & 512 & ${47.57}_{0.84}$ & ${66.20}_{0.78}$ & ${68.42_{1.07}}$ &  ${82.70_{0.65}}$ \\
        Simpleshot \cite{wang2019simpleshot} & ResNet18 & KNN & 512 & ${62.61}_{0.80}$ & ${78.33}_{0.59}$ & ${63.50_{0.88}}$ & ${82.63_{0.65}}$ \\
        Simpleshot \cite{wang2019simpleshot} & DenseNet121 & KNN & 1024 & $63.53_{0.79}$ & $79.11_{0.55}$ & ${62.82}_{0.21}$ & ${81.82}_{0.15}$ \\
        Simpleshot \cite{wang2019simpleshot} & WideResNet & KNN & 640 & $64.34_{0.79}$ & $78.80_{0.59}$ & ${62.18}_{0.22}$ & ${78.38}_{0.16}$ \\
        DeepEMD \cite{zhang2020deepemd} & ResNet12 & EMD & 512 & $65.21_{0.75}$ & $80.51_{0.54}$ & ${74.39_{0.85}}$ & ${87.65_{0.55}}$ \\
        TADAM \cite{oreshkin2018tadam} & ResNet12 & L2 & 640 & $58.50_{0.30}^{\dagger}$ & $76.70_{0.30}^{\dagger}$ & - & - \\
        FEAT \cite{ye2020few}  & WideResNet & Cosine & 640 & $61.72_{0.11}^{\dagger}$ & $78.49_{0.15}^{\dagger}$ & $68.87_{0.22}^{\dagger}$ & $82.90_{0.15}^{\dagger}$ \\
        LEO \cite{rusu2018meta} & WideResNet & KLDiv & 640 & $61.76_{0.08}^{\dagger}$ & $77.59_{0.12}^{\dagger}$ & - & - \\
        Robust-20$^*$ \cite{dvornik2019diversity} & ResNet18 & Cosine & 512 & $63.73_{0.62}^{\dagger}$ & $81.19_{0.43}^{\dagger}$ & - & - \\
        EASY$^*$ \cite{bendou2022easy} & 3$\times$ResNet12 & L2 & $3\times640$ & $68.56_{0.76}$ & $83.02_{0.34}$ & ${73.16}_{0.91}$ & ${89.58_{0.47}}$ \\
        \hline
        (baseline) \textbf{FusionAll}, $m=10$ & \multicolumn{2}{l}{(RN18-34, DN121, WR, CN4) $+$ (Simpleshot, EMD, Proto, Relation, MAML, Matching) } & ($7\times 512$, 1024, 640, 640) & ${65.34}_{0.44}$ & ${79.98}_{0.39}$ & ${75.65}_{0.51}$ & ${78.49}_{0.55}$ \\
        (ours) $\mathbf{FusionShot}^{\mathrm{dist}}, m=3$ & \multicolumn{2}{l}{ResNet18 $+$ (Prototypical, Simpleshot, EMD)} & (512, 512, 512) & ${66.38}_{0.10}$ & ${81.58}_{0.36}$ & $\mathbf{78.64_{0.38}}$ & ${88.95}_{0.29}$ \\
        (ours) $\mathbf{FusionShot}^{\mathrm{bb}}, m=4$ & \multicolumn{2}{l}{(RN18-34, DN121, WR, CN4) $+$ Simpleshot } & (512, 1024, 640, 640) & ${66.97}_{0.10}$ & ${81.12}_{0.36}$ & ${64.30}_{0.65}$ & ${82.20}_{0.50}$ \\
        (ours) $\mathbf{FusionShot}^{\mathrm{hybrid}}, m=5$ & \multicolumn{2}{l}{(3$\times$ RN12, RN18, DN121) $+$(EASY, SimpleShot, EMD)} & ( $3\times640$, 512, 1024) & $\mathbf{70.76}_{0.44}$ & $\mathbf{83.84}_{0.31}$ & ${77.36}_{0.54}$ & $\mathbf{90.94}_{0.36}$\\
        \hline
    \end{tabular}
    \end{adjustbox}
    \label{table:sota}
  \caption{{\small Comparison with existing SOTA methods on 1-shot 5-way and 5-shot 5-way performance using (a) \textit{mini}-Imagenet and CUB. The works marked with the $^*$ symbol represent other ensemble methods. Data with the $^{\dagger}$ symbol is taken from the corresponding work.}}
  \label{table:all_scores}
  \vspace{-9pt}
\end{table*}

\textbf{Comparison with existing SOTA Few-shot methods.\/} Here we compare {\sc FusionShot} performance with the SOTA methods in the literature under three scenarios: object recognition, fine-grained image classification, and cross-domain classification. In our comparisons, three different {\sc FusionShot} versions are given; fusion model trained on pruned ensemble set containing backbone diversity $\mathrm{FusionShot}^{\mathrm{bb}}$, distance function diversity $\mathrm{FusionShot}^{\mathrm{dist}}$ and mixture $\mathrm{FusionShot}^{\mathrm{hybrid}}$. We also provide a fusion model that is trained with the best SOTA method to show one can always implement {\sc FusionShot} to further improve the performance.

We choose ResNet18 as the fixed backbone architecture for feature extraction with six latent distance comparison methods for latent-distance-based ensemble learning. By focal diversity ensemble pruning, we obtain Protonet, SimpleShot, and DeepEMD with ResNet18 as the best-performing sub-ensemble, denoted by $\mathrm{FusionShot}^{\mathrm{dist}}$. 
Similarly, we choose seven DNN backbone architectures with SimpleShot as the fixed latent distance function as shown in Figure~\ref{fig:scatter_method}. By focal-diversity ensemble pruning, we obtain SimpleShot with Conv4, ResNet18-34, DenseNet121, and WideResNet as the best-performing ensemble, which has four models and outperforms their superset ensemble teams, denoted by $\mathrm{FusionShot}^{\mathrm{bb}}$. 
The comparison includes
several recent SOTA algorithms in Table \ref{table:all_scores}, e.g., TADAM~\cite{oreshkin2018tadam}, FEAT~\cite{ye2020few}, LEO~\cite{rusu2018meta}, Robust-20~\cite{dvornik2019diversity}, and EASY~\cite{bendou2022easy}. We show that the ensembles recommended by our focal-diversity optimized {\sc FusionShot} improves the top-1 prediction performance by up to 4\%, compared to the other methods on both \textit{mini}-Imagenet and CUB datasets for 5-way 1-shot and 5-way 5-shot scenarios. In the last row of Figure~\ref{table:all_scores}, the ensemble of $\mathrm{FusionShot}^{\mathrm{dist}}$ and EASY achieves 2\% performance gain.

\begin{table}[t]
    \centering
    \small
    \begin{tabular}{l p{1.1cm} p{1.5cm} c}
        \hline
        \multirow{2}{1cm}{Method} & \multirow{2}{1cm}{Backbone} & \multirow{2}{1.7cm}{Dist. Func.} & \textit{mini}-Image$\rightarrow$CUB \\
        \cline{4-4}
        & & & 5-shot \\
        \hline
        Matching \cite{vinyals2016matching} & ResNet18 & Cosine & $52.17_{0.74}$ \\
        Prototypical \cite{snell2017prototypical} & ResNet18 & L2 & $55.24_{0.72}$ \\
        Relation \cite{sung2018learning} & ResNet18 & CNN & $50.93_{0.73}$ \\
        MAML \cite{finn2017model} & ResNet18 & MLP & $46.85_{0.72}$ \\
        Simpleshot \cite{wang2019simpleshot} & ResNet18 & KNN & $67.38_{0.70}$ \\
        DeepEMD \cite{zhang2020deepemd} & ResNet12 & EMD & $77.44_{0.70}$ \\
        Robust-20 $^*$ \cite{dvornik2019diversity} & ResNet18 & L2 & $65.04_{0.57}^{\dagger}$ \\
        \hline
        $\mathbf{FusionShot}^{\mathrm{dist}}$ (ours) & \multicolumn{2}{l}{ResNet18 $+$ (Proto, SS, EMD)} & $\mathbf{78.02_{0.38}}$ \\
        \hline
    \end{tabular}
    \caption{{\small We compare the cross-domain performance of base models, Robust-20, and FusionShot, which fuses Protoypical, MAML, SimpleShot, and DeepEMD.}}
    \label{table:cross_domain}
\end{table}

\textbf{Cross-domain Performance.\/} We evaluate the cross-domain performance of {\sc FusionShot} by  training on the \textit{mini}-Imagenet and then testing on the CUB dataset. The ensemble consists of the base models trained on mini-ImageNet and never sees the new CUB dataset, neither in training nor validation. The CUB dataset is used only for testing. This \textit{blind} setting simulates closely the real-world scenarios where each base model is trained on a different dataset. We can take any off-the-shelf models regardless of the training datasets, FusionShot will leverage the task-agnostic properties of the base models and boost their performance on the new dataset. As shown in Table \ref{table:cross_domain}, {\sc FusionShot} shows comparable performance with the best base model, even though there is a 10\% gap between the first two best-performing base models. When we remove DeepEMD, {\sc FusionShot} improves the best base model performance by up to 6\% (see Table \ref{table:closer_look_more} in the Appendix). We also show that  {\sc FusionShot} outperforms Robust-20~\cite{dvornik2019diversity}, an ensemble method using 20 ResNet18 models.

\vspace{-4pt}
\subsection{Robustness Against Adversarial Attacks}
This set of experiments is designed to show that {\sc FusionShot} can be leveraged to protect a victim few-shot model against adversarial attacks. Here DeepEMD is the victim and the PGD attack~\cite{madry2017towards} is used to attack DeepEMD. The details on the PGD attack for few-shot models are provided in Appendix B5. 
In FusionShot defense, the victim model is used as the focal model to compute the focal diversity of all candidate ensembles.
The top-$k$ ensembles that exhibit the highest focal diversity will form our defense ensemble team(s), which will be selected randomly for each few-shot query. Next, we retrieve the logits produced by the defense ensemble when the victim model is under attack and incorporate them into the training set,  denoted as $\mathcal{Y}_{\mathrm{train}}$, for learning how to produce ensemble fused prediction by the learn-to-combine model. Together with the benign and attacked logits, we train the learn-to-combine ensemble fusion model to learn how to combine multiple independent and possibly conflicting predictions under adversarial or benign settings. Training with both attack and benign logits allows the defense ensemble to repair/replace the predictions of the victim model (DeepEMD) with the predictions of our focal diversity-based ensemble fusion model. 

\begin{figure}[t]
\centering
    \begin{subfigure}{0.40\textwidth}
        \centering
        \includegraphics[width=\textwidth]{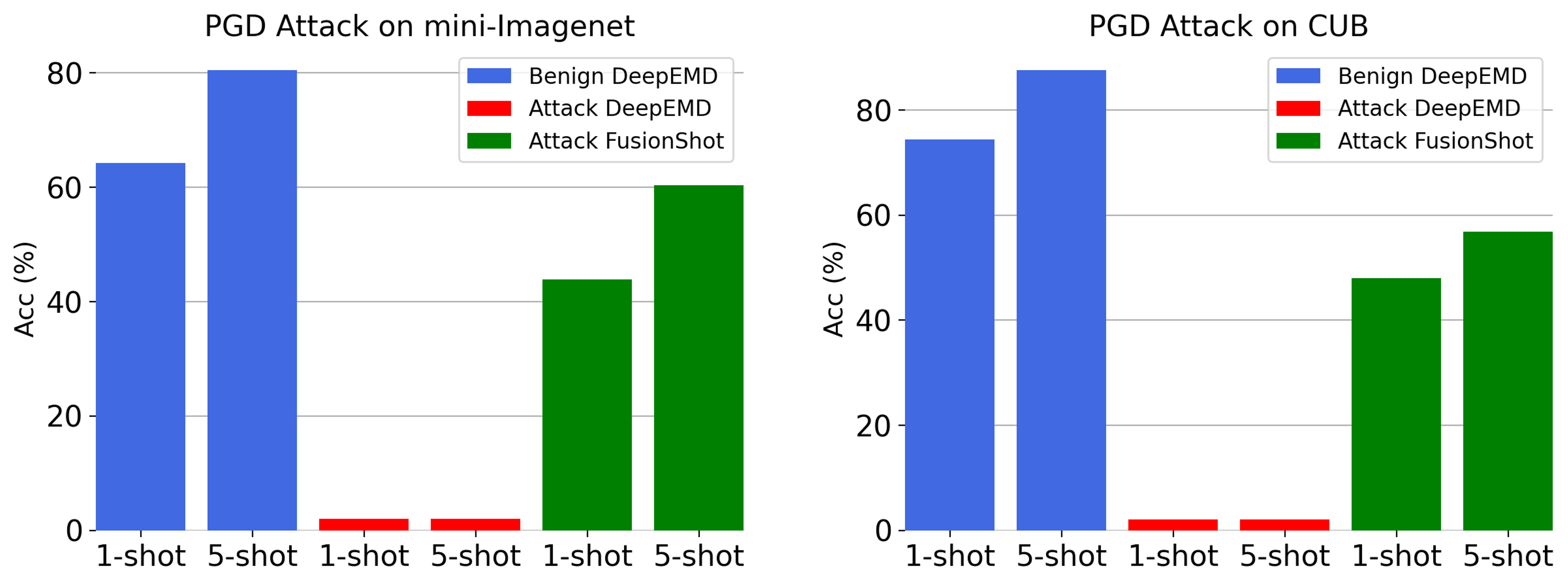}
    \end{subfigure}
    \begin{subfigure}{0.3\textwidth}
        \centering
        \includegraphics[width=\textwidth]{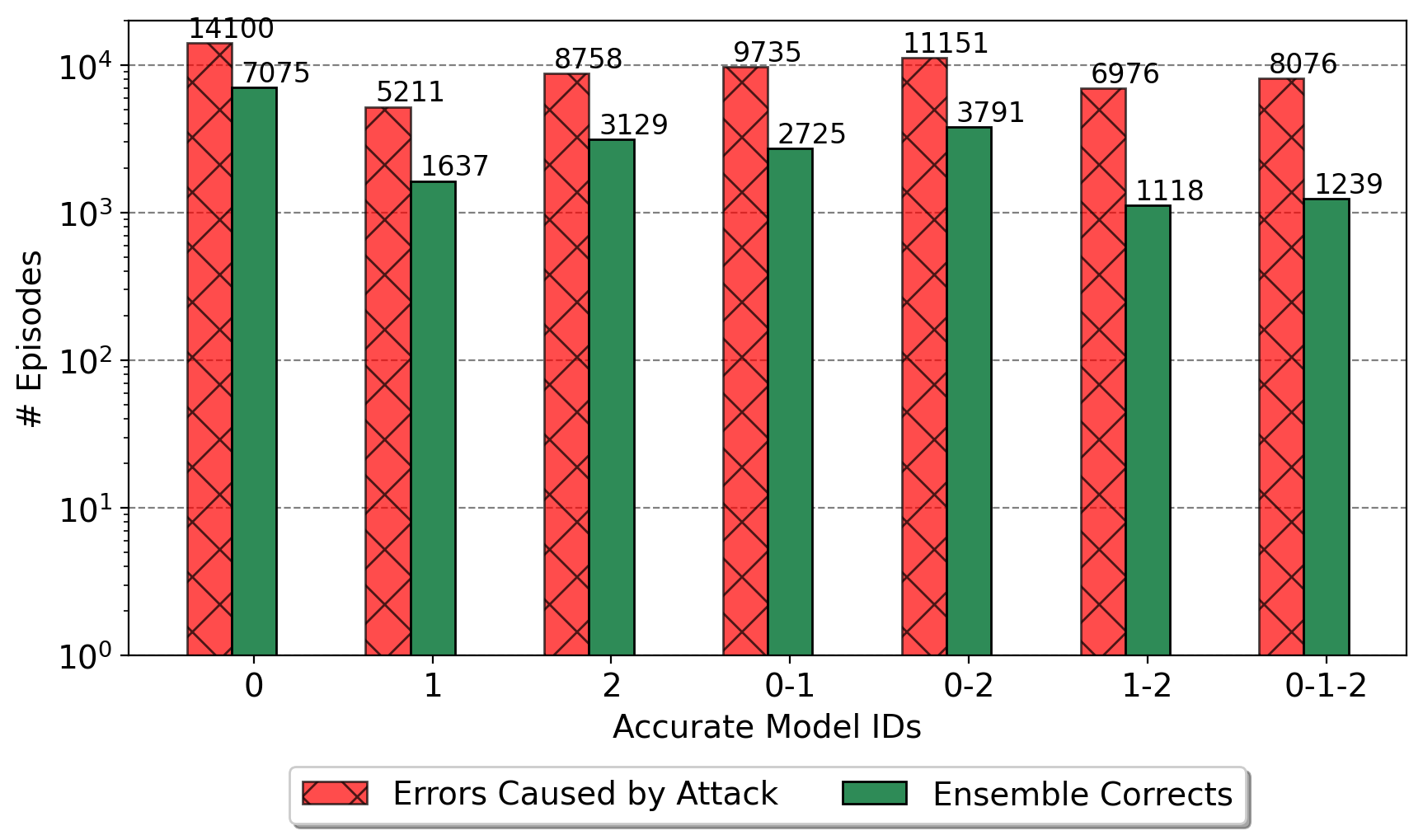}
    \end{subfigure}
    \vspace{-10pt}
    \caption{{\small (Top) The mean accuracy of DeepEMD at 1, 5-shot settings to benign episodes (blue) and PGD attacked episodes (red). We show the mean accuracy of FusionShot to the attacked episodes in green bars. 
    (Bottom) Red bar: \# errors made by single base model or all models in a team out of 45000 novel episodes (1-shot 5-way, \textit{mini}-Imagenet). Green bars: \# corrected episodes by FusionShot.}}
   \label{fig:comparison}
\end{figure}

\begin{table}[htbp]
  \centering
  \small
  \begin{tabular}{lcccccc}
    \hline
    \multirow{2}{1cm}{Method} & \multirow{2}{1cm}{Backbone} & \multicolumn{2}{c}{Benign Acc.} & \multicolumn{2}{c}{Acc. Under Attack}\\
    \cline{3-6}
    & &  1-shot & 5-shot & 1-shot & 5-shot \\
    \hline
    (victim) DeepEMD & Resnet18 & 65.21  & 80.51 & 0.00  & 0.01    \\
    Def: AQ~\cite{goldblum2020adversarially} & Resnet10 & 37.91 & 57.87 & 20.59 & 31.52 \\
    Def: DFSL~\cite{li2022defensive} & Resnet18 & 48.79 & 68.98 & 40.17 & 58.04 \\
    $\mathbf{FusionShot}^{\mathrm{dist}}$ (ours) & Resnet18 & \textbf{66.38}  & \textbf{81.58} & \textbf{43.31}  & \textbf{59.74}  \\
    \hline
  \end{tabular}
  \caption{{\small \textit{Mini}-Imagenet performance of the DeepEMD under PGD attack and FusionShot defense. We also compare with the other defense algorithms in the literature.}}
  \label{tab:performance}
\end{table}

The top bar charts in Figure \ref{fig:comparison} show the mean accuracy of episodes when they are under PGD attack, and benign environment for \textit{mini}-Imagenet and CUB datasets respectively. As shown in Figure \ref{fig:comparison}, FusionShot protects the victim model against adversarial attacks and raises its performance to 43\% and 59\% at 1 and 5 shot settings. Following this result, we dive into the comparison by calculating the number of episodes in which our defense mechanism was successful. The red bar in Figure \ref{fig:comparison} shows the number of episodes in which each model made errors individually or together when the queries are infused with a PGD attack. The x-axis shows the model index of the combinations. The green bar shows the number of infused episodes that are corrected by our focal diversity selected ensemble to protect the victim model. We make three observations: (1) The right-most red bar shows that the ensemble of three ResNet18 models with 3-distance methods (DeepEMD, Simpleshot, and RelationNet) failed together in 8076 attacked episodes, and our ensemble fusion model can successfully correct 1239 of them (15\%). (2) Consider the 2-model ensemble \{0-2\}, there are 11151 episodes in which Simpleshot and DeepEMD made incorrect predictions, and our ensemble fusion method can correct 3791 of them (33\%). (3) For the victim model DeepEMD (left-most), our {\sc FusionShot} ensemble can correct 7075 out of 14100 incorrect episodes, offering over 50\% performance improvement.

Table \ref{tab:performance} shows the comparison of FusionShot with two representative defense methods in the literature. 
 DeepEMD is the SOTA few-shot model and as the victim model here, it is severely impacted by the adversarial attack. In comparison, FusionShot-powered defense can raise the few-shot performance and outperform the other two defense methods in the presence of attack. This shows that our focal diversity-enhanced few-shot ensemble learning can strengthen the adversarial resilience of the victim model against attacks while maintaining superior performance under benign scenarios. 

\begin{figure}[!hbt]
\centering
    \begin{subfigure}{0.37\textwidth}
        \centering
        \includegraphics[width=\textwidth]{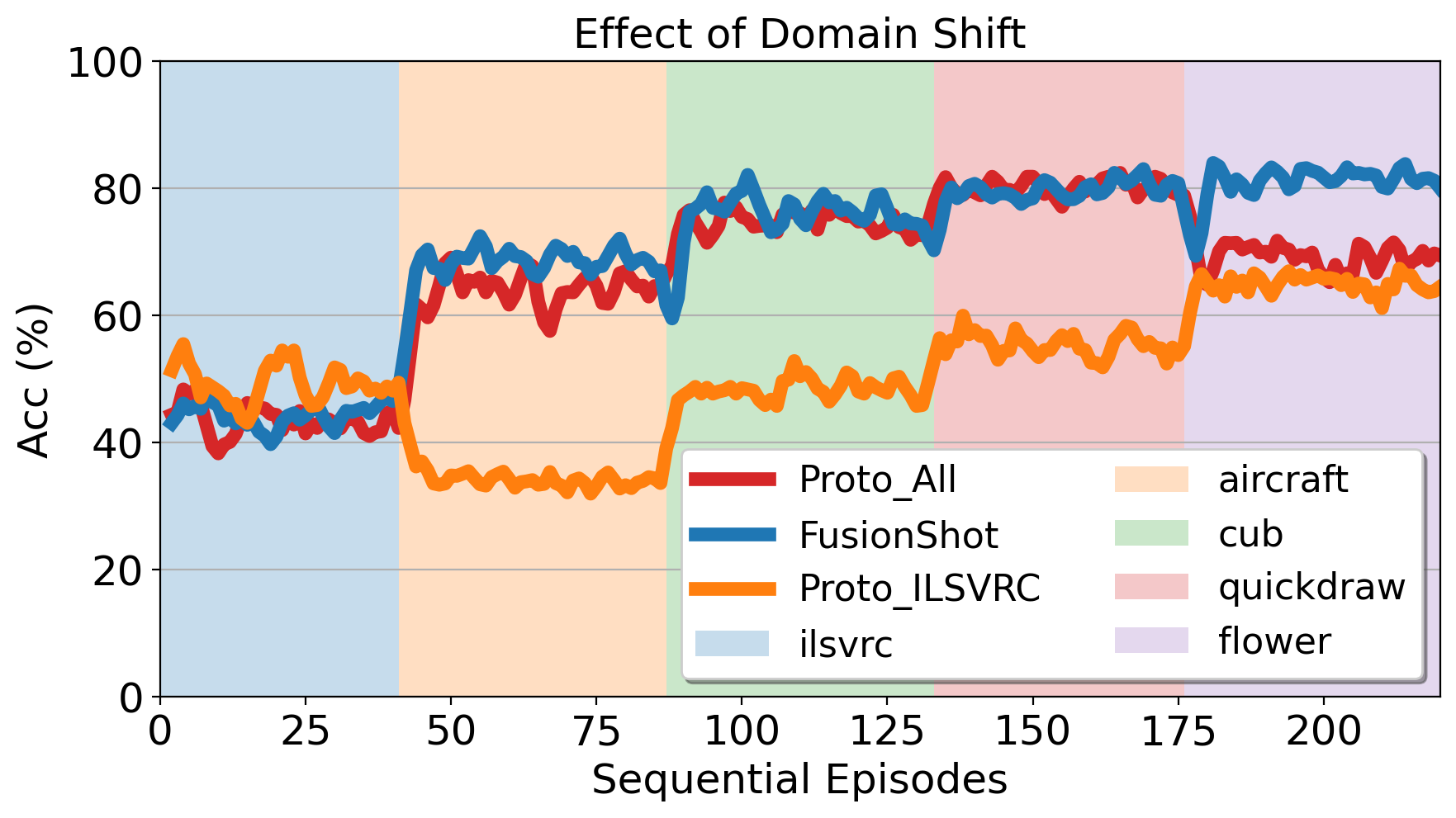}
    \end{subfigure}
    \begin{subfigure}{0.37\textwidth}
        \centering
        \includegraphics[width=\textwidth]{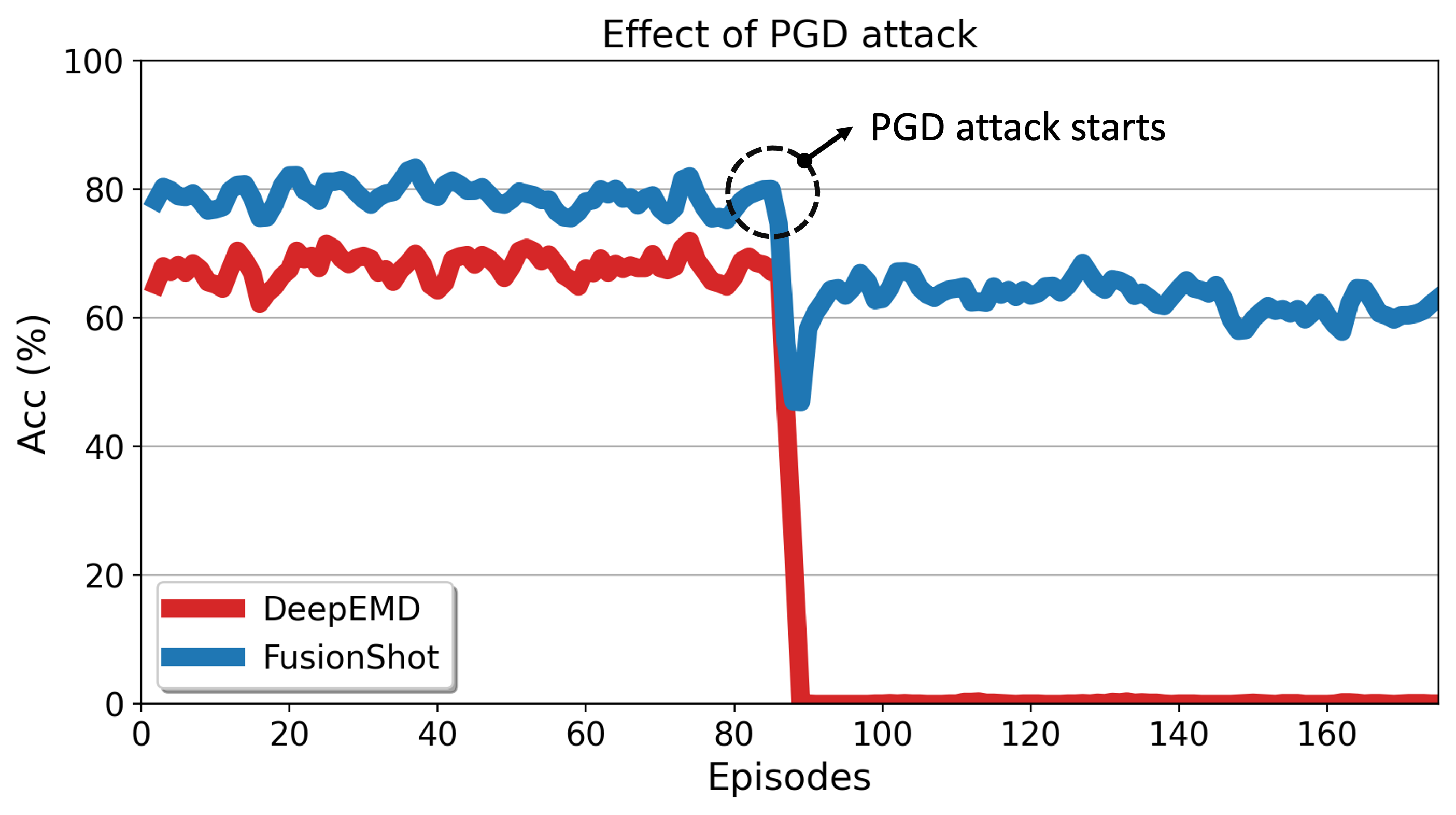}
    \end{subfigure}
    \vspace{-10pt}
    \caption{{\small 
    FusionShot under domain shift and adversarial setting.}}
    \vspace{-13pt}
   \label{fig:domainshift}
\end{figure}

\subsection{Adaptability}


\textbf{Adaptation to Domain Switches.\/} 
In addition to structural varieties, each base model in the pool can be trained using different datasets and may or may not be successful with novel images due to the class-agnostic capability of few-shot models. The models are successful when the dissimilarity between the domain of the trained dataset and that of the incoming image is not distant. For example, a model trained on a dataset containing images of cars can classify images of buses but may fail when presented with images of flowers. We tested whether FusionShot can adapt to changes in the domain by assigning the correct weight to the appropriate model based on the domain shift.

We selected 5 different datasets, ILSVRC2012 (ImageNet) \cite{deng2009imagenet}, Aircraft \cite{maji13fine-grained}, CUB, QuickDraw \cite{triantafillou2019meta}, and VGG-Flower \cite{Nilsback08}, and trained a Prototypical Network with ResNet18 backbone for each dataset individually with a total of 5 independent few-shot learners. Then, the batch of episodes each containing 2000 episodes with 5-way 5-shot settings sequentially passes into FusionShot, where 1500 of them are used for training, 300 for validation, and 200 for testing. Figure \ref{fig:domainshift} shows the performance in each incoming test set and the background color refers to which dataset the episodes are coming from. We compare the performance of FusionShot with two other baseline models, Prototypical trained using ImageNet only, and Prototypical trained using all the datasets. We make three observations: (1) Even though FusionShot experiences fluctuation in the first batches of the new dataset, it adjusts to the new domain. (2) Despite ImageNet containing 1000 classes, it is not enough to represent all the domains since Protonet\_ILSVRC suffers from the domain shift severely. (3) Compared to Protonet\_All, FusionShot outperforms in two domains and shows comparable performance in 3 domains. This shows that FusionShot can represent dataset variety with multiple models better than a single model trained on the entire dataset.

\textbf{Adaptation to Adversarial Attacks.\/} 
In this experiment, DeepEMD is used as the victim model under the PGD attack on mini-Imagenet with the 5-shot 5-way setting. Figure \ref{fig:domainshift} reports the result. The performance of FusionShot drops in the first batch where the samples are infused to attack. Then, it adjusts itself and performs around 60\%. On the other hand, without a defense mechanism, DeepEMD drops to 0.1\% after the PGD attack. We conclude that FusionShot creates a defensive barrier to mitigate the severe impact of adversarial attacks on few-shot models.

\subsection{Few-Shot Ensemble Pruning: Comparison}
\begin{table}[htbp]
  \centering
  \small
  \begin{adjustbox}{width=0.5\textwidth, center}
  \begin{tabular}{l | p{1.3cm} | p{0.7cm} p{0.7cm} p{0.7cm} p{0.7cm} p{0.7cm} }
    \hline
     \multirow{2}{1.8cm}{Ensemble Team} & Baseline & \multicolumn{5}{c}{Top-5 by Focal Diversity Pruning} \\
    \cline{2-7}
     & 123456789 & 789 & 3789 & 6789 & 36789 & 2789 \\
    \hline
    Focal Diversity Ranking & 763 & 1 & 2 & 3 & 4 & 5 \\
    Focal Diversity Score & 0.4328 & 0.4591 & 0.4575 & 0.4572 & 0.4552 & 0.4550 \\
    Kappa Diversity Ranking & 545 & 192 & 134 & 848 & 41 & 238 \\
    Kappa Score & 0.319 & 0.3055 & 0.3012 & 0.3366 & 0.2891 & 0.3082 \\
    \hline
    Simple Mean Acc (\%) & 60.52 & 62.00 & 62.57 & 61.37 & 61.81 & 62.20 \\
    Plurality Voting Acc (\%) & 65.12 & 64.82 & 64.06 & 63.51 & 63.26 & 63.24 \\
    Fusion Model Acc (\%) & \textbf{65.15} & \textbf{65.78} & \textbf{66.22} & \textbf{65.44} & \textbf{65.82} & \textbf{65.78} \\
    \hline
  \end{tabular}
  \end{adjustbox}
  \caption{{\small We compare our focal-diversity pruning mechanism with Kappa Ranking on \textit{mini}-Imagenet. We also show a comparison between baseline ensemble methods with FusionShot.}}
  \vspace{-17pt}
  \label{tab:ablation}
\end{table}

We performed an ablation study of our focal diversity-based ensemble pruning method with two cases: (i) without using any pruning method and (ii) pruning with focal diversity-based pruning and compared to the pruning using the well-known Kappa diversity metric~\cite{Cohen07}. Without pruning, the only ensemble will be the one with all base models in a given model pool. Table \ref{tab:ablation} reports the results. We make two observations. (i) The largest ensemble of 9 models will be placed at the rank position of $763^{\mathrm{th}}$ by our focal diversity. The top-1 ensemble has size 3 and outperforms the largest ensemble by 1.1\%. (ii) Comparing focal diversity-based ensemble pruning method with Kappa diversity ranking, which uses Cohen's Kappa score~\cite{Cohen07} to calculate the disagreement between two models. The low Kappa score indicates high Kappa diversity. The top-5 ensembles by focal diversity ranking (row-1) show the best-performing accuracy (last row). However, for the top-5 focal ranked ensemble teams, the ranking using their respective Kappa diversity scores (row-3) will place them at much lower rank positions. 
For example, Kappa diversity places the best-performing ensemble by our focal diversity score at the $134^{\mathrm{th}}$ rank position.

The next ablation study is on ensemble fusion by our learn-to-combine module. We compare FusionShot with ensemble fusion using a simple mean method 
and plurality voting. As shown in the 5th-6th rows of Table \ref{tab:ablation}, FusionShot learn-to-combine can outperform the two popularly used ensemble consensus methods for the top-5 ensemble teams with 3~4\% accuracy gain over simple means and 1\% to 2.5\% accuracy gain over plurality voting. Even for the largest ensemble team of 9 models, the FusionShot learn-to-combine model shows slightly better performance than plurality voting and 4.6\% over the simple means method. 

\begin{figure}[t]
    \centering
    \includegraphics[width=0.3\textwidth]{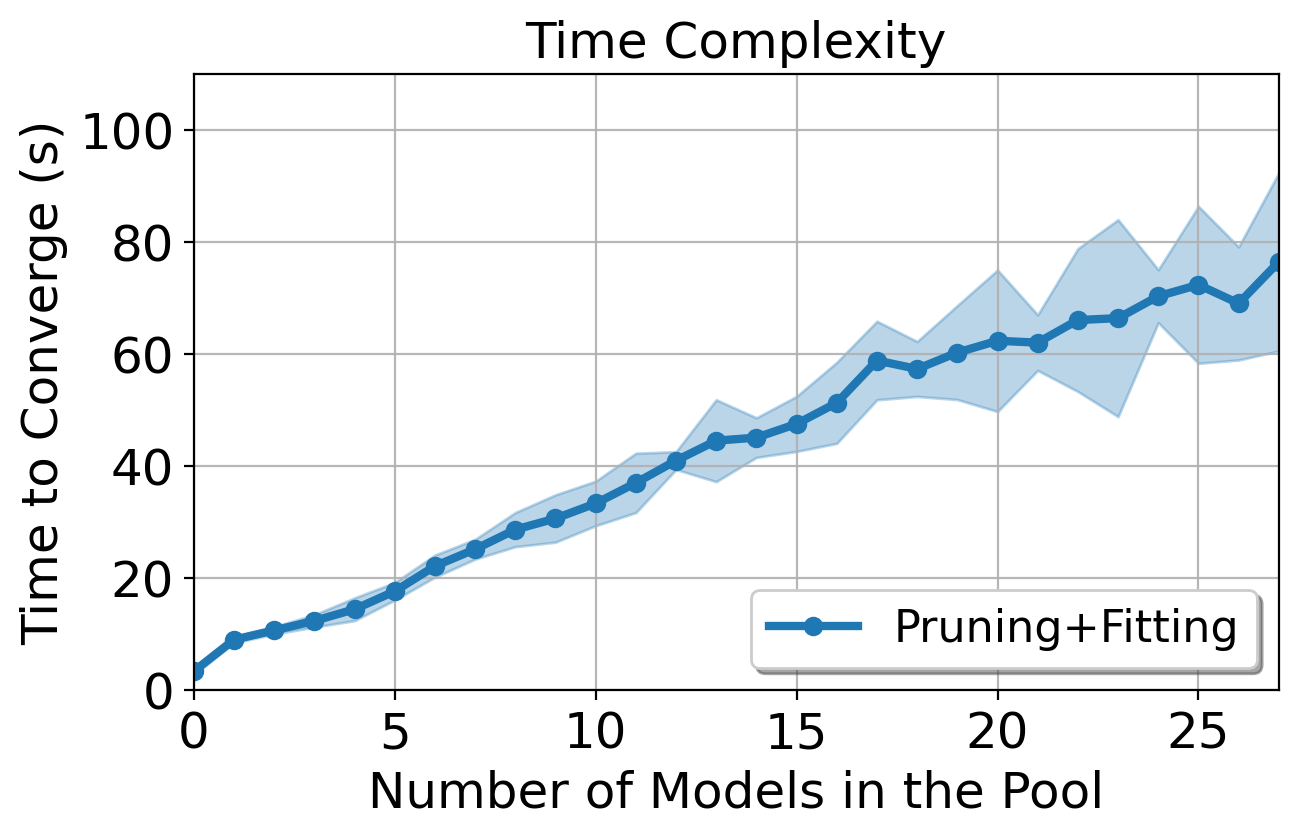}
    \vspace{-10pt}
    \caption{{\small The scaling of FusionShot as a new model being added to the base model pool.}}
    \label{fig:complexity}
\end{figure}

\vspace{-5pt}
\subsection{Complexity}
\vspace{-2pt}
Given that FusionShot only requires the logits of the base models to perform learn-to-combine, and we leverage GA algorithm to speed up the focal diversity pruning speed, FusionShot is fast and scalable to integrate a new model into the pool as shown in Figure \ref{fig:complexity}. As the new model is introduced, the input weight dimension is increased by $K \times M$ where K is the number of ways and $M$ is the number of new models. Thus, the effect of adding a new model on the complexity of the FusionShot ensemble learner is small and minimal. The dominating factor of FusionShot's time complexity is the pruning method which scales linearly (Table \ref{fig:speed} in Section~3).


\vspace{-4pt}
\section{Conclusion}
\vspace{-2pt}
We have presented {\sc FusionShot}, a focal diversity optimized few-shot ensemble method with three original contributions: (i) We explore alternative fusion channels to ensemble multiple few-shot models and leverage the learn-to-combine method to enable dynamic weight assignment to integrate multiple independent and possibly conflict predictions. (ii) We introduce a focal-diversity-optimized few-shot ensemble pruning method to further boost the performance of few-shot ensemble learning. (iii) We evaluate FusionShot with extensive experiments on mini-Imagenet and CUB benchmarks with three learning scenarios: object recognition, fine-grained image classification, and cross-domain classification and the adversarial setting. We show that the ensembles recommended by FusionShot outperform the representative SOTA few-shot models on novel tasks, even when a majority of the base models fail and the 
FusionShot defense is effective in repairing victim few-shot model under adversarial attacks.

\bibliographystyle{ACM-Reference-Format}
\bibliography{sample-base}


\begin{thebibliography}{47}


\ifx \showCODEN    \undefined \def \showCODEN     #1{\unskip}     \fi
\ifx \showDOI      \undefined \def \showDOI       #1{#1}\fi
\ifx \showISBNx    \undefined \def \showISBNx     #1{\unskip}     \fi
\ifx \showISBNxiii \undefined \def \showISBNxiii  #1{\unskip}     \fi
\ifx \showISSN     \undefined \def \showISSN      #1{\unskip}     \fi
\ifx \showLCCN     \undefined \def \showLCCN      #1{\unskip}     \fi
\ifx \shownote     \undefined \def \shownote      #1{#1}          \fi
\ifx \showarticletitle \undefined \def \showarticletitle #1{#1}   \fi
\ifx \showURL      \undefined \def \showURL       {\relax}        \fi
\providecommand\bibfield[2]{#2}
\providecommand\bibinfo[2]{#2}
\providecommand\natexlab[1]{#1}
\providecommand\showeprint[2][]{arXiv:#2}

\bibitem[Bateni et~al\mbox{.}(2020)]%
        {bateni2020improved}
\bibfield{author}{\bibinfo{person}{Peyman Bateni}, \bibinfo{person}{Raghav Goyal}, \bibinfo{person}{Vaden Masrani}, \bibinfo{person}{Frank Wood}, {and} \bibinfo{person}{Leonid Sigal}.} \bibinfo{year}{2020}\natexlab{}.
\newblock \showarticletitle{Improved few-shot visual classification}. In \bibinfo{booktitle}{\emph{Proceedings of the IEEE/CVF Conference on Computer Vision and Pattern Recognition}}. \bibinfo{pages}{14493--14502}.
\newblock


\bibitem[Bendou et~al\mbox{.}(2022)]%
        {bendou2022easy}
\bibfield{author}{\bibinfo{person}{Yassir Bendou}, \bibinfo{person}{Yuqing Hu}, \bibinfo{person}{Raphael Lafargue}, \bibinfo{person}{Giulia Lioi}, \bibinfo{person}{Bastien Pasdeloup}, \bibinfo{person}{St{\'e}phane Pateux}, {and} \bibinfo{person}{Vincent Gripon}.} \bibinfo{year}{2022}\natexlab{}.
\newblock \showarticletitle{Easy—ensemble augmented-shot-y-shaped learning: State-of-the-art few-shot classification with simple components}.
\newblock \bibinfo{journal}{\emph{Journal of Imaging}} \bibinfo{volume}{8}, \bibinfo{number}{7} (\bibinfo{year}{2022}), \bibinfo{pages}{179}.
\newblock


\bibitem[Bronskill et~al\mbox{.}(2020)]%
        {bronskill2020tasknorm}
\bibfield{author}{\bibinfo{person}{John Bronskill}, \bibinfo{person}{Jonathan Gordon}, \bibinfo{person}{James Requeima}, \bibinfo{person}{Sebastian Nowozin}, {and} \bibinfo{person}{Richard Turner}.} \bibinfo{year}{2020}\natexlab{}.
\newblock \showarticletitle{Tasknorm: Rethinking batch normalization for meta-learning}. In \bibinfo{booktitle}{\emph{International Conference on Machine Learning}}. PMLR, \bibinfo{pages}{1153--1164}.
\newblock


\bibitem[Chen et~al\mbox{.}(2019)]%
        {chen2019closer}
\bibfield{author}{\bibinfo{person}{Wei-Yu Chen}, \bibinfo{person}{Yen-Cheng Liu}, \bibinfo{person}{Zsolt Kira}, \bibinfo{person}{Yu-Chiang~Frank Wang}, {and} \bibinfo{person}{Jia-Bin Huang}.} \bibinfo{year}{2019}\natexlab{}.
\newblock \showarticletitle{A closer look at few-shot classification}.
\newblock \bibinfo{journal}{\emph{arXiv preprint arXiv:1904.04232}} (\bibinfo{year}{2019}).
\newblock


\bibitem[Cohen et~al\mbox{.}(2007)]%
        {Cohen07}
\bibfield{author}{\bibinfo{person}{Sarah Cohen}, \bibinfo{person}{Werner Nutt}, {and} \bibinfo{person}{Yehoshua Sagic}.} \bibinfo{year}{2007}\natexlab{}.
\newblock \showarticletitle{Deciding equivalances among conjunctive aggregate queries}.
\newblock \bibinfo{journal}{\emph{J. ACM}} \bibinfo{volume}{54}, \bibinfo{number}{2}, Article \bibinfo{articleno}{5} (\bibinfo{date}{April} \bibinfo{year}{2007}), \bibinfo{numpages}{50}~pages.
\newblock
\urldef\tempurl%
\url{https://doi.org/10.1145/1219092.1219093}
\showDOI{\tempurl}


\bibitem[Deng et~al\mbox{.}(2009)]%
        {deng2009imagenet}
\bibfield{author}{\bibinfo{person}{Jia Deng}, \bibinfo{person}{Wei Dong}, \bibinfo{person}{Richard Socher}, \bibinfo{person}{Li-Jia Li}, \bibinfo{person}{Kai Li}, {and} \bibinfo{person}{Li Fei-Fei}.} \bibinfo{year}{2009}\natexlab{}.
\newblock \showarticletitle{Imagenet: A large-scale hierarchical image database}. In \bibinfo{booktitle}{\emph{2009 IEEE conference on computer vision and pattern recognition}}. Ieee, \bibinfo{pages}{248--255}.
\newblock


\bibitem[Dvornik et~al\mbox{.}(2019)]%
        {dvornik2019diversity}
\bibfield{author}{\bibinfo{person}{Nikita Dvornik}, \bibinfo{person}{Cordelia Schmid}, {and} \bibinfo{person}{Julien Mairal}.} \bibinfo{year}{2019}\natexlab{}.
\newblock \showarticletitle{Diversity with cooperation: Ensemble methods for few-shot classification}. In \bibinfo{booktitle}{\emph{Proceedings of the IEEE/CVF international conference on computer vision}}. \bibinfo{pages}{3723--3731}.
\newblock


\bibitem[Dvornik et~al\mbox{.}(2020)]%
        {dvornik2020selecting}
\bibfield{author}{\bibinfo{person}{Nikita Dvornik}, \bibinfo{person}{Cordelia Schmid}, {and} \bibinfo{person}{Julien Mairal}.} \bibinfo{year}{2020}\natexlab{}.
\newblock \showarticletitle{Selecting relevant features from a multi-domain representation for few-shot classification}. In \bibinfo{booktitle}{\emph{Computer Vision--ECCV 2020: 16th European Conference, Glasgow, UK, August 23--28, 2020, Proceedings, Part X 16}}. Springer, \bibinfo{pages}{769--786}.
\newblock


\bibitem[Finn et~al\mbox{.}(2017)]%
        {finn2017model}
\bibfield{author}{\bibinfo{person}{Chelsea Finn}, \bibinfo{person}{Pieter Abbeel}, {and} \bibinfo{person}{Sergey Levine}.} \bibinfo{year}{2017}\natexlab{}.
\newblock \showarticletitle{Model-agnostic meta-learning for fast adaptation of deep networks}. In \bibinfo{booktitle}{\emph{International conference on machine learning}}. PMLR, \bibinfo{pages}{1126--1135}.
\newblock


\bibitem[Gad(2021)]%
        {gad2021pygad}
\bibfield{author}{\bibinfo{person}{Ahmed~Fawzy Gad}.} \bibinfo{year}{2021}\natexlab{}.
\newblock \bibinfo{title}{PyGAD: An Intuitive Genetic Algorithm Python Library}.
\newblock
\newblock
\showeprint[arxiv]{2106.06158}~[cs.NE]


\bibitem[Garcia and Bruna(2017)]%
        {garcia2017few}
\bibfield{author}{\bibinfo{person}{Victor Garcia} {and} \bibinfo{person}{Joan Bruna}.} \bibinfo{year}{2017}\natexlab{}.
\newblock \showarticletitle{Few-shot learning with graph neural networks}.
\newblock \bibinfo{journal}{\emph{arXiv preprint arXiv:1711.04043}} (\bibinfo{year}{2017}).
\newblock


\bibitem[Goldblum et~al\mbox{.}(2020)]%
        {goldblum2020adversarially}
\bibfield{author}{\bibinfo{person}{Micah Goldblum}, \bibinfo{person}{Liam Fowl}, {and} \bibinfo{person}{Tom Goldstein}.} \bibinfo{year}{2020}\natexlab{}.
\newblock \showarticletitle{Adversarially robust few-shot learning: A meta-learning approach}.
\newblock \bibinfo{journal}{\emph{Advances in Neural Information Processing Systems}}  \bibinfo{volume}{33} (\bibinfo{year}{2020}), \bibinfo{pages}{17886--17895}.
\newblock


\bibitem[Hiller et~al\mbox{.}(2022)]%
        {hiller2022rethinking}
\bibfield{author}{\bibinfo{person}{Markus Hiller}, \bibinfo{person}{Rongkai Ma}, \bibinfo{person}{Mehrtash Harandi}, {and} \bibinfo{person}{Tom Drummond}.} \bibinfo{year}{2022}\natexlab{}.
\newblock \showarticletitle{Rethinking generalization in few-shot classification}.
\newblock \bibinfo{journal}{\emph{Advances in Neural Information Processing Systems}}  \bibinfo{volume}{35} (\bibinfo{year}{2022}), \bibinfo{pages}{3582--3595}.
\newblock


\bibitem[Hilliard et~al\mbox{.}(2018)]%
        {hilliard2018few}
\bibfield{author}{\bibinfo{person}{Nathan Hilliard}, \bibinfo{person}{Lawrence Phillips}, \bibinfo{person}{Scott Howland}, \bibinfo{person}{Art{\"e}m Yankov}, \bibinfo{person}{Courtney~D Corley}, {and} \bibinfo{person}{Nathan~O Hodas}.} \bibinfo{year}{2018}\natexlab{}.
\newblock \showarticletitle{Few-shot learning with metric-agnostic conditional embeddings}.
\newblock \bibinfo{journal}{\emph{arXiv preprint arXiv:1802.04376}} (\bibinfo{year}{2018}).
\newblock


\bibitem[Kim et~al\mbox{.}(2019)]%
        {kim2019edge}
\bibfield{author}{\bibinfo{person}{Jongmin Kim}, \bibinfo{person}{Taesup Kim}, \bibinfo{person}{Sungwoong Kim}, {and} \bibinfo{person}{Chang~D Yoo}.} \bibinfo{year}{2019}\natexlab{}.
\newblock \showarticletitle{Edge-labeling graph neural network for few-shot learning}. In \bibinfo{booktitle}{\emph{Proceedings of the IEEE/CVF conference on computer vision and pattern recognition}}. \bibinfo{pages}{11--20}.
\newblock


\bibitem[Koch et~al\mbox{.}(2015)]%
        {koch2015siamese}
\bibfield{author}{\bibinfo{person}{Gregory Koch}, \bibinfo{person}{Richard Zemel}, \bibinfo{person}{Ruslan Salakhutdinov}, {et~al\mbox{.}}} \bibinfo{year}{2015}\natexlab{}.
\newblock \showarticletitle{Siamese neural networks for one-shot image recognition}. In \bibinfo{booktitle}{\emph{ICML deep learning workshop}}, Vol.~\bibinfo{volume}{2}. Lille.
\newblock


\bibitem[Kurakin et~al\mbox{.}(2018)]%
        {kurakin2018adversarial}
\bibfield{author}{\bibinfo{person}{Alexey Kurakin}, \bibinfo{person}{Ian~J Goodfellow}, {and} \bibinfo{person}{Samy Bengio}.} \bibinfo{year}{2018}\natexlab{}.
\newblock \showarticletitle{Adversarial examples in the physical world}.
\newblock In \bibinfo{booktitle}{\emph{Artificial intelligence safety and security}}. \bibinfo{publisher}{Chapman and Hall/CRC}, \bibinfo{pages}{99--112}.
\newblock


\bibitem[Li et~al\mbox{.}(2020)]%
        {li2020adversarial}
\bibfield{author}{\bibinfo{person}{Kai Li}, \bibinfo{person}{Yulun Zhang}, \bibinfo{person}{Kunpeng Li}, {and} \bibinfo{person}{Yun Fu}.} \bibinfo{year}{2020}\natexlab{}.
\newblock \showarticletitle{Adversarial feature hallucination networks for few-shot learning}. In \bibinfo{booktitle}{\emph{Proceedings of the IEEE/CVF conference on computer vision and pattern recognition}}. \bibinfo{pages}{13470--13479}.
\newblock


\bibitem[Li et~al\mbox{.}(2022)]%
        {li2022defensive}
\bibfield{author}{\bibinfo{person}{Wenbin Li}, \bibinfo{person}{Lei Wang}, \bibinfo{person}{Xingxing Zhang}, \bibinfo{person}{Lei Qi}, \bibinfo{person}{Jing Huo}, \bibinfo{person}{Yang Gao}, {and} \bibinfo{person}{Jiebo Luo}.} \bibinfo{year}{2022}\natexlab{}.
\newblock \showarticletitle{Defensive Few-Shot Learning}.
\newblock \bibinfo{journal}{\emph{IEEE Transactions on Pattern Analysis and Machine Intelligence}} \bibinfo{volume}{45}, \bibinfo{number}{5} (\bibinfo{year}{2022}), \bibinfo{pages}{5649--5667}.
\newblock


\bibitem[Li et~al\mbox{.}(2021)]%
        {li2021universal}
\bibfield{author}{\bibinfo{person}{Wei-Hong Li}, \bibinfo{person}{Xialei Liu}, {and} \bibinfo{person}{Hakan Bilen}.} \bibinfo{year}{2021}\natexlab{}.
\newblock \showarticletitle{Universal representation learning from multiple domains for few-shot classification}. In \bibinfo{booktitle}{\emph{Proceedings of the IEEE/CVF International Conference on Computer Vision}}. \bibinfo{pages}{9526--9535}.
\newblock


\bibitem[Liu et~al\mbox{.}(2020)]%
        {liu2020universal}
\bibfield{author}{\bibinfo{person}{Lu Liu}, \bibinfo{person}{William Hamilton}, \bibinfo{person}{Guodong Long}, \bibinfo{person}{Jing Jiang}, {and} \bibinfo{person}{Hugo Larochelle}.} \bibinfo{year}{2020}\natexlab{}.
\newblock \showarticletitle{A universal representation transformer layer for few-shot image classification}.
\newblock \bibinfo{journal}{\emph{arXiv preprint arXiv:2006.11702}} (\bibinfo{year}{2020}).
\newblock


\bibitem[Liu et~al\mbox{.}(2018)]%
        {liu2018learning}
\bibfield{author}{\bibinfo{person}{Yanbin Liu}, \bibinfo{person}{Juho Lee}, \bibinfo{person}{Minseop Park}, \bibinfo{person}{Saehoon Kim}, \bibinfo{person}{Eunho Yang}, \bibinfo{person}{Sung~Ju Hwang}, {and} \bibinfo{person}{Yi Yang}.} \bibinfo{year}{2018}\natexlab{}.
\newblock \showarticletitle{Learning to propagate labels: Transductive propagation network for few-shot learning}.
\newblock \bibinfo{journal}{\emph{arXiv preprint arXiv:1805.10002}} (\bibinfo{year}{2018}).
\newblock


\bibitem[Luo et~al\mbox{.}(2021)]%
        {luo2021rectifying}
\bibfield{author}{\bibinfo{person}{Xu Luo}, \bibinfo{person}{Longhui Wei}, \bibinfo{person}{Liangjian Wen}, \bibinfo{person}{Jinrong Yang}, \bibinfo{person}{Lingxi Xie}, \bibinfo{person}{Zenglin Xu}, {and} \bibinfo{person}{Qi Tian}.} \bibinfo{year}{2021}\natexlab{}.
\newblock \showarticletitle{Rectifying the shortcut learning of background for few-shot learning}.
\newblock \bibinfo{journal}{\emph{Advances in Neural Information Processing Systems}}  \bibinfo{volume}{34} (\bibinfo{year}{2021}), \bibinfo{pages}{13073--13085}.
\newblock


\bibitem[Ma et~al\mbox{.}(2022)]%
        {ma2022few}
\bibfield{author}{\bibinfo{person}{Chunwei Ma}, \bibinfo{person}{Ziyun Huang}, \bibinfo{person}{Mingchen Gao}, {and} \bibinfo{person}{Jinhui Xu}.} \bibinfo{year}{2022}\natexlab{}.
\newblock \showarticletitle{Few-shot learning as cluster-induced Voronoi diagrams: a geometric approach}.
\newblock \bibinfo{journal}{\emph{arXiv preprint arXiv:2202.02471}} (\bibinfo{year}{2022}).
\newblock


\bibitem[Ma et~al\mbox{.}(2021)]%
        {ma2021partner}
\bibfield{author}{\bibinfo{person}{Jiawei Ma}, \bibinfo{person}{Hanchen Xie}, \bibinfo{person}{Guangxing Han}, \bibinfo{person}{Shih-Fu Chang}, \bibinfo{person}{Aram Galstyan}, {and} \bibinfo{person}{Wael Abd-Almageed}.} \bibinfo{year}{2021}\natexlab{}.
\newblock \showarticletitle{Partner-assisted learning for few-shot image classification}. In \bibinfo{booktitle}{\emph{Proceedings of the IEEE/CVF International Conference on Computer Vision}}. \bibinfo{pages}{10573--10582}.
\newblock


\bibitem[Madry et~al\mbox{.}(2017)]%
        {madry2017towards}
\bibfield{author}{\bibinfo{person}{Aleksander Madry}, \bibinfo{person}{Aleksandar Makelov}, \bibinfo{person}{Ludwig Schmidt}, \bibinfo{person}{Dimitris Tsipras}, {and} \bibinfo{person}{Adrian Vladu}.} \bibinfo{year}{2017}\natexlab{}.
\newblock \showarticletitle{Towards deep learning models resistant to adversarial attacks}.
\newblock \bibinfo{journal}{\emph{arXiv preprint arXiv:1706.06083}} (\bibinfo{year}{2017}).
\newblock


\bibitem[Maji et~al\mbox{.}(2013)]%
        {maji13fine-grained}
\bibfield{author}{\bibinfo{person}{S. Maji}, \bibinfo{person}{J. Kannala}, \bibinfo{person}{E. Rahtu}, \bibinfo{person}{M. Blaschko}, {and} \bibinfo{person}{A. Vedaldi}.} \bibinfo{year}{2013}\natexlab{}.
\newblock \bibinfo{booktitle}{\emph{Fine-Grained Visual Classification of Aircraft}}.
\newblock \bibinfo{type}{{T}echnical {R}eport}.
\newblock
\showeprint[arxiv]{1306.5151}~[cs-cv]


\bibitem[Mangla et~al\mbox{.}(2020)]%
        {mangla2020charting}
\bibfield{author}{\bibinfo{person}{Puneet Mangla}, \bibinfo{person}{Nupur Kumari}, \bibinfo{person}{Abhishek Sinha}, \bibinfo{person}{Mayank Singh}, \bibinfo{person}{Balaji Krishnamurthy}, {and} \bibinfo{person}{Vineeth~N Balasubramanian}.} \bibinfo{year}{2020}\natexlab{}.
\newblock \showarticletitle{Charting the right manifold: Manifold mixup for few-shot learning}. In \bibinfo{booktitle}{\emph{Proceedings of the IEEE/CVF winter conference on applications of computer vision}}. \bibinfo{pages}{2218--2227}.
\newblock


\bibitem[Mirjalili and Mirjalili(2019)]%
        {mirjalili2019genetic}
\bibfield{author}{\bibinfo{person}{Seyedali Mirjalili} {and} \bibinfo{person}{Seyedali Mirjalili}.} \bibinfo{year}{2019}\natexlab{}.
\newblock \showarticletitle{Genetic algorithm}.
\newblock \bibinfo{journal}{\emph{Evolutionary Algorithms and Neural Networks: Theory and Applications}} (\bibinfo{year}{2019}), \bibinfo{pages}{43--55}.
\newblock


\bibitem[Nilsback and Zisserman(2008)]%
        {Nilsback08}
\bibfield{author}{\bibinfo{person}{M-E. Nilsback} {and} \bibinfo{person}{A. Zisserman}.} \bibinfo{year}{2008}\natexlab{}.
\newblock \showarticletitle{Automated Flower Classification over a Large Number of Classes}. In \bibinfo{booktitle}{\emph{Proceedings of the Indian Conference on Computer Vision, Graphics and Image Processing}}.
\newblock


\bibitem[Oreshkin et~al\mbox{.}(2018)]%
        {oreshkin2018tadam}
\bibfield{author}{\bibinfo{person}{Boris Oreshkin}, \bibinfo{person}{Pau Rodr{\'\i}guez~L{\'o}pez}, {and} \bibinfo{person}{Alexandre Lacoste}.} \bibinfo{year}{2018}\natexlab{}.
\newblock \showarticletitle{Tadam: Task dependent adaptive metric for improved few-shot learning}.
\newblock \bibinfo{journal}{\emph{Advances in neural information processing systems}}  \bibinfo{volume}{31} (\bibinfo{year}{2018}).
\newblock


\bibitem[Radford et~al\mbox{.}(2021)]%
        {radford2021learning}
\bibfield{author}{\bibinfo{person}{Alec Radford}, \bibinfo{person}{Jong~Wook Kim}, \bibinfo{person}{Chris Hallacy}, \bibinfo{person}{Aditya Ramesh}, \bibinfo{person}{Gabriel Goh}, \bibinfo{person}{Sandhini Agarwal}, \bibinfo{person}{Girish Sastry}, \bibinfo{person}{Amanda Askell}, \bibinfo{person}{Pamela Mishkin}, \bibinfo{person}{Jack Clark}, {et~al\mbox{.}}} \bibinfo{year}{2021}\natexlab{}.
\newblock \showarticletitle{Learning transferable visual models from natural language supervision}. In \bibinfo{booktitle}{\emph{International conference on machine learning}}. PMLR, \bibinfo{pages}{8748--8763}.
\newblock


\bibitem[Ravi and Larochelle(2016)]%
        {ravi2016optimization}
\bibfield{author}{\bibinfo{person}{Sachin Ravi} {and} \bibinfo{person}{Hugo Larochelle}.} \bibinfo{year}{2016}\natexlab{}.
\newblock \showarticletitle{Optimization as a model for few-shot learning}. In \bibinfo{booktitle}{\emph{International conference on learning representations}}.
\newblock


\bibitem[Rusu et~al\mbox{.}(2018)]%
        {rusu2018meta}
\bibfield{author}{\bibinfo{person}{Andrei~A Rusu}, \bibinfo{person}{Dushyant Rao}, \bibinfo{person}{Jakub Sygnowski}, \bibinfo{person}{Oriol Vinyals}, \bibinfo{person}{Razvan Pascanu}, \bibinfo{person}{Simon Osindero}, {and} \bibinfo{person}{Raia Hadsell}.} \bibinfo{year}{2018}\natexlab{}.
\newblock \showarticletitle{Meta-learning with latent embedding optimization}.
\newblock \bibinfo{journal}{\emph{arXiv preprint arXiv:1807.05960}} (\bibinfo{year}{2018}).
\newblock


\bibitem[Snell et~al\mbox{.}(2017)]%
        {snell2017prototypical}
\bibfield{author}{\bibinfo{person}{Jake Snell}, \bibinfo{person}{Kevin Swersky}, {and} \bibinfo{person}{Richard Zemel}.} \bibinfo{year}{2017}\natexlab{}.
\newblock \showarticletitle{Prototypical networks for few-shot learning}.
\newblock \bibinfo{journal}{\emph{Advances in neural information processing systems}}  \bibinfo{volume}{30} (\bibinfo{year}{2017}).
\newblock


\bibitem[Sung et~al\mbox{.}(2018)]%
        {sung2018learning}
\bibfield{author}{\bibinfo{person}{Flood Sung}, \bibinfo{person}{Yongxin Yang}, \bibinfo{person}{Li Zhang}, \bibinfo{person}{Tao Xiang}, \bibinfo{person}{Philip~HS Torr}, {and} \bibinfo{person}{Timothy~M Hospedales}.} \bibinfo{year}{2018}\natexlab{}.
\newblock \showarticletitle{Learning to compare: Relation network for few-shot learning}. In \bibinfo{booktitle}{\emph{Proceedings of the IEEE conference on computer vision and pattern recognition}}. \bibinfo{pages}{1199--1208}.
\newblock


\bibitem[Tian et~al\mbox{.}(2020)]%
        {tian2020rethinking}
\bibfield{author}{\bibinfo{person}{Yonglong Tian}, \bibinfo{person}{Yue Wang}, \bibinfo{person}{Dilip Krishnan}, \bibinfo{person}{Joshua~B Tenenbaum}, {and} \bibinfo{person}{Phillip Isola}.} \bibinfo{year}{2020}\natexlab{}.
\newblock \showarticletitle{Rethinking few-shot image classification: a good embedding is all you need?}. In \bibinfo{booktitle}{\emph{Computer Vision--ECCV 2020: 16th European Conference, Glasgow, UK, August 23--28, 2020, Proceedings, Part XIV 16}}. Springer, \bibinfo{pages}{266--282}.
\newblock


\bibitem[Triantafillou et~al\mbox{.}(2021)]%
        {triantafillou2021learning}
\bibfield{author}{\bibinfo{person}{Eleni Triantafillou}, \bibinfo{person}{Hugo Larochelle}, \bibinfo{person}{Richard Zemel}, {and} \bibinfo{person}{Vincent Dumoulin}.} \bibinfo{year}{2021}\natexlab{}.
\newblock \showarticletitle{Learning a universal template for few-shot dataset generalization}. In \bibinfo{booktitle}{\emph{International Conference on Machine Learning}}. PMLR, \bibinfo{pages}{10424--10433}.
\newblock


\bibitem[Triantafillou et~al\mbox{.}(2019)]%
        {triantafillou2019meta}
\bibfield{author}{\bibinfo{person}{Eleni Triantafillou}, \bibinfo{person}{Tyler Zhu}, \bibinfo{person}{Vincent Dumoulin}, \bibinfo{person}{Pascal Lamblin}, \bibinfo{person}{Utku Evci}, \bibinfo{person}{Kelvin Xu}, \bibinfo{person}{Ross Goroshin}, \bibinfo{person}{Carles Gelada}, \bibinfo{person}{Kevin Swersky}, \bibinfo{person}{Pierre-Antoine Manzagol}, {et~al\mbox{.}}} \bibinfo{year}{2019}\natexlab{}.
\newblock \showarticletitle{Meta-dataset: A dataset of datasets for learning to learn from few examples}.
\newblock \bibinfo{journal}{\emph{arXiv preprint arXiv:1903.03096}} (\bibinfo{year}{2019}).
\newblock


\bibitem[Vinyals et~al\mbox{.}(2016)]%
        {vinyals2016matching}
\bibfield{author}{\bibinfo{person}{Oriol Vinyals}, \bibinfo{person}{Charles Blundell}, \bibinfo{person}{Timothy Lillicrap}, \bibinfo{person}{Daan Wierstra}, {et~al\mbox{.}}} \bibinfo{year}{2016}\natexlab{}.
\newblock \showarticletitle{Matching networks for one shot learning}.
\newblock \bibinfo{journal}{\emph{Advances in neural information processing systems}}  \bibinfo{volume}{29} (\bibinfo{year}{2016}).
\newblock


\bibitem[Wang et~al\mbox{.}(2019)]%
        {wang2019simpleshot}
\bibfield{author}{\bibinfo{person}{Yan Wang}, \bibinfo{person}{Wei-Lun Chao}, \bibinfo{person}{Kilian~Q Weinberger}, {and} \bibinfo{person}{Laurens van~der Maaten}.} \bibinfo{year}{2019}\natexlab{}.
\newblock \showarticletitle{Simpleshot: Revisiting nearest-neighbor classification for few-shot learning}.
\newblock \bibinfo{journal}{\emph{arXiv preprint arXiv:1911.04623}} (\bibinfo{year}{2019}).
\newblock


\bibitem[Wei and Liu(2021)]%
        {wenqiWei-TDSC}
\bibfield{author}{\bibinfo{person}{Wenqi Wei} {and} \bibinfo{person}{Ling Liu}.} \bibinfo{year}{2021}\natexlab{}.
\newblock \showarticletitle{Robust Deep Learning Ensemble against Deception}.
\newblock \bibinfo{journal}{\emph{IEEE Transaction on Dependable and Secure Computing (TDSC)}} (\bibinfo{year}{2021}).
\newblock


\bibitem[Wu et~al\mbox{.}(2021)]%
        {wu2021boosting}
\bibfield{author}{\bibinfo{person}{Yanzhao Wu}, \bibinfo{person}{Ling Liu}, \bibinfo{person}{Zhongwei Xie}, \bibinfo{person}{Ka-Ho Chow}, {and} \bibinfo{person}{Wenqi Wei}.} \bibinfo{year}{2021}\natexlab{}.
\newblock \showarticletitle{Boosting ensemble accuracy by revisiting ensemble diversity metrics}. In \bibinfo{booktitle}{\emph{Proceedings of the IEEE/CVF Conference on Computer Vision and Pattern Recognition}}. \bibinfo{pages}{16469--16477}.
\newblock


\bibitem[Xie et~al\mbox{.}(2022)]%
        {xie2022joint}
\bibfield{author}{\bibinfo{person}{Jiangtao Xie}, \bibinfo{person}{Fei Long}, \bibinfo{person}{Jiaming Lv}, \bibinfo{person}{Qilong Wang}, {and} \bibinfo{person}{Peihua Li}.} \bibinfo{year}{2022}\natexlab{}.
\newblock \showarticletitle{Joint distribution matters: Deep brownian distance covariance for few-shot classification}. In \bibinfo{booktitle}{\emph{Proceedings of the IEEE/CVF conference on computer vision and pattern recognition}}. \bibinfo{pages}{7972--7981}.
\newblock


\bibitem[Yang et~al\mbox{.}(2021)]%
        {yang2021free}
\bibfield{author}{\bibinfo{person}{Shuo Yang}, \bibinfo{person}{Lu Liu}, {and} \bibinfo{person}{Min Xu}.} \bibinfo{year}{2021}\natexlab{}.
\newblock \showarticletitle{Free lunch for few-shot learning: Distribution calibration}.
\newblock \bibinfo{journal}{\emph{arXiv preprint arXiv:2101.06395}} (\bibinfo{year}{2021}).
\newblock


\bibitem[Ye et~al\mbox{.}(2020)]%
        {ye2020few}
\bibfield{author}{\bibinfo{person}{Han-Jia Ye}, \bibinfo{person}{Hexiang Hu}, \bibinfo{person}{De-Chuan Zhan}, {and} \bibinfo{person}{Fei Sha}.} \bibinfo{year}{2020}\natexlab{}.
\newblock \showarticletitle{Few-shot learning via embedding adaptation with set-to-set functions}. In \bibinfo{booktitle}{\emph{Proceedings of the IEEE/CVF conference on computer vision and pattern recognition}}. \bibinfo{pages}{8808--8817}.
\newblock


\bibitem[Zhang et~al\mbox{.}(2020)]%
        {zhang2020deepemd}
\bibfield{author}{\bibinfo{person}{Chi Zhang}, \bibinfo{person}{Yujun Cai}, \bibinfo{person}{Guosheng Lin}, {and} \bibinfo{person}{Chunhua Shen}.} \bibinfo{year}{2020}\natexlab{}.
\newblock \showarticletitle{Deepemd: Few-shot image classification with differentiable earth mover's distance and structured classifiers}. In \bibinfo{booktitle}{\emph{Proceedings of the IEEE/CVF conference on computer vision and pattern recognition}}. \bibinfo{pages}{12203--12213}.
\newblock


\end{thebibliography}

\appendix

\section{Reproducibility Statement}
We make the following effort to enhance the reproducibility of our results. 
\begin{itemize}
    \item For {\sc FusionShot} implementation, a link to an anonymous downloadable source is included in our abstract.
    \item We show a brief description of the representative SOTA few-shot models in Appendix B.1 to B.4. Detailed settings and the hyper-parameter selection logistics can be found in Appendix B.5.
\end{itemize}

\section{Experiment Details}

\begin{figure*}[hbt!]
    \centering
    \includegraphics[width=0.7\textwidth]{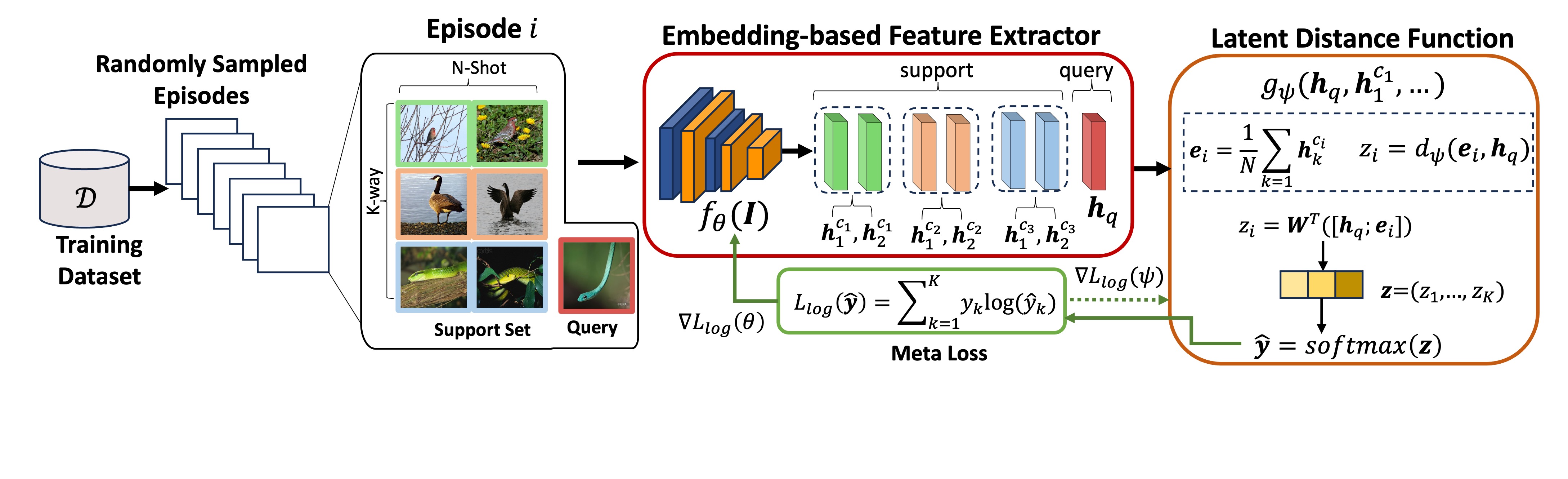}
    \vspace{-6pt}
    \caption{A Reference Framework for Few Shot Learning}
    \label{fig:episode}
\end{figure*}

\subsection{Base Model Details}

\begin{table*}[hbt!]
    \begin{subtable}{\linewidth}
    \small
    \centering
        \begin{tabular}{c c c p{4cm} c }
            \hline \\
            Method & \#Train Episodes & Input Dim. & Train Aug. & \\
            \hline
            protonet & 60,000 & $3\times 224\times 224$ & Resize, CenterCrop, Normalize \\
            MatchinNet & 60,000 & $3\times 224\times 224$ & Resize, CenterCrop, Normalize \\
            RelationNet & 60,000 & $3\times 224\times 224$ & Resize, CenterCrop, Normalize\\
            MAML & 60,000 & $3\times 224\times 224$ & Resize, CenterCrop, Normalize\\
            DeepEMD & 5,000 & $3\times 84\times 84$ & RandomResizedCrop, RandomHorizontalFlip, Normalize  \\
    SimpleShot & 9,000 & $3\times 224\times 224$ & Resize, CenterCrop, Normalize\\        
            \hline
        \end{tabular}
    \caption{}
    \label{table:hyperparams_a}
    \end{subtable}
    \begin{subtable}{\linewidth}
        \centering
        \begin{tabular}{lllll}
        \hline
        Method & Distance Function & Pre-Train Loss & Meta-Loss Function & Optimizer \\
        \hline
        protonet & Euclidean & - & Cross Entropy & Adam \\
        matchingnet & Cosine & - & Cross Entropy & Adam \\
        relationnet & CNN & - & MSE & Adam \\
        maml & MLP & - & Cross Entropy & Adam \\
        deepemd & EMD & Cross Entropy & Cross Entropy & SGD \\
        simpleshot & KNN (Cosine) & Cross Entropy & - & SGD \\
        \hline
        \end{tabular}
        \caption{}
        \label{table:hyperparams_b}
    \end{subtable}
    \begin{subtable}{\linewidth}
    \centering
    \small
        \begin{tabular}{c c c c c c c c }
            \hline
            Method & Backbone & Embed. Dim. & Pretrain & Episode Time & Total Size \\
            \hline
            Protonet & ResNet18 &  512 & No & 29ms & 42.672 MB \\
            MatchinNet & ResNet18 & 512 & No & 60ms & 70.719 MB \\
            RelationNet & ResNet18 &  512 & No & 28ms & 69.707 MB \\
            MAML & ResNet18 &  512 & No & 161ms & 42.682 MB \\
            DeepEMD & ResNet18 &  512 & Yes & 354ms & 47.431 MB\\
            SimpleShot & ResNet18 &  512 & Yes & 94ms & 42.768 MB \\  
            \hline
        \end{tabular}
    \caption{}
    \label{table:hyperparams_c}
    \end{subtable}
    \caption{We show the training setting of each base model in (a) and (b). The cost of each model in terms of spatial and temporal is shown in (c)}
    \label{table:hyperparams}
\end{table*}

\begin{figure*}[t]
    \centering
    \includegraphics[width=0.8\textwidth]{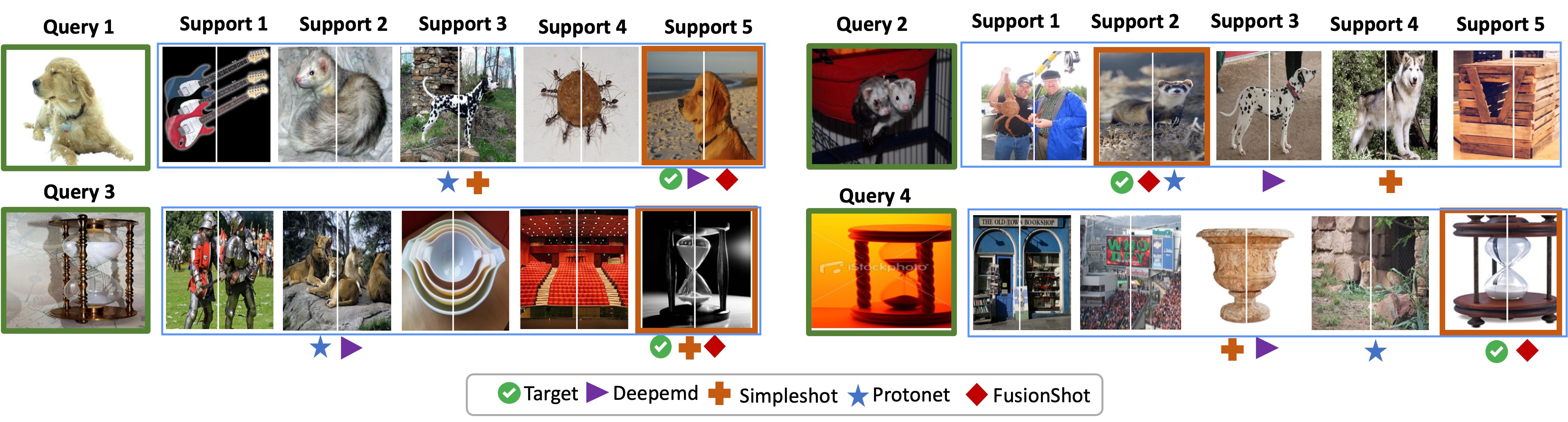}
    \caption{{\small We compare four few-shot models: ProtoNet, DeepEMD, SimpleShot, and FusionShot. For $\mathrm{Query}$-1, DeepEMD and FusionShot succeed in top-1 prediction while SimpleShot and ProtoNet both fail. However, DeepEMD fails on $\mathrm{Query}$-2, whereas all others succeed with the correct top-1 result, and  $\mathrm{Query}$-3, whereas ProtoNet also fails, but SimpleShot and FusionShot succeed. $\mathrm{Query}$-4 is a novel test example where DeepEMD, SimpleShot, and ProtoNet all fail, but FusionShot succeeds in finding the correct top-1 result by integrating focal diversity optimized ensemble pruning and learn-to-combine.}} 
    \label{fig:queryexamples}
\end{figure*}

In this section, we give details on our models, where each of them utilizes a different latent distance function. For all the extracted query and support embeddings we use the notation of $f_{\theta}(\mathbf{Q}) = \mathbf{h}_q$ and $f_{\theta}(\mathbf{I}_i)=\mathbf{h}_i$, where $\mathbf{I}_i \in \mathcal{S}$ and $\theta$ is the backbone parameters. On below we show that $\hat{y}_{k}$ is the $k^{th}$ value of probability vector $\hat{\mathbf{y}}_k$.

\textbf{Prototypical Networks.\/}
After the backbone architecture produces each embedding, Prototypical Networks take the average of embeddings that are in the same class and call them class prototypes. Then, they classify the query by looking at the nearest Euclidian distance from the query embedding to the class prototypes. When all the distances are calculated, a probability value is calculated for each class as follows:
\begin{equation}
    p(\hat{y}_k|\mathbf{Q}, \mathcal{S}) = \frac{exp(-||\mathbf{h}_q - \mathbf{e}_k||_{2}^{2})}{\sum_{k'} exp(-||\mathbf{h}_q - \mathbf{e}_{k^{'}}||_{2}^{2})},
\end{equation}
where we use the same notation $\mathbf{e}_k$ for class prototypes. Prototypical Networks suffer the cross-entropy loss.

\textbf{Matching Networks}, however, in the simplest form calculate softmax over the cosine distances between the query and all the support embeddings, 
\begin{equation}
\alpha(\mathbf{h}_q, \mathbf{h}_i) = \frac{exp(cos(\mathbf{h}_q, \mathbf{h}_i)))}{\sum_j^{|\mathcal{S}|} exp(cos(\mathbf{h}_q, \mathbf{h}_j))}    
\end{equation}
 
which they call an attention value to the class. Then, Matching Networks perform a linear combination of the support labels:
\begin{equation}
    p(\hat{y}_k|\mathbf{Q}, \mathcal{S}) = \sum_{i=1}^{|\mathcal{S}|}\alpha(\mathbf{h}_q, \mathbf{h}_i)\mathbf{1}({y_{i} = k}),
\end{equation}
where $\mathbf{1}$ is the identity function which takes 1 corresponding to the calculated class probability. Matching Networks, also, suffer the cross-entropy loss.

\textbf{Relation Networks} utilize another CNN architecture called {\sc relation module} to perform a comparison between the query and support embeddings. The relation module takes the concatenated query embedding and a support embedding to produce a relation score $r$ representing the relation between the query and the sample. In the case of multiple shots, the relation networks sum all the relation scores for individual classes. Differently, the relation networks suffer mean squared error loss where the matched pairs have similarity 1 and mismatched pairs have similarity 0.

\textbf{MAML} is also a parametric model that employs a linear layer with parameters $\mathbf{W}$ and $\mathbf{b}$ on top of the backbone $f_{\theta}(.)$ following a softmax operation. Differently, in each episode, it performs a small number of learning steps on the given support set, starting from the initial parameters ($\mathbf{W}$, $\mathbf{b}$, $\theta$). The inner loop performs supervised learning by grouping the images in the same class and labeling them. After the inner loop iteration, the model predicts the query as follows:
\begin{equation}
    p(\hat{y}_{k} | \mathbf{Q}, \mathcal{S}) = \mathrm{softmax}(\mathbf{b}' + \mathbf{W}'f_{\theta'}(\mathbf{Q})),
\end{equation}
where ($\mathbf{W}'$, $\mathbf{b}'$, $\theta'$) are the updated parameters by the inner loop by suffering the cross-entropy loss. Normally, the model is trained by updating the second-order gradients from inner loop parameters into the initial parameters but first-order approximation is taken to reduce the amount of memory cost. Dissimilar to other methods, MAML aims to learn the best initialization parameters.

\textbf{Simple Shot} does not perform meta-training. Simpleshot, first, trains a classifier on top of the backbone using supervised labels by minimizing the cross-entropy loss, ${l}(\mathbf{W}f_{\theta}(\mathbf{I}), y)$. Secondly, it removes the classifier and performs meta-testing by assigning the closest support embedding to the query embedding, i.e., it performs nearest-neighbor classification:
\begin{equation}
    \hat{y}_{k} = \argmin_{c_{i} \in \{c_1, \dots, c_K\}}d(\mathbf{h}_q, \mathbf{h}^{c_i})
\end{equation}
where $d$ either can be Euclidian or cosine distance. Simpleshot provides simple transformations on the embeddings by centering and L2 normalization.

\textbf{DeepEMD}, first, pre-trains its backbone architecture by following the process we show in Appendix B.3. Second, it removes the classifier layer and performs meta-training. Deep EMD employs Earth Mover's Distance function as their distance metric to the extracted features of query and support samples:
\begin{equation}
    \mathrm{EMD}(\mathbf{Q}, \mathbf{I}) =  \sum_{i,j}\Tilde{x}_{i,j}\zeta_{i,j},
\end{equation}
where $\mathbf{I}\in\mathcal{S}$ and $\Tilde{x}$ is the maximum flow of sending query weights to support weights and $\zeta$ is the cost between weights. DeepEMD calculates the distance by solving this Linear Problem (LP). Thus, it calls an LP solver in each iteration. To perform end-to-end training on the backbone, it calculates the Jacobian matrix on the solution of the LP. After propagating the gradients coming from the distance function, it suffers cross-entropy loss.


\subsection{Base Model Implementation Details}
In the implementation of the Protonet, Matchingnet, Relationnet, and MAML; we used the code provided by \citep{chen2019closer}. For the DeepEMD and Simpleshot, we used the code provided by \citep{zhang2020deepemd} and \citep{wang2019simpleshot}, respectively. DeepEMD contains three different feature extractor types, fully connected, grid, and sampling, we use the first type to be fair with the other methods since the other types of DeepEMD perform patching on the query image whereas Fully Connected takes plain images. We separately train each base model with the hyperparameters shown in Table \ref{table:hyperparams_a} and \ref{table:hyperparams_b}. For all the hyperparameters we suggest checking our library.

In terms of spatial and temporal cost, we provide the capacity of each model with a ResNet18 backbone, in Table \ref{table:hyperparams_c}. Note that, ResNet18 has a size of 42.67MB alone. Secondly, we provided the duration of one forward pass for each model. In an online set-up where each base model runs in parallel, the bottle-neck model will be DeepEMD. 

In each epoch, we perform multiple iterations of episode creation and forward pass of the created episode. The classes, $K$, are selected randomly from the $\mathcal{C}^{\mathrm{train}}$. Then, we sample multiple images to create queries and a support set from the corresponding samples of the classes. Note that, the sampled data is in $K\times (N+M)$ shape. We take the first $N$ columns as support set, and the other columns as queries. Thus, in each iteration, we have $M$ amount of query for each class. One forward pass on the model produces the logits, and we compare the logits with the class IDs. The class IDs, $y$, are integers showing the positions of the classes in the support set for each query. Note that, $y$ is the same for each iteration regardless of the class sampled from $\mathcal{C}^{\mathrm{train}}$. Lastly, we compare the logits with the $y$ and obtain episode accuracy and loss.

The episode creation process is the same as the training algorithm with the difference of selected classes for the samples, i.e., $\mathcal{C}^{\mathrm{train}}$, $\mathcal{C}^{\mathrm{val}}$, and $\mathcal{C}^{\mathrm{novel}}$. Since $y$ is the same for each iteration regardless of which class it is sampled from, the process is class invariant, which makes the models few-shot learners. 

\subsection{ Pretraining before Few-shot Meta-training\/}
Pretraining is one of the options to further improve the transferability and adaptability of few-shot learning to unseen novel data. 
Instead of beginning with randomly initialized weights, this optimization will choose to pre-train the DNN embedding-based feature extractor $f_{\theta}(.)$ on the training set by adding a supervised classifier layer. Let $\tilde{y}_i = \mathrm{softmax}(\mathbf{W}^{T}f_{\theta}(\mathbf{I}))$ represent the predictions based on the jointly trained model. Here $\mathbf{I}\in\mathcal{D}^{\mathrm{train}}$ representing an image sampled from the training dataset and $\mathbf{W}^T \in \mathbb{R}^{m}\times z$ is the classifier layer, where $m$ denotes the embedding dimensions of the feature extractor and $z$ denotes the number of classes in $\mathcal{D}^{train}$. This approach  jointly trains the embedding with the cross-entropy loss to minimize the classification error:
\begin{equation}
    \mathcal{L}_{cross} (\mathbf{\theta}, \mathbf{W}) = \argmin_{\theta, \mathbf{W}} - \sum_{\mathbf{I}\sim \mathcal{D}^{\mathrm{train}}}\tilde{y}\log (p_{\theta, \mathbf{W}}(\tilde{y}\mid\mathbf{I}))
\end{equation}
Contrary to the few-shot learning, here we perform standard supervised training where the classifier layer makes predictions among $z$ classes. To further improve the adaption, during the training process, the model evaluated on few-shot learning tasks using $\mathcal{D}^{val}$. However, this technique varies from one method to another.

\subsection{Base Model Performances on Novel Classes}
\begin{figure}[hbt!]
\centering
    \begin{subfigure}{0.35\textwidth}
        \centering
        \includegraphics[width=\textwidth]{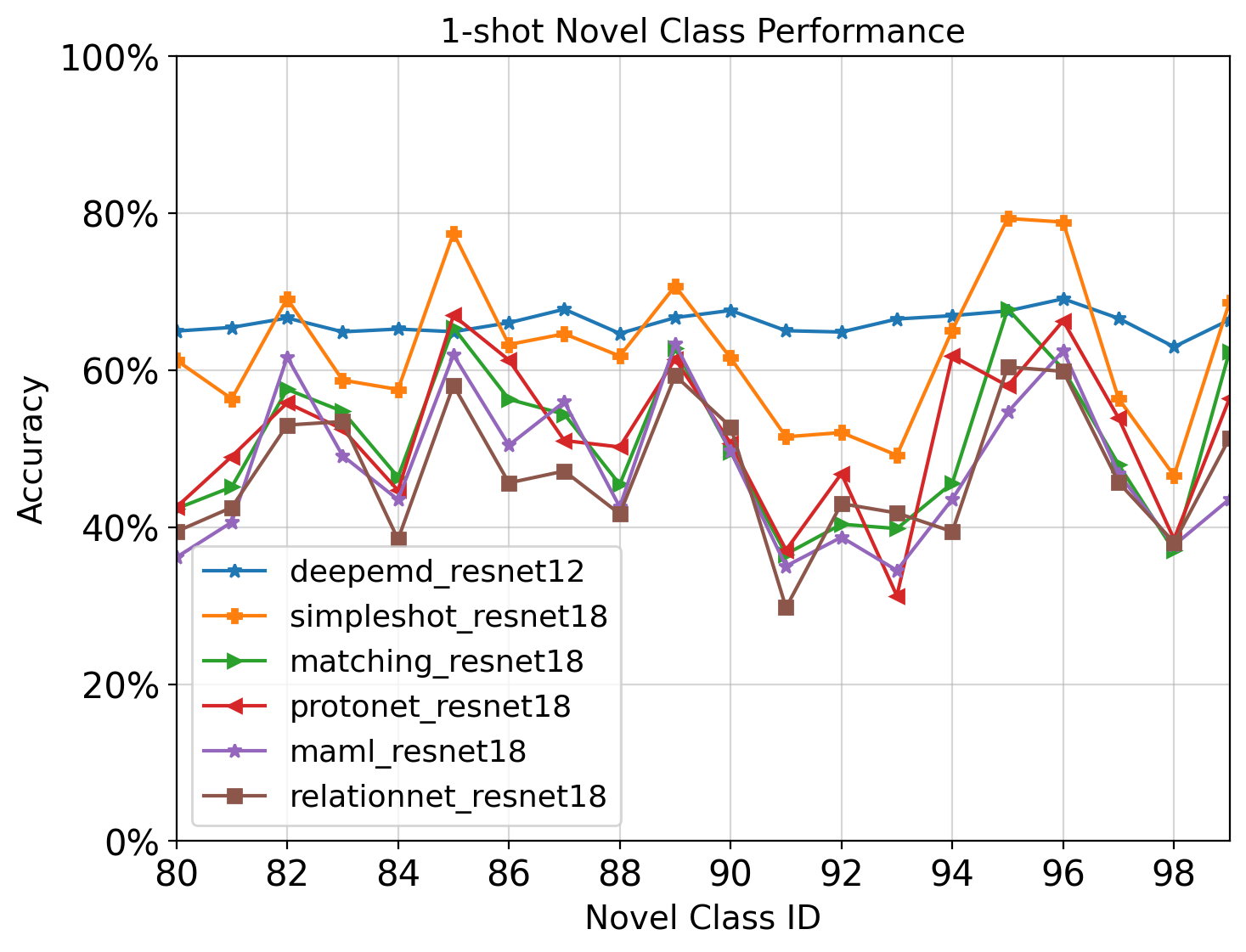}
    \end{subfigure}
    \begin{subfigure}{0.35\textwidth}
        \centering
        \includegraphics[width=\textwidth]{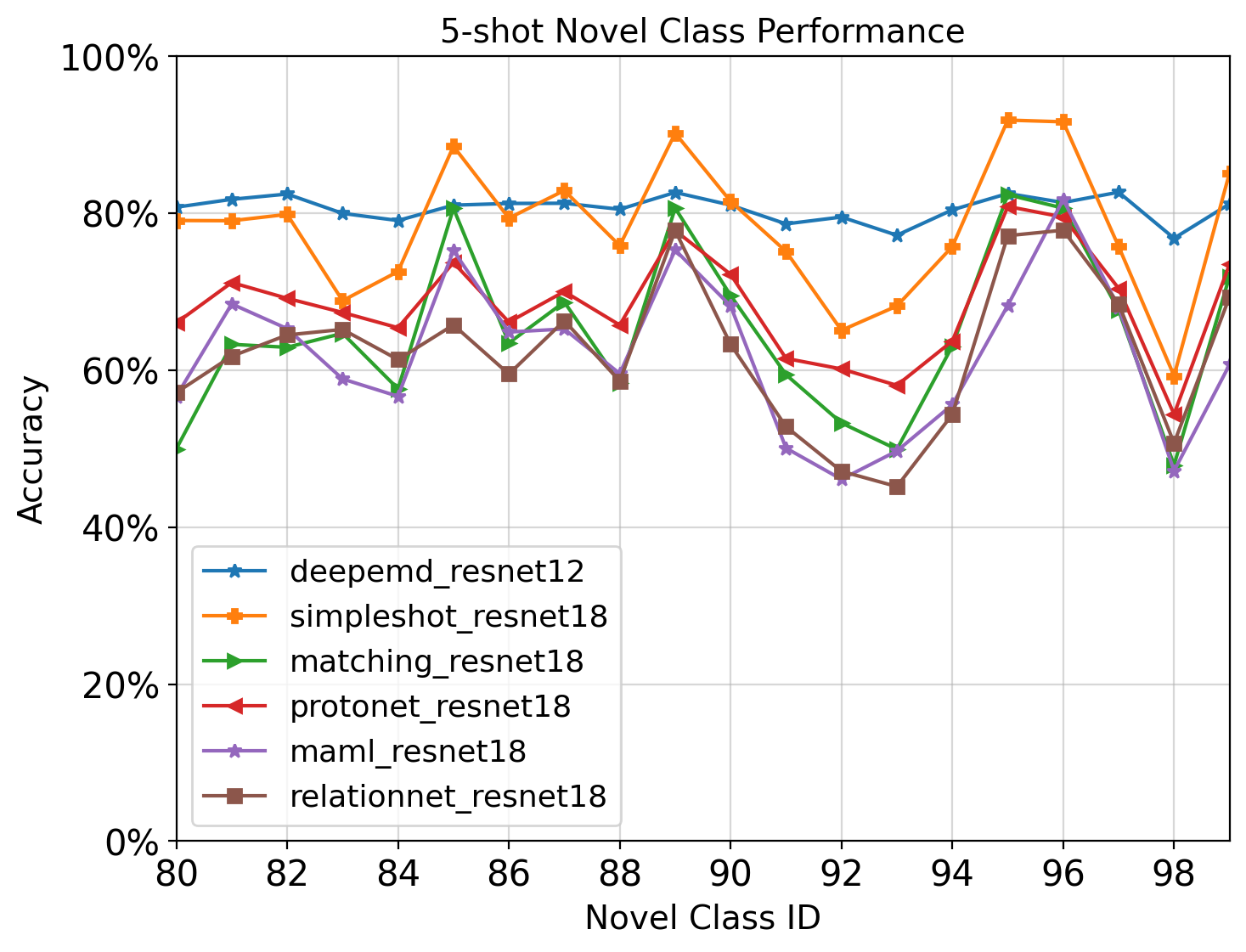}
    \end{subfigure}
    \caption{We show the performance of each base model on novel classes of \textit{mini}-Imagenet, where each model is trained on \textit{mini}-Imagenet for 1-shot-5way(a) and 5-shot-5way}
    \label{fig:novel_class_perf_base}
\end{figure}

We analyze and compare the performance of the base models at the novel classes of mini-Imagenet in Figure  \ref{fig:novel_class_perf_base}, where the models use the same backbone architecture type, ResNet18. We make two interesting observations. First, by observing the performance increase and decrease, we can say that most of the base models follow similar patterns over the novel classes. However, the variance of the models' performances in a particular class is high. Second, DeepEMD shows more stable and overall better top-1 and top-5 performance across different novel classes. However, for some novel classes, Simpleshot clearly outperforms DeepEMD, such as novel class IDs 85, 89, 94, and 96 for both top-1 and top-5 performance.

\begin{figure*}[hbt!]
\centering
    \begin{subfigure}{0.31\textwidth}
        \centering
        \includegraphics[width=\textwidth]{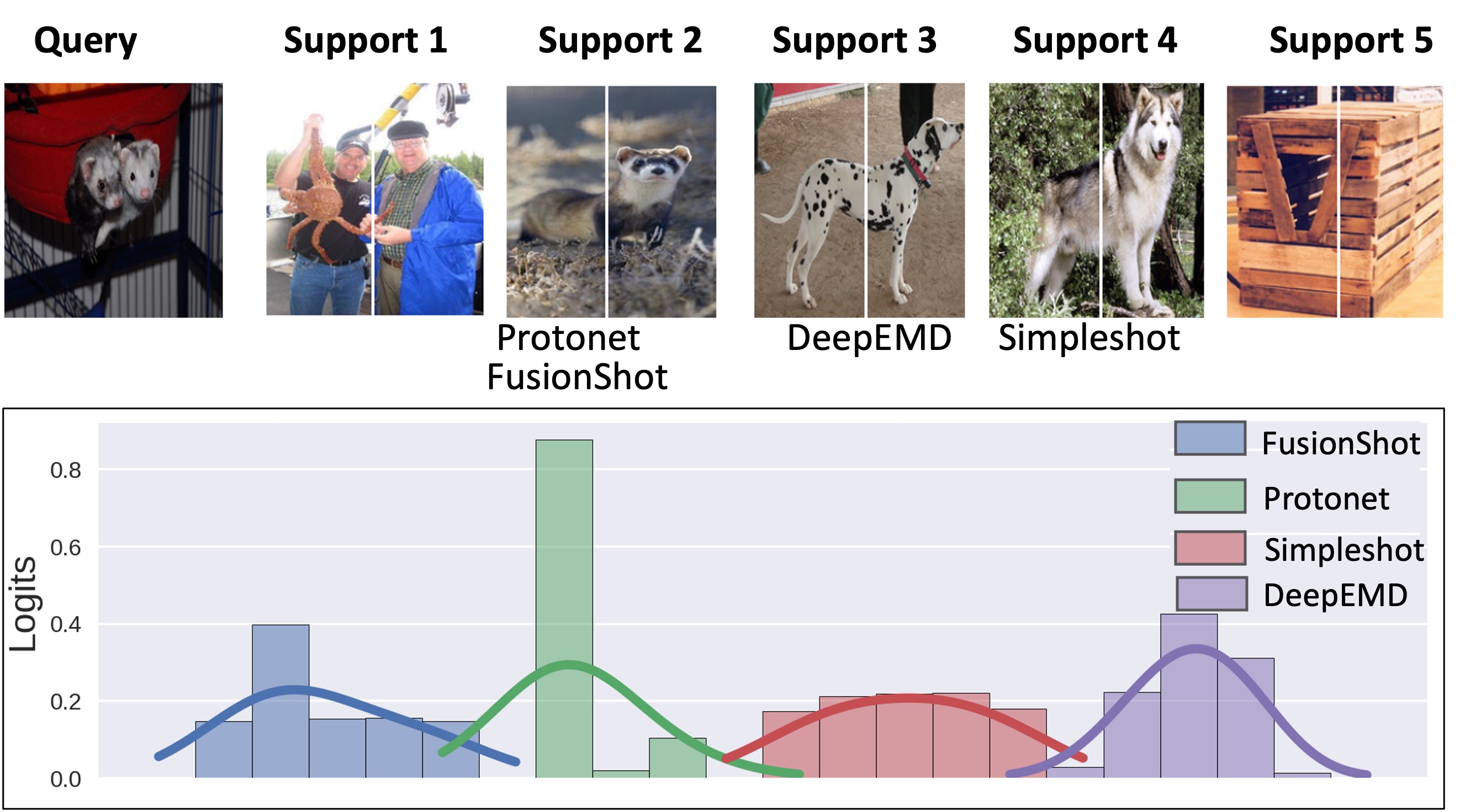}
        \label{fig:obs_density_a}
    \end{subfigure}
    \begin{subfigure}{0.31\textwidth}
        \centering
        \includegraphics[width=\textwidth]{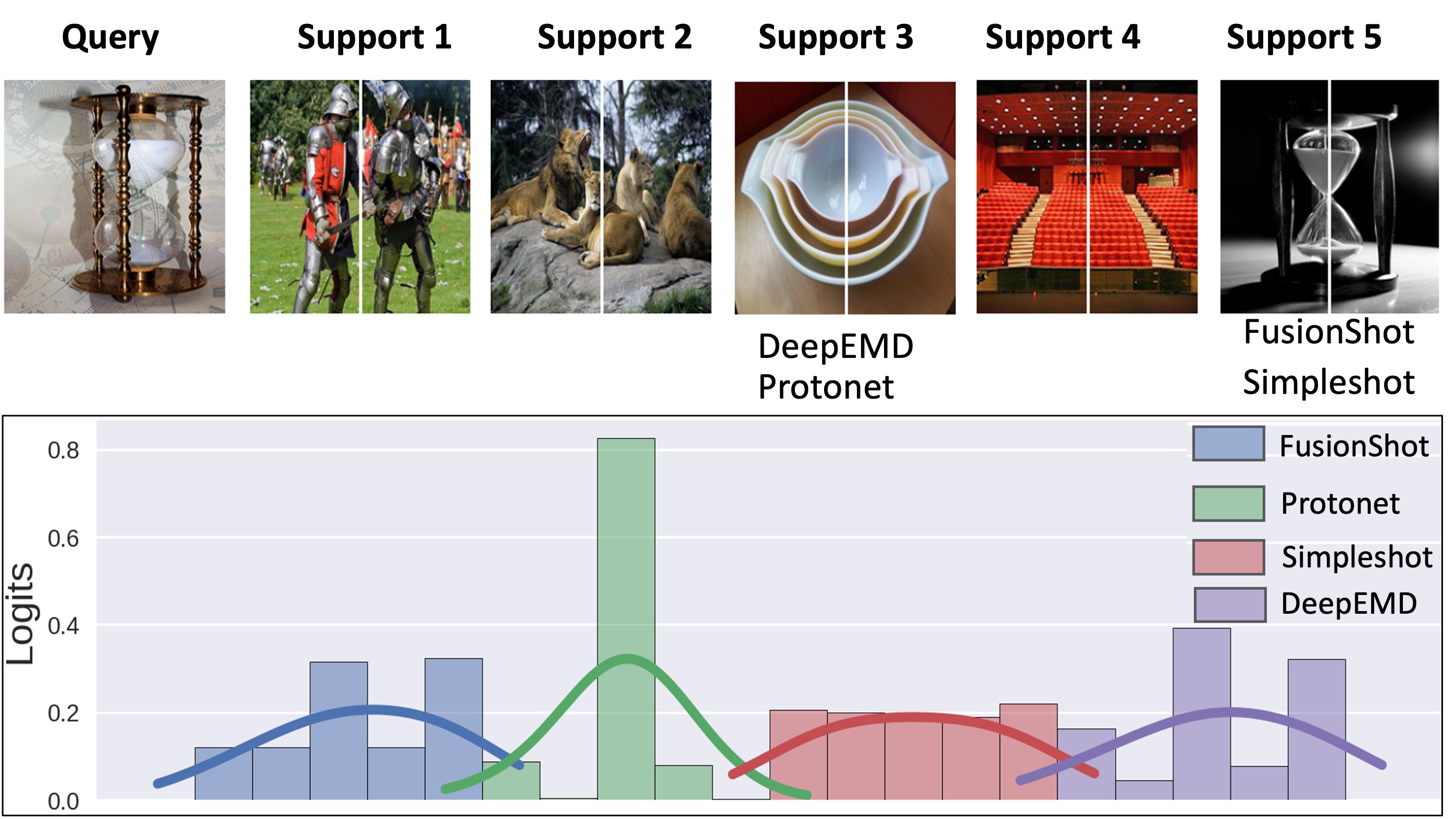}
        \label{fig:obs_density_b}
    \end{subfigure}
        \begin{subfigure}{0.3\textwidth}
        \centering
        \includegraphics[width=\textwidth]{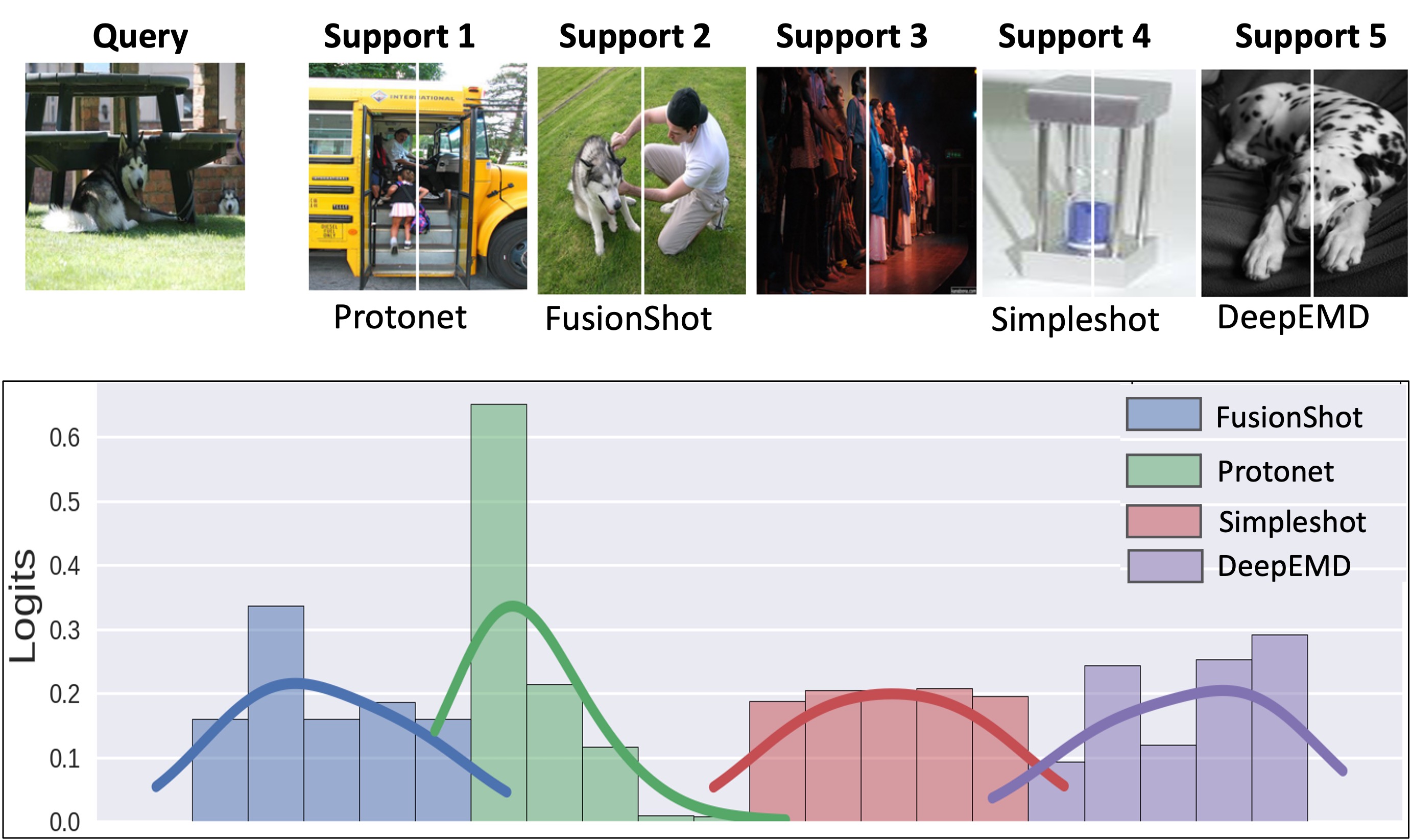}
        \label{fig:obs_density_c}
    \end{subfigure}
    \vspace{-5pt}
    \caption{For given queries, we show the probability densities assigned by each model to the support images. We want to emphasize the importance of the secondary probabilities, i.e., the probabilities that do not correspond to the ground truth.}
    \label{fig:obs_density}
\end{figure*}

Despite the fact that the accumulative behavior of base models shows similar patterns in terms of top-1 and top-5 performance ups and downs, their individual episode decisions tend to provide different probability densities across the novel classes in the corresponding support sets as shown in Figure \ref{fig:obs_density}. The ensemble fusion method can effectively refine the top-1 and top-5 predictions by not only taking into account of the highest scores of the models but also resolving the top-1 and top-5 prediction inconsistencies among the component base models of the ensemble. 

\subsection{Details on Adversarial Querying and Defending with FusionShot}

\begin{table}[t]
  \centering
  \small
  \begin{adjustbox}{width=0.45\textwidth, center}
  \begin{tabular}{p{1.2cm}p{1.1cm}ccccc}
    \hline
    \multirow{2}{1cm}{Method} & \multirow{2}{1cm}{Backbone} & \multicolumn{2}{c}{Acc. Under Attack} & \multicolumn{2}{c}{Benign Acc}\\
    \cline{3-6}
    & &  1-shot & 5-shot & 1-shot & 5-shot \\
    \hline
    Matching & Resnet18 & $0.20_{0.04}$  & $0.24_{0.4}$  & $50.89_{0.78}$  & $64.13_{1.03}$  \\
    Prototypical & Resnet18 & $0.60_{0.03}$ & $0.09_{0.05}$ & $51.72_{0.81}$ & $68.28_{0.95}$ \\
    MAML & Resnet18 & $0.12_{0.03}$ & $0.06_{0.06}$ & $47.57_0.84$ & $60.55_{1.05}$ \\
    Relation & Resnet18 & $0.65_{0.08}$  & $0.20_{0.06}$  & $47.07_{0.77}$  & $61.87_{1.00}$  \\
    SimpleShot & Resnet18 & $0.00_{0.01}$  & $0.00_{0.00}$  & $62.61_{0.80}$  & $77.56_{0.89}$  \\
    DeepEMD & Resnet18 & $0.00_{0.00}$  & $0.00_{0.01}$  & $64.2_{0.75}$  & $80.51_{0.54}$  \\
    \hline
  \end{tabular}
  \end{adjustbox}
  \caption{\textit{Mini}-Imagenet performance of the base models under 20-step PGD white-box attack compared to the benign case.}
  \label{table:adv_attack_base}
\end{table}

\begin{table}[hbt!]
  \centering
  \small
  \begin{subtable}{\linewidth}
  \begin{adjustbox}{width=1\textwidth, center}
  \begin{tabular}{p{2cm}p{1.1cm}ccccc}
    \hline
    \multirow{2}{1cm}{Method} & \multirow{2}{1cm}{Backbone} & \multicolumn{2}{c}{Acc. under Attack} & \multicolumn{2}{c}{Benign Acc.}\\
    \cline{3-6}
    & &  1-shot & 5-shot & 1-shot & 5-shot \\
    \hline
    Prototypical & Resnet18 & $28.16_{0.84}$   & $40.80_{0.94}$  & $51.72_{0.81}$ & $73.88_{0.65}$ \\
    Matching & Resnet18 & $31.59_{0.96}$   & $42.02_{0.88}$  & $50.89_{0.78}$  & $68.13_{1.03}$  \\
    Relation & Resnet18 & $33.16_{0.9}$    & $45.31_{0.81}$  & $47.07_{0.77}$  & $69.03_{0.70}$  \\
    MAML & Resnet18 & $30.75_{0.80}$   & $42.51_{0.91}$  & $47.57_0.84$ & $66.20_{0.78}$ \\
    SimpleShot & Resnet18 &$28.05_{0.83}$   & $34.74_{0.83}$  &  $62.61_{0.80}$  & $78.33_{0.59}$  \\
    \hline
    $\mathbf{FusionShot}^{\mathrm{dist}}$ & Resnet18 & $\mathbf{43.31_{0.65}}$ & $\mathbf{59.74_{0.64}}$ & $\mathbf{66.38_{0.10}}$ & $\mathbf{81.58_{0.36}}$ \\
    \hline
  \end{tabular}
  \end{adjustbox}
  \caption{}
  \label{table:adv_1}
  \end{subtable}
    \begin{subtable}{\linewidth}
  \begin{adjustbox}{width=1\textwidth, center}
  \begin{tabular}{p{2cm}p{1.1cm}ccccc}
    \hline
    \multirow{2}{1cm}{Method} & \multirow{2}{1cm}{Backbone} & \multicolumn{2}{c}{Acc. Under Attack} & \multicolumn{2}{c}{Benign Acc.}\\
    \cline{3-6}
    & &  1-shot & 5-shot & 1-shot & 5-shot \\
    \hline
    Matching & Resnet18 & $32.38_{0.88}$ & $36.25_{0.77}$ & $73.49_{0.89}$ & $83.64_{0.61}$\\
    Prototypical & Resnet18 & $33.87_{0.96}$ & $36.75_{0.78}$ & $72.07_{0.93}$ & $85.01_{0.60}$\\
    MAML & Resnet18 & $36.38_{1.08}$ & $40.52_{0.87}$ & ${68.42}_{1.07}$ & ${82.70}_{0.65}$\\
    Relation & Resnet18 & $35.38_{1.01}$ & $39.88_{0.94}$ & $68.58_{0.94}$ & $82.75_{0.58}$ \\
    SimpleShot & Resnet18 & $25.58_{0.74}$ & $26.68_{0.67}$ & $63.20_{1.26}$ & $82.63_{0.87}$\\
    \hline
    $\mathbf{FusionShot}^{\mathrm{dist}}$ & Resnet18 & $\mathbf{47.44_{0.64}}$ & $\mathbf{56.21_{0.62}}$ & $\mathbf{78.64_{0.38}}$ & $\mathbf{88.95_{0.29}}$ &\\
    \hline
  \end{tabular}
  \end{adjustbox}
  \caption{}
  \label{table:adv_2}
  \end{subtable}
  \caption{{\small We show the transfer 20-step PGD attack on the base models for (a) mini-Imagenet (b) CUB datasets. The attacks are generated by attacking DeepEMD with the ResNet18 backbone. In (a) FusionShot combines the predictions of RelationNet, SimpleShot, and DeepEMD. In (b) FusionShot combines the predictions of MAML, SimpleShot, Protonet, and DeepEMD.}}
  \label{table:adv_attack_black}
\end{table}

In this section, we give further details on adversarial attacks and our defense mechanism. In order to test the robustness of our method, we employ a projected gradient descent attack (PGD) which is one of the most effective algorithms \cite{madry2017towards}. The objective of the attack algorithm is to produce a perturbation $\delta$ such that when it is added to the input, $\mathbf{x}$, the deep learning model produces a wrong target. Thus, the goal is to find the perturbation which maximizes the loss function:
\begin{equation}
    \delta_{adv} = \max_{\norm{\delta}_{p} < \epsilon}\mathcal{L}(\mathbf{x} + \delta, y),
    \label{eq:adv_1}
\end{equation}
where $\mathcal{L}$ is the loss function and $y$ is the correct target. In the case of an image classifier where $\mathbf{x}$ is an image, the created perturbation, $\delta_{adv}$, is bounded by the number of bits that are used to represent the image. Here, we use $\ell_{\infty}$, i.e., $p=\infty$ as the norm degree and $\epsilon = \nicefrac{8}{255}$ as the precision for 64-bit images. The equation \ref{eq:adv_1} represents an untargeted attack. For the targetted attack equation reverses to minimization where the goal is to find the $\delta_{adv}$ that leads to the selected target which is different than the label:
\begin{equation}
    \delta_{adv} = \min_{\norm{\delta}_{p} < \epsilon}\mathcal{L}(\mathbf{x} + \delta, \Tilde{y}),
    \label{eq:adv_2}
\end{equation}
where $\Tilde{y}\neq y$ and $\mathcal{L}$ is the same loss function. PGD attack minimizes this loss by calculating the gradient and taking a small step in that direction, where it repeats the procedure for multiple iterations. In the case of a few-shot classifier, we rewrite the equation by targeting the query image:
\begin{equation}
    \delta_{adv} = \min_{\norm{\delta}_{p} < \epsilon}\mathcal{L}_{cross}[F_{\phi}(\mathbf{Q} + \delta, \mathcal{S}), \Tilde{y}],
    \label{eq:adv_3}
\end{equation}
where $F_{\phi}$ is the few-shot learner with the $\phi$ parameters, $\mathbf{Q}, \mathcal{S}$ are the query image and support set as before, and $\Tilde{y}$ represents the class to which the query image does not pertain. The idea to attack query image rather than support is explained in \cite{goldblum2020adversarially}, and we recommend the reader to check further details. We employ the equation \ref{eq:adv_3} to attack our few-shot learners. We start with the random perturbation and iterate 20 steps, each with a step size of $\nicefrac{2}{255}$. We show the performance of each base model after the PGD attack in Table \ref{table:adv_attack_base}. In other words, Table \ref{table:adv_attack_base} also shows the white-box attack performance, where the attacker knows the type and the architecture of the model.

Our focus, however, is on black-box attacks where the attacker does not know the model inside our system and uses transferability of attacks \cite{kurakin2018adversarial}. In our experiments, we select DeepEMD as the victim model to generate the perturbations, and we add the perturbations to the queries of other base models. The performances of the base models under the transferability scenario are shown in Table \ref{table:adv_attack_black} for mini-Imagenet and CUB datasets. Although the performance of models in a transfer attack is higher than the direct attack, the accuracy values are still extremely low compared to benign cases. Thus, we coupled the victim model, DeepEMD, with the focal diversity selected ensemble set. We train the FusionShot on the predictions of the ensemble set base models for benign and attack cases by utilizing the $\mathcal{D}^{\mathrm{train}}$ and $\mathcal{D}^{\mathrm{val}}$. In the final step, we test FusionShot on novel queries that are attacked by the perturbations created using the victim model, which is DeepEMD. The performance on both datasets is shown in Table \ref{table:adv_attack_black}.

\subsection{Convergence Analysis of Fusionshot}

\begin{figure}[t]
    \centering
    \includegraphics[width=0.3\textwidth]{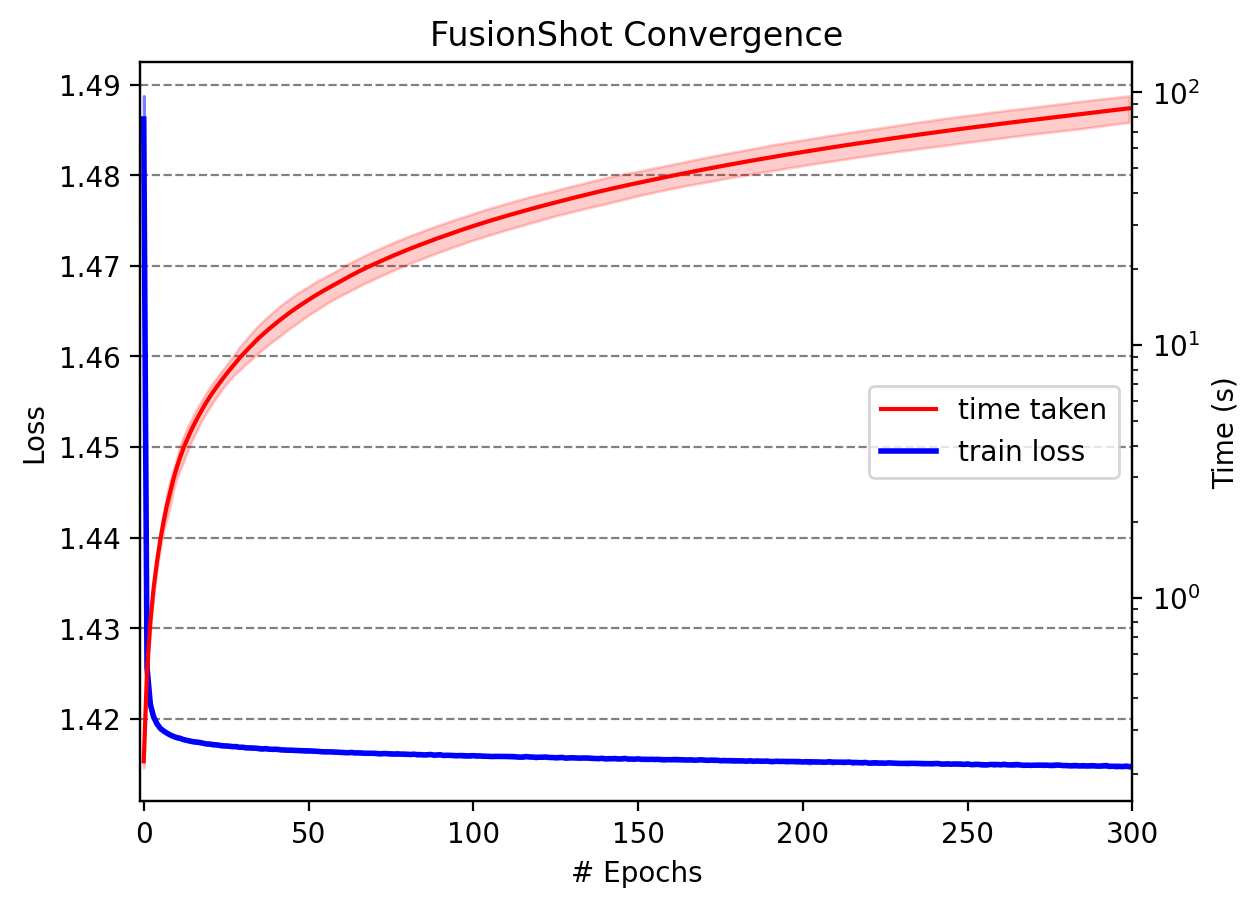}
    \caption{We show the convergence of the FusionShot model which learns to combine 3 few-shot learners; DeepEMD, SimpleShot, and Prototypical with ResNet18 architecture.}
    \label{fig:convergence}
\end{figure}

In Figure \ref{fig:convergence}, we illustrate the convergence behavior of FusionShot. Our training process involves validation at each epoch, and we choose the model parameters that yield the highest validation accuracy. Although the peak validation accuracy is typically achieved around the $150^\mathrm{th}$ epoch in our experiments, we continue training until the $300^\mathrm{th}$ epoch. As depicted in Figure \ref{fig:convergence}, our model completes training in 100 seconds. Notably, the size of our model is approximately $0.047$MB since it contains only 3 linear layers with 100 hidden dimensions. We perform all our experiments on a PC with the following specifications: 12th Gen Intel(R) Core(TM) i5-12400, GeForce RTX 3060 GA106 12GB, and 32GB of DDR4 RAM.

\subsection{Model Pool Details}
In Figure \ref{fig:heat_map}, we illistrate the important role of focal diversity based ensemble pruning using a pool of 10 pre-trained few shot models trained on \textit{mini}-ImageNet. These 10 models are MatchingNet-ResNet18, ProtoNet-ResNet18, Relationnet-ResNet18, MAMLResNet18, ProtoNet-Conv6, MatchingNet-Conv6, MAML-Conv6, SimpleShotWideRes, RelationNet-Conv6, DeepEMD-ResNet18.

\section{Additional Experiments}

\subsection{Interpretation of Ensemble Fusion}
\begin{figure*}[hbt!]
\centering
    \begin{subfigure}{0.2\textwidth}
        \centering
        \includegraphics[width=\textwidth]{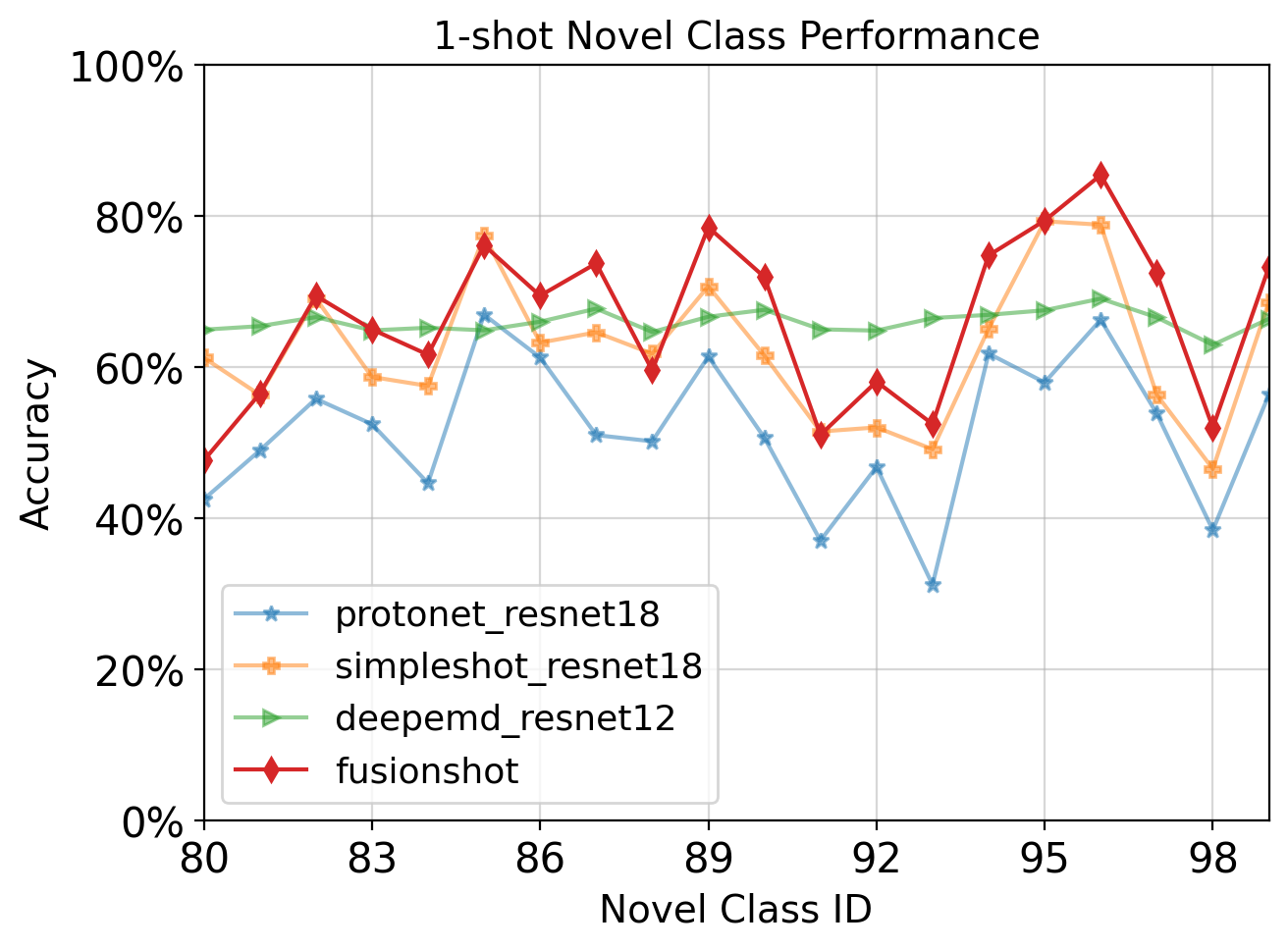}
        \caption{}
        \label{fig:cls_plot_1shot}
    \end{subfigure}
    \begin{subfigure}{0.25\textwidth}
        \centering
        \includegraphics[width=\textwidth]{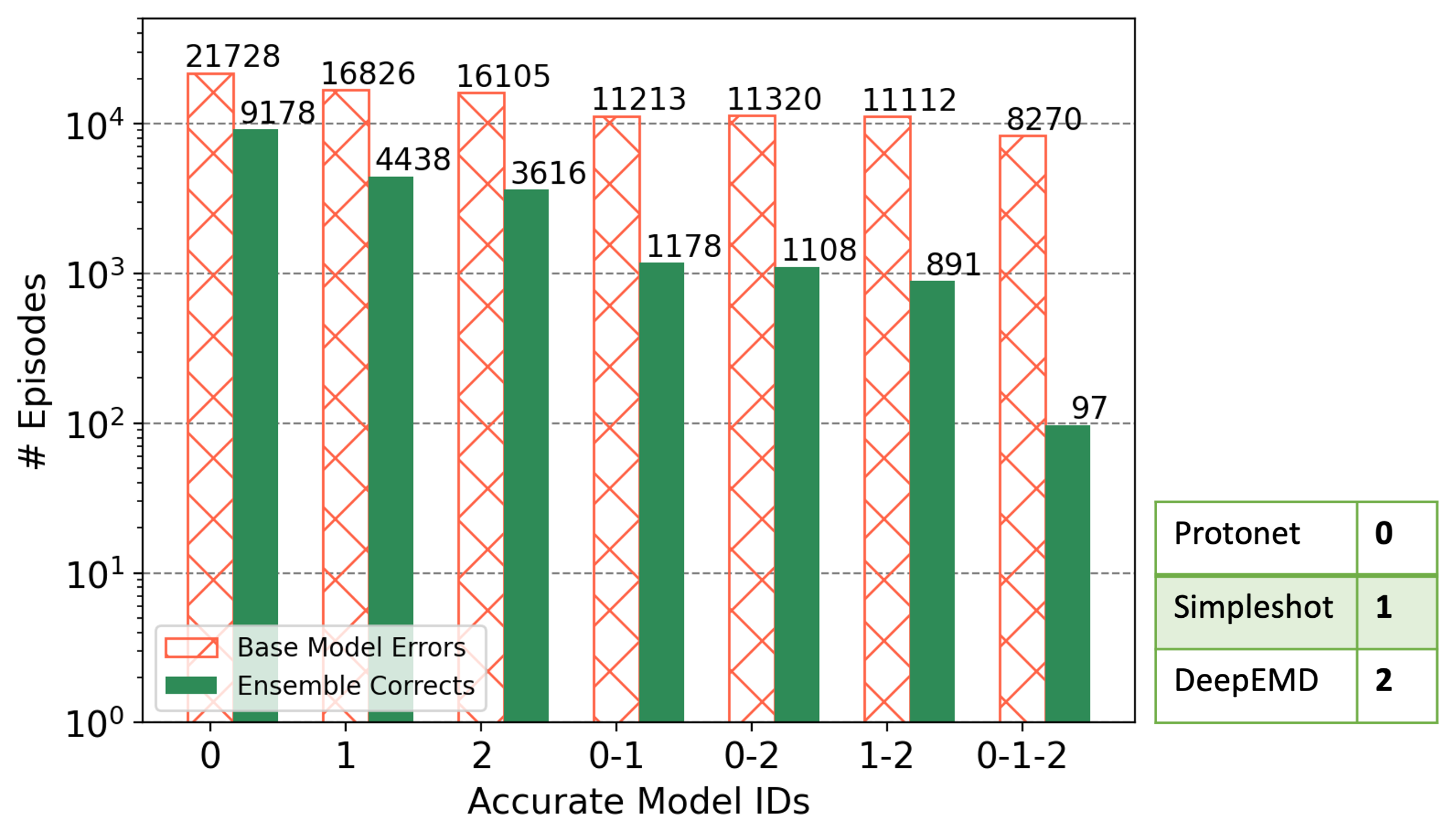}
        \caption{}
        \label{fig:improvement_benign}
    \end{subfigure}
    \begin{subfigure}{0.2\textwidth}
        \centering
        \includegraphics[width=\textwidth]{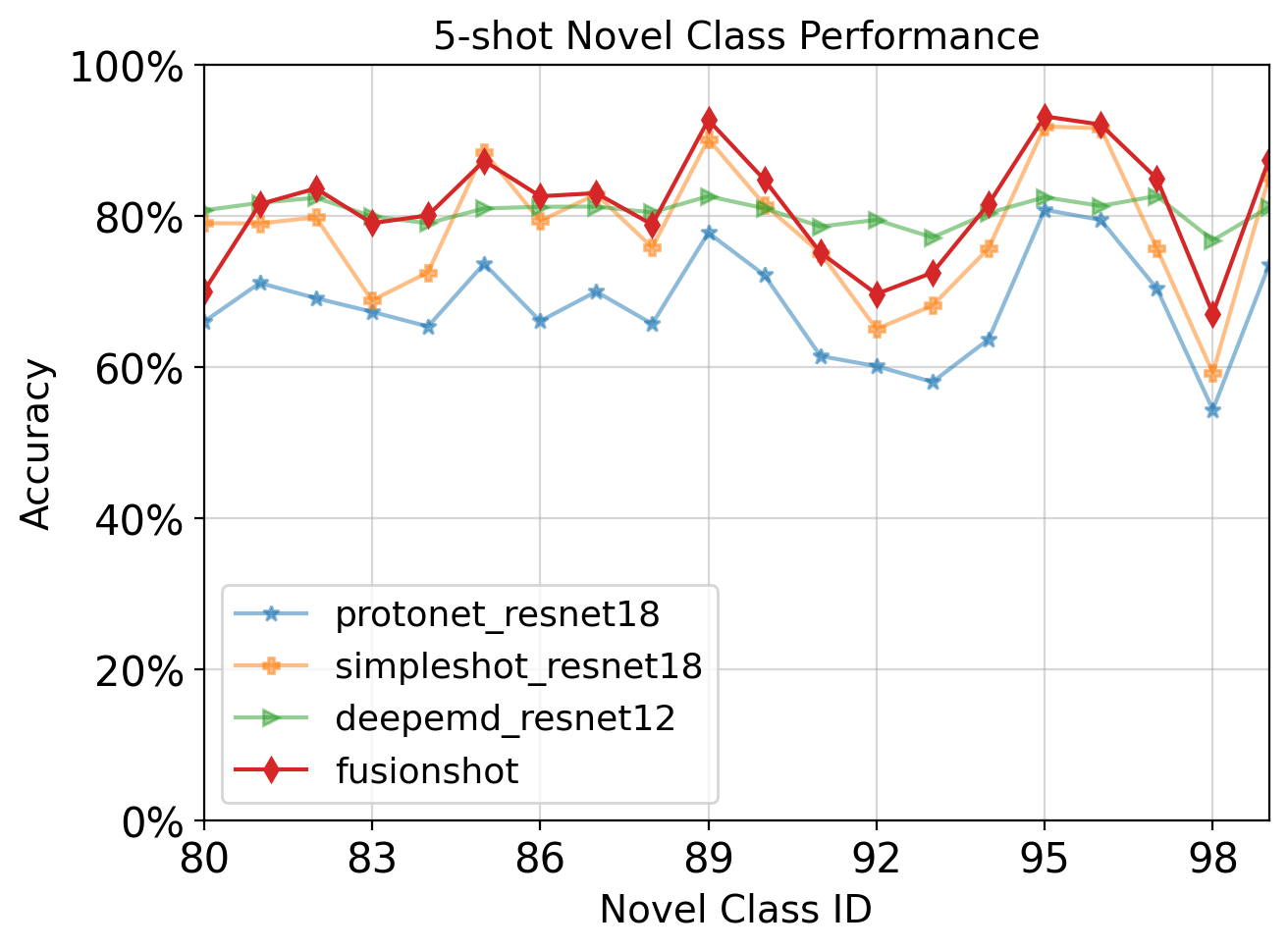}
        \caption{}
        \label{fig:apx_improvement_a}
    \end{subfigure}
    \begin{subfigure}{0.25\textwidth}
        \centering
        \includegraphics[width=\textwidth]{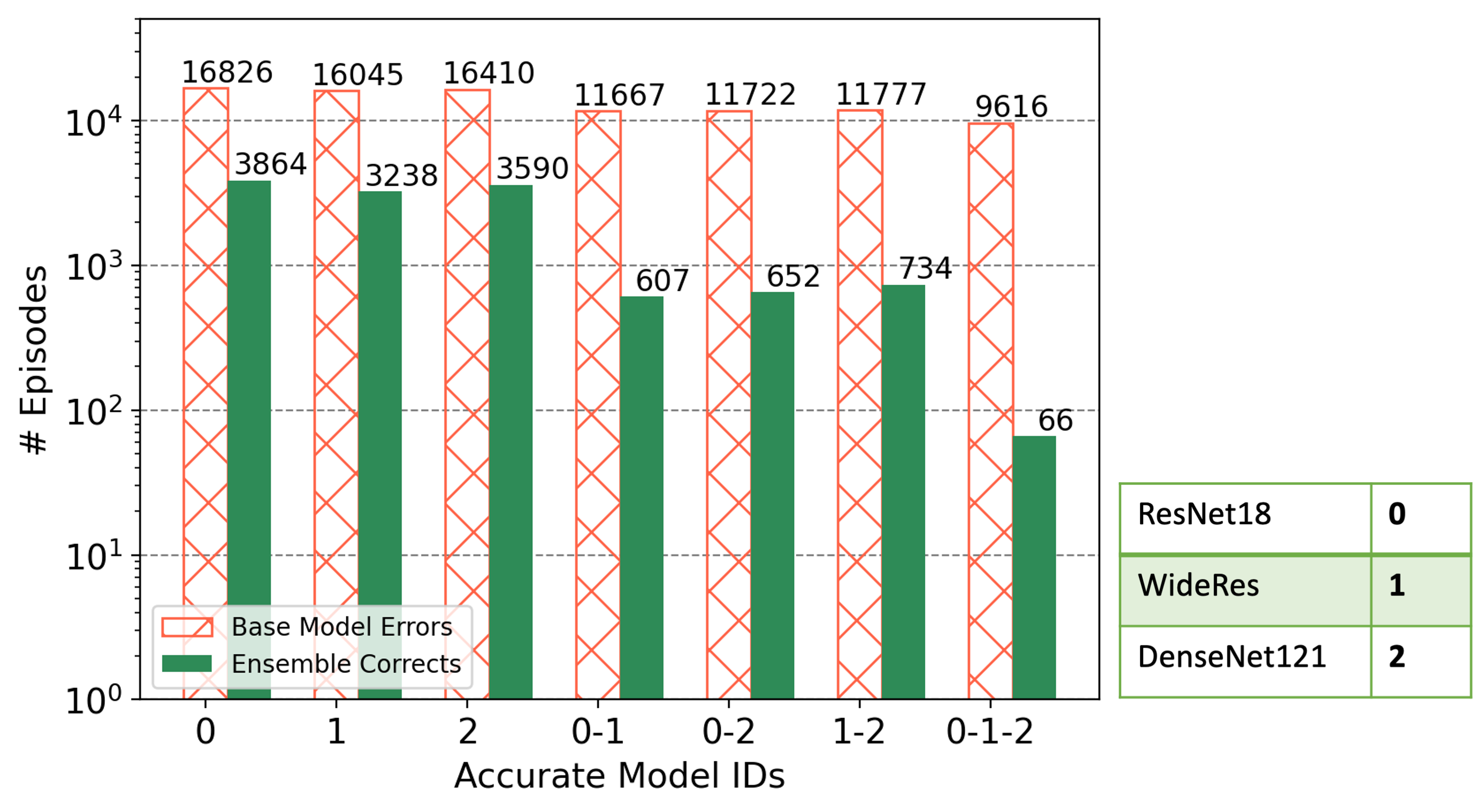}
        \caption{}
        \label{fig:apx_improvement_b}
    \end{subfigure}
    \caption{{\small (a) (c) We show the accuracy of base models and the FusionShot on each novel class out of 45000 episodes for the 1-shot/5-shot 5-way performance. (b) (d) Red bar: \# errors made by single base model or all models in a team out of 45000 novel episodes (1-shot 5-way, \textit{mini}-Imagenet). Green bars: \# corrected episodes by FusionShot. In (b) we keep the backbone architecture type the same, ResNet18, change the latent distance function, and in (d) we perform the vice-versa.}}
    \label{fig:apx_improvement}
\end{figure*}

This section provides experiments for the interpretation of ensemble fusion. Figure \ref{fig:cls_plot_1shot} compares the performance of each base model with the {\sc FusionShot} for each of the 20 novel classes of mini-Imagenet. We make three observations. (i) The performance of {\sc FusionShot} ensemble fusion can serve as the upper bound for each base model. (ii) Protonet and Simpleshot show similar high-low performance patterns in the set of novel classes. Although Simpleshot uses the k-nearest neighbor distance comparison method, both use L2 distance as their vector similarity metric. (iii) DeepEMD displays nearly uniform performance across all novel classes, likely due to the use of earth movers' distance (EMD).

Figure \ref{fig:improvement_benign}, on the other hand, shows all the base model predictions in 45000 episodes and counts the errors that each model made individually or together. The red bars show the number of errors that each individual makes or the number of errors made by all models in the 2-model or 3-model combinations, while the green bar shows the number of episodes that are corrected by our focal diversity selected ensemble for top-1 prediction performance. We make three observations. First, the right-most red bar shows that the ensemble of 3-distance methods (Protonet, Simpleshot, and DeepEMD) with ResNet18 backbone architecture failed together in 8270 episodes, and our ensemble fusion model can successfully correct 97 of them (1.2\%). 
Second, consider the 2-model ensemble \{1-2\}, there are 11112 episodes in which both Simpleshot and DeepEMD made incorrect top-1 predictions, and our ensemble fusion method can correct 891 of them (8\%). Finally, for Protonet (left-most), our {\sc FusionShot} ensemble can correct 9178 out of 21728 incorrect episodes, offering over 42\% performance improvement. 

Similarly, Figure \ref{fig:apx_improvement_a} compares the 5-shot novel class performance of DeepEMD, SimpleShot, Protonet, and {\sc FusionShot}. We observe that {\sc FusionShot} improves the top-5 performance of the best base model even though the worst model fails miserably. Furthermore, {\sc FusionShot} can reduce the failure rate in scenarios where the majority of the base models in an ensemble fail.

Figure~\ref{fig:apx_improvement_b} shows that a similar analysis we showed in Figure \ref{fig:improvement_benign} is also applicable to the top-5 performance. Here we use SimpleShot as the latent distance function while changing the backbone architectures. The percentage of corrections made by the {\sc FusionShot} ensemble is 20\% when the minority of the base models in an ensemble make incorrect predictions and 10\% when the majority of the base models in an ensemble make incorrect predictions. Even when all of the base models in the ensemble make incorrect decisions, {\sc FusionShot} selected ensembles can still improve the top-5 performance by correcting 1\% of the errors.  

\subsection{Cross-Domain Evaluation}
\begin{table}[hbt!]
    \small
    \centering
    \begin{adjustbox}{width=0.45\textwidth, center}
    \begin{tabular}{l c}
        \hline
        Method & \textit{mini}-Image$\rightarrow$CUB\\
        \hline
        Matching & $52.17_{0.74}$ \\
        Prototypical & $55.24_{0.72}$ \\
        Relation & $50.93_{0.73}$ \\
        MAML & $46.85_{0.72}$ \\
        Simpleshot & $67.38_{0.70}$ \\
        \hline
        MAML-Matching-Proto &  ${59.82_{0.46}}$ \\
        MAML-Matching-Protonet-Relation &  ${61.17_{0.46}}$ \\
        MAML-Matching-Protonet-Relation-SimpleShot &  $\mathbf{69.91}_{0.44}$  \\
    \hline
    \end{tabular}
    \end{adjustbox}
    \caption{Performing Cross-Domain experiments without DeepEMD. All the methods use the ResNet18 backbone.}
    \label{table:closer_look_more}
\end{table}

In addition to the results reported in Table \ref{table:all_scores}, we perform {\sc FusionShot} in the blind setting for a pool of base models without DeepEMD, which outperforms other base models by about 10\%. Table \ref{table:closer_look_more} shows that {\sc FusionShot} selected ensembles can improve the Simpleshot performance by 2\% and improve other base model performances up to 6\% when Simpleshot is removed.

\subsection{FusionShot Vs Plurality Voting}
\begin{figure}[t]
    \centering
    \includegraphics[width=0.4\textwidth]{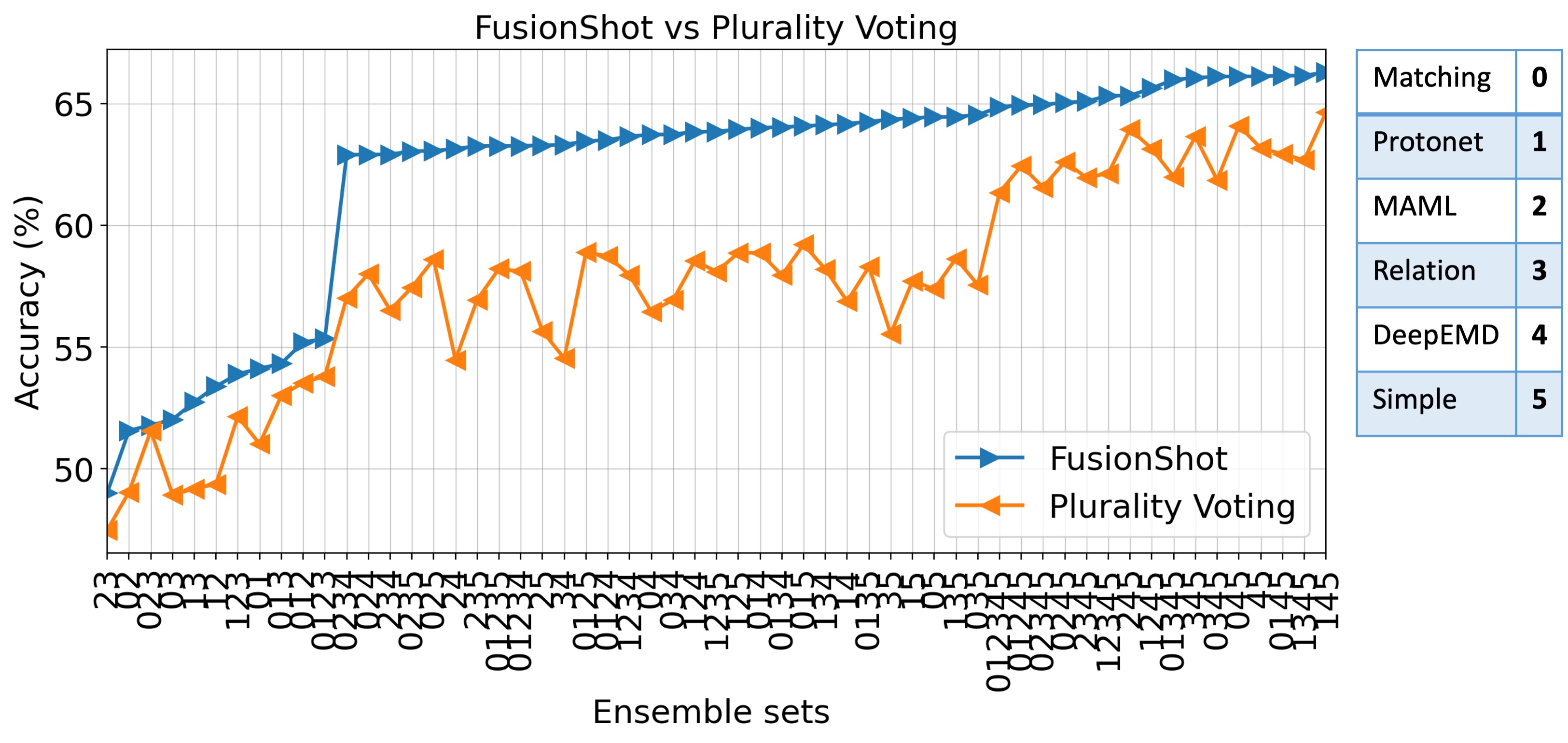}
    \caption{{\small Comparison of FusionShot with Plurality Voting applied to the ensemble sets with Matching, Proto, MAML, Relation, DeepEMD, and SimpleShot methods with ResNet18 architectures.}}
    \label{fig:fs_vs_p}
\end{figure}

This set of experiments evaluates the performance gain of using MLP (learn-to-combine) in the ensemble fusion stage. Figure \ref{fig:fs_vs_p} shows the sorted accuracy measurements for FusionShot learn to combine based ensemble fusion compared to  FusionShot plurality voting consensus-based ensemble fusion on the $\mathcal{Y}^{\mathrm{novel}}$ dataset of \textit{mini}-Imagenet benchmark. While the y-axis shows the accuracy, the x-axis shows the ensemble set base members. Note that we sort the ensemble sets according to their performance when FusionShot combines base model predictions.

\begin{figure*}[t]
\centering
    \begin{subfigure}{0.5\textwidth}
        \centering
        \includegraphics[width=\textwidth]{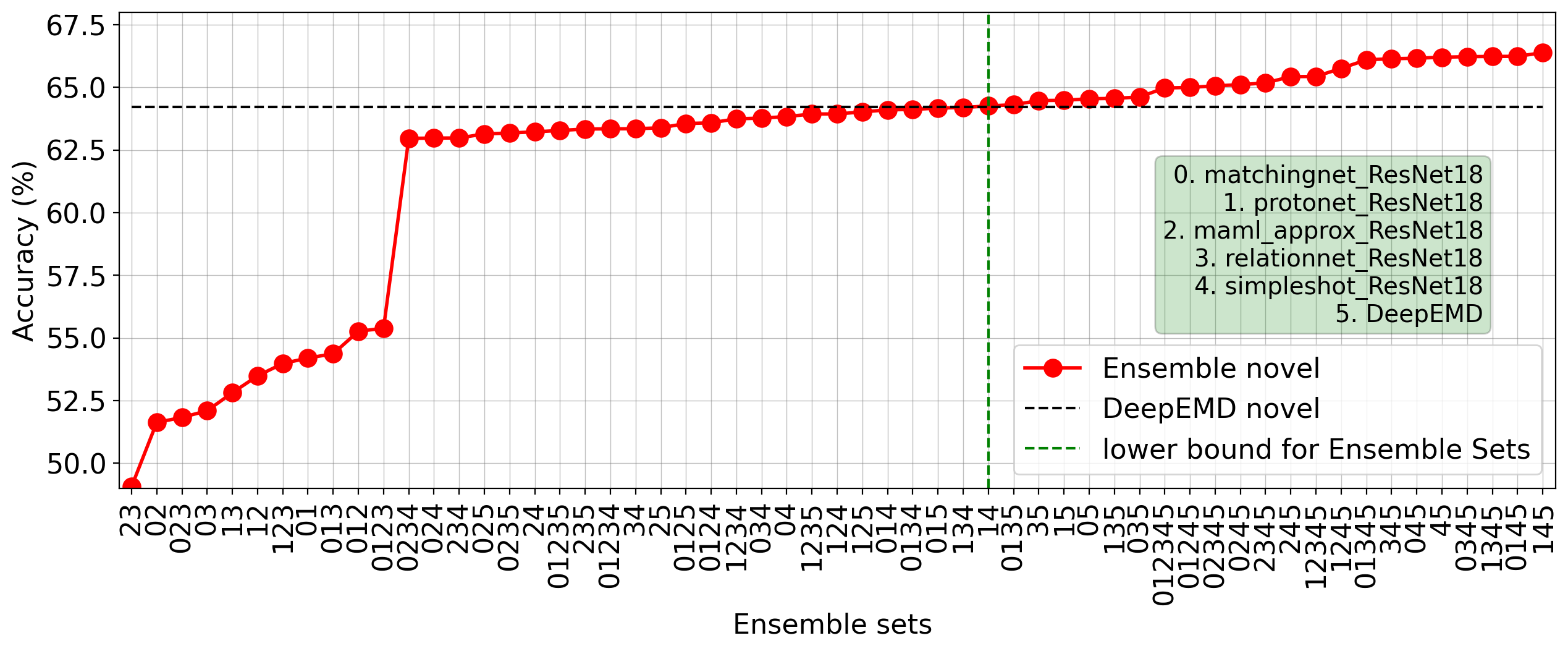}
        \caption{}
        \label{fig:line_a}
    \end{subfigure}
    \begin{subfigure}{0.8\textwidth}
        \centering
        \includegraphics[width=\textwidth]{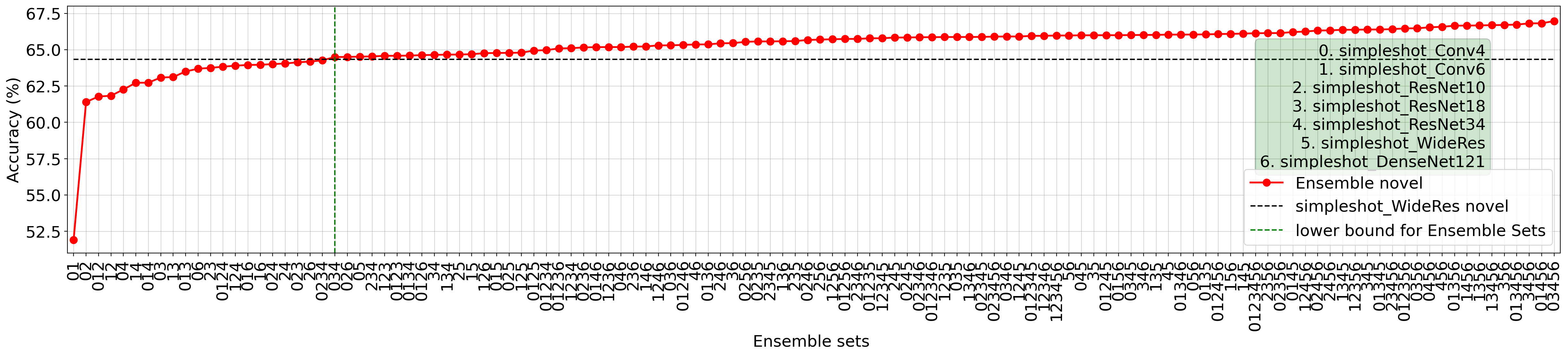}
        \caption{}
        \label{fig:line_b}
    \end{subfigure}
    \caption{The sorted performances of each FusionShot model that is trained on each possible sub-ensemble set predictions in 1-shot 5-way setting on \textit{mini}-Imagenet dataset. We provide all the scores in the Table \ref{table:method_acc} and \ref{table:bb_acc} for more details.}
    \label{fig:line}
\end{figure*}

\subsection{Backbone Feature Extractors and Latent Distance Method Analysis}
\vspace{-3pt}
Figure \ref{fig:line} provides some additional details for the results reported in Figure \ref{fig:scatter_method} and \ref{fig:scatter_backbone}. The hyperparameters, learning rate, number of epochs, and FusionShot architecture are kept the same. We use the same datasets, $\mathcal{Y}^{\mathrm{train}}, \mathcal{Y}^{\mathrm{val}}, \mathcal{Y}^{\mathrm{novel}}$, during training, validation, and testing. We use the validation data to decide when to stop the learning process and select the best-performing validation model. 

 Figure \ref{fig:line_a} shows the performances of several {\sc FusionShot} models, each is trained by using a different latent distance function under the same backbone DNN architectures.  Figure \ref{fig:line_b} shows the performances of several {\sc FusionShot} models, each is trained by using a different backbone DNN architecture but all use the same latent distance function for metric space comparison. We make two observations: 
 (i) Our ensemble fusion approach outperforms the best-performing component base model, e.g., DeepEMD (64.21\%).
 (ii) A fair number of ensembles can outperform DeepEMD (see those after the vertical line in both figures), indicating the opportunity for few-shot ensemble learning, and the opportunity for our focal diversity optimized ensemble pruning to effectively select top-performing ensemble teams. 

Table \ref{table:method_acc} and Table \ref{table:bb_acc} show all the ensemble teams included in Figure \ref{fig:line_a} and Figure \ref{fig:line_b} respectively using in a table, ranked by their novel accuracy in ascending order. We make the following four highlights for Table \ref{table:method_acc}. (1) The best-performing 3-model ensemble set is highlighted by $^1$ in yellow. It shows the effectiveness of our focal diversity-optimized ensemble pruning algorithm. (2) In contrast, removing the Protonet base model and simply putting the two best-performing base models (DeepEMD and SimpleShot) into a 2-model ensemble does not yield the top-performing ensemble, see highlight by $^2$ in yellow. (3) The four-model ensemble highlighted by $^3$ in yellow cannot outperform the 2-model ensembles even when the two base models are selected from the component models of this 4-model ensemble. However, all of them are outperforming DeepEMD which has 64.21\% novel accuracy. This shows that our ensemble fusion approach can effectively compose the top-performing ensemble learners that outperform the SOTA method. (4) As highlighted by $^4$ in yellow, we show the best performing 4-model ensemble without including Simpleshot or DeepEMD as a member base model. This indicates the important role of a strong few-shot model in ensemble learning, as demonstrated by the top 16 performing ensemble teams identified by {\sc FusionShot}, although we have shown that {\sc FusionShot} can compose ensembles that outperform DeepEMD, as discussed in the observation (3).

Similar observations can be made for Table \ref{table:bb_acc}.
(1) As highlighted by $^1$ in yellow, the best-performing ensemble is the 4-model team, each trained using different backbone architectures. (2) As highlighted by $^2$ in yellow, the 3-model ensemble has 66.70\% novel accuracy lower than the top-performing 4-model ensemble (66.97\%), but both outperform the ensemble with all 7 base models (66.13\%).  (3) We can also find the best performing 2-model ensemble that the majority of the models, highlighted by $^3$ in yellow, followed by the second best performing 2-model ensemble, highlighted by $^4$ in yellow, showing that the ensemble of the larger team size may not outperform the ensemble of smaller size. 

\subsection{Traversing the Solution Space with Genetic Algorithm}
\begin{figure}[t]
  \centering
  \begin{subfigure}{0.25\textwidth} 
    \centering
    \includegraphics[width=\textwidth]{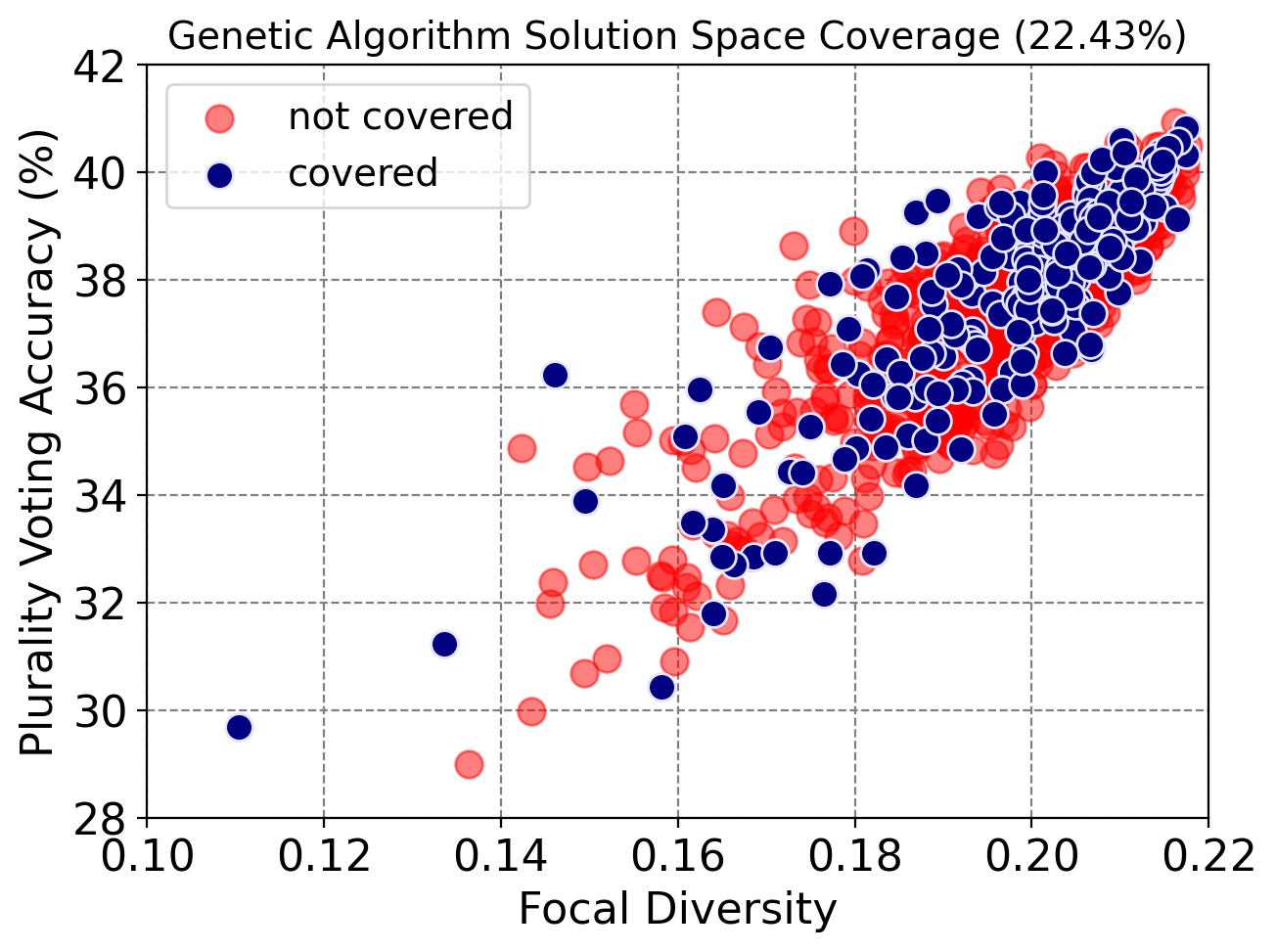}
    \caption{}
    \label{fig:acc_div}
  \end{subfigure}
  \begin{subfigure}{0.22\textwidth} 
    \centering
    \includegraphics[width=\textwidth]{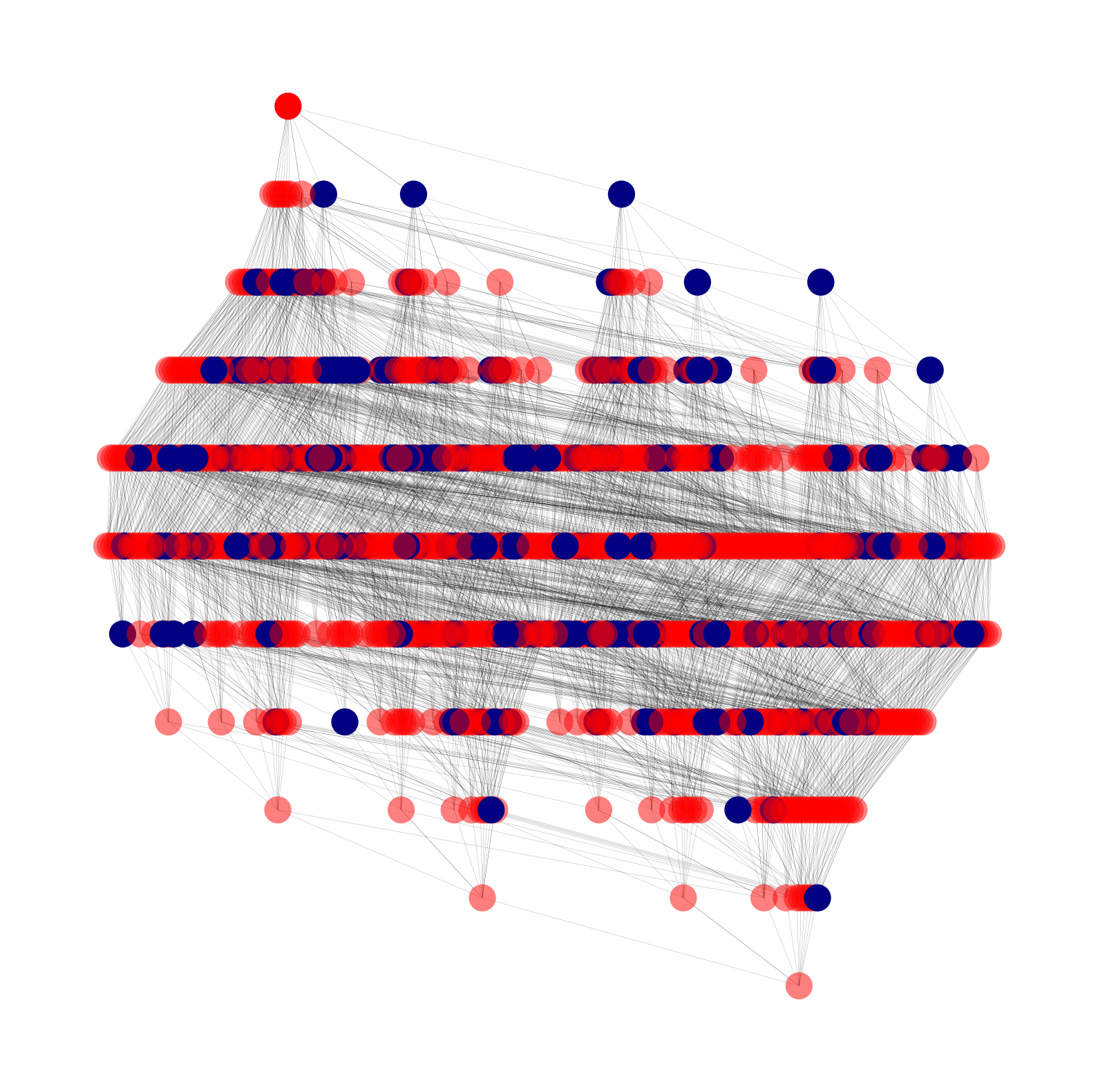}
    \caption{}
    \label{fig:conv}
  \end{subfigure}
  \caption{{\small We show the Genetic Algorithm solution space coverage for 10 base models. In (a) we show plurality voting accuracy against focal diversity metrics of each candidate ensemble set and whether they are covered by the GA. In (b) the tree representation of the solution space is shown where each layer represents the size of the ensemble set and each node represents a candidate ensemble set.}}
  \label{fig:ga_traverse}
\end{figure}

In this experiment, we show that GA traverses about 25\% of the solution space to reach the best solution. We record every distinct solution produced in each population and accept it as traversed. As shown in Figure \ref{fig:ga_traverse}, GA traverses towards the best solution due to the selection process, and as a result of cross-over it explores a new solution. In each iteration, the algorithm is getting close to the optimal solution.
We used \citep{gad2021pygad} library to implement the Genetic Algorithm.

\section{Cases Where FusionShot is Ineffective}

\begin{figure*}[hbt!]
\centering
    \begin{subfigure}{0.28\textwidth}
        \centering
        \includegraphics[width=\textwidth]{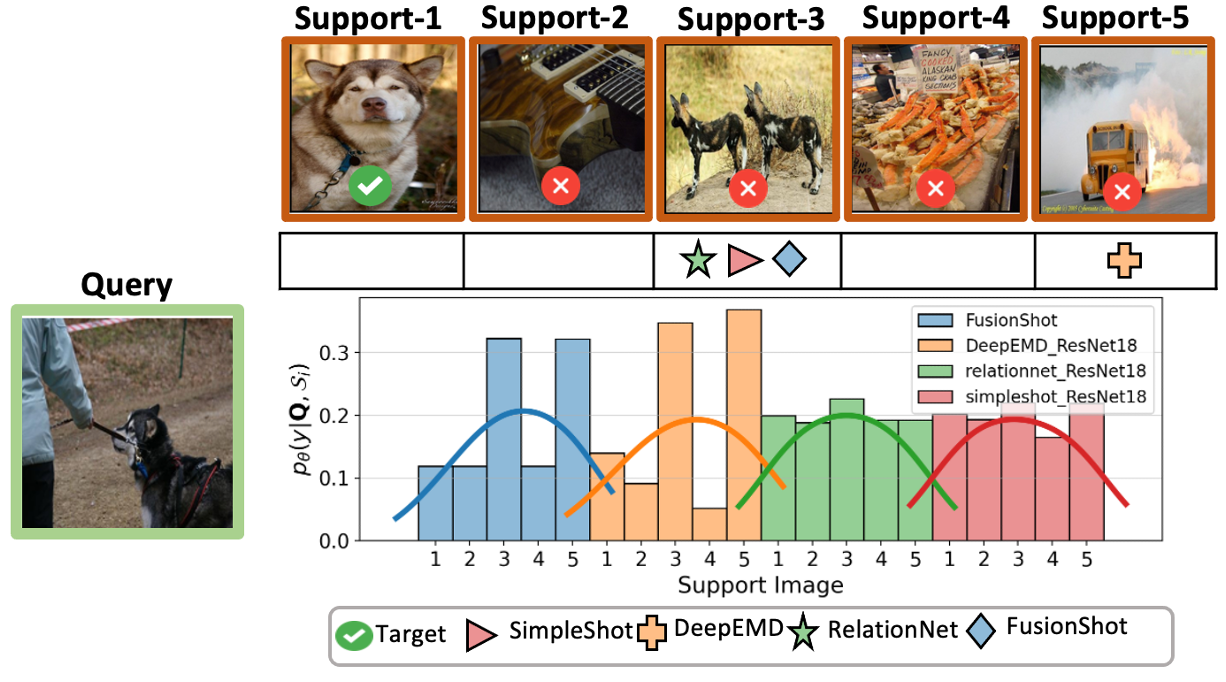}
        \caption{}
        \label{fig:fail1}
    \end{subfigure}
    \begin{subfigure}{0.28\textwidth}
        \centering
        \includegraphics[width=\textwidth]{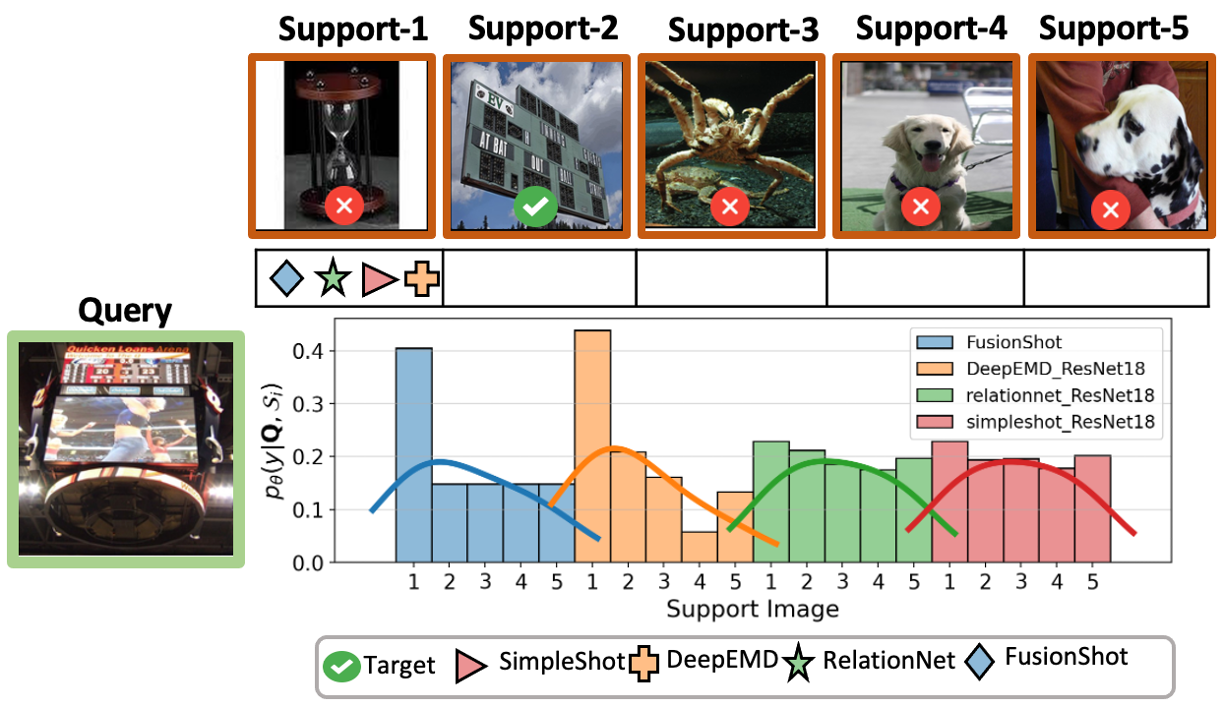}
        \caption{}
        \label{fig:fail2}
    \end{subfigure}
        \begin{subfigure}{0.32\textwidth}
        \centering
        \includegraphics[width=\textwidth]{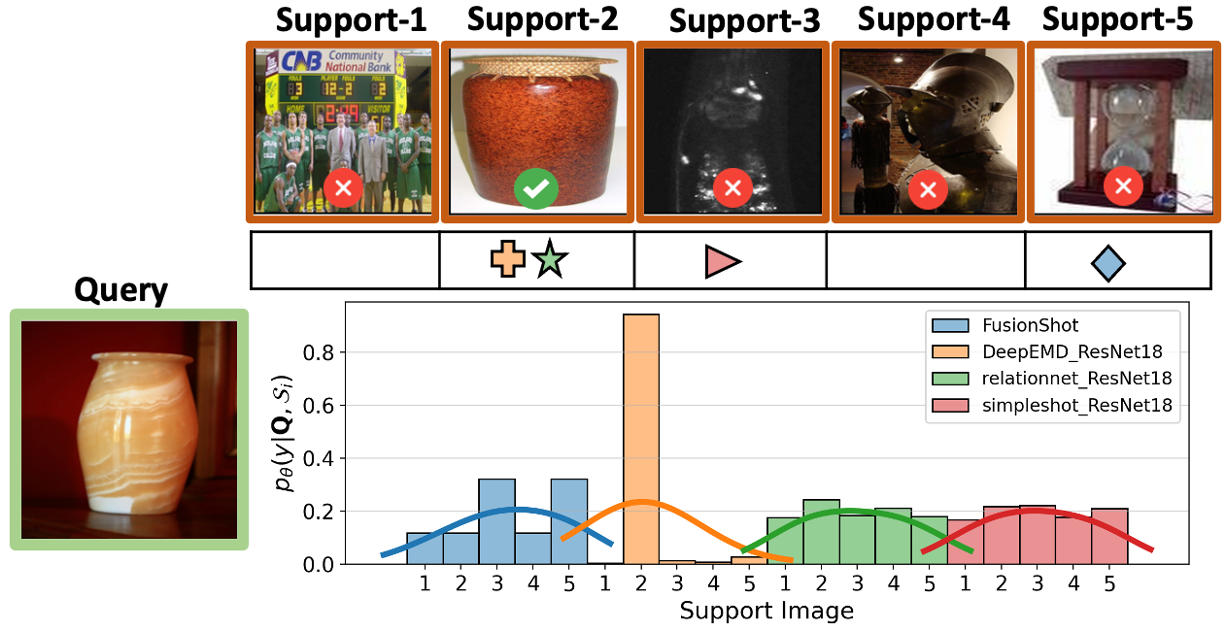}
        \caption{}
        \label{fig:fail3}
    \end{subfigure}
    \caption{Failing cases of FusionShot}
    \label{fig:fails}
\end{figure*}

In this set of experiments, we observe the cases where FusionShot makes incorrect decisions. As shown in Figure \ref{fig:fail1}, FusionShot is indecisive among two options where both are considered by the candidate models. It chooses the third support set which is selected by the two base models and the second option of the strong model, DeepEMD. Even though the selected choice is reasonable due to the choices of base models, it is wrong because the query image is a husky whereas the third support image shows jackals. One may argue that the query image lacks clarity, and the dog is not facing the camera which makes it hard to distinguish its breed. Thus, as the base models become hesitant, FusionShot also makes uncertain decisions.

In the second case shown in \ref{fig:fail2}, all the base models select the first image as the target, however, the correct target is the second image. When there is a big contrast difference between the query and the target image and there is another image in the support set that resembles the query in terms of shapes, the base models make a wrong decision. Hence, FusionShot also follows that decision. As we show in our previous experiments, such as Figure \ref{fig:improvement_benign}, FusionShot can only fix 1\% of errors when all the base models make a wrong decision.

Lastly, in Figure \ref{fig:fail3} we use a FusionShot when it is trained with adversarial and benign examples. The query image remains unattacked, and DeepEMD makes the correct decision with high certainty. However, SimpleShot makes an incorrect decision, which consequently causes FusionShot to perceive the query image as attacked and DeepEMD as the target.

\section{More Discussions on Few-shot Recent Literature}

Few-shot image classifiers can be categorized based on their learning strategies into two types: inductive and transductive. The transductive learning methods can provide high accuracy, \citep{liu2018learning}, but they require comparison with the fitted data for every new query which limits their applicability to other datasets, and slows the inference time performance. The inductive methods, however, aim for a task-agnostic function that can be applied to any dataset. In this work, we focus on the few-shot classifiers that follow inductive learning.

Numerous efforts have been dedicated to improving the performance of the few shot classifiers in the inductive setting by either designing a more complex latent distance function or improving the expressiveness of the feature extractor. The precedential methods, \citep{koch2015siamese, snell2017prototypical, vinyals2016matching} focused on the non-parametric latent distance functions such as Manhattan, Euclidian, and cosine distances. Following these works, relation networks \citep{sung2018learning}, and recent Graph-CNN variants \citep{garcia2017few, kim2019edge} offered the parameterized approaches by learning the distance function with neural networks. The SOTA works show the importance of the correlations between the embeddings by measuring the joint distribution with parameterized metrics, such as Earth Mover's Distance \citep{zhang2020deepemd}, Mahalanobis distance \citep{bateni2020improved}, and Brownian Distance \citep{xie2022joint}.

On the other side, various works that focus on the feature extractor proposed different techniques for training and testing. \citep{chen2019closer} performed a supervised pre-training on the feature extractor fraught with a fully connected layer that is fine-tuned during test time with few samples. Subsequently, \citep{wang2019simpleshot} simply implemented k-nearest-neighbor to the outputs of a pre-trained feature extractor and showed the performance with various feature extractor architectures. \citep{tian2020rethinking} defend that pre-trained embedding can outperform many classical methods. To further enhance the feature extractor's expressiveness, several approaches have been proposed, including the addition of extra self-supervised loss \citep{mangla2020charting}, the utilization of augmentation techniques \citep{luo2021rectifying, yang2021free}, and the generation of supplementary data \citep{li2020adversarial}. 
Research on feature extractor adaptation during test time~\citep{finn2017model, rusu2018meta} finds the best initial parameters for learning novel tasks in testing, and \citep{ye2020few, bateni2020improved} shows adaptive embedding to the target class during testing. 

In terms of task-agnostic and generalization capability of few-shot learners, the literature interest shifted towards improving the generalization of few-shot learners in unseen datasets \citep{triantafillou2019meta}, which mirrors the setting we experimented with in the blind cross-domain setting. The focus is on creating a universal representation to generalize to unseen domains \citep{bronskill2020tasknorm, liu2020universal, triantafillou2021learning}, and creating few-shot learners that are able to distinguish common features in between datasets \citep{dvornik2020selecting, li2021universal}. A recent approach, \citep{hiller2022rethinking} proposes the use of Vision Transformers to overcome the lack of fine-grained labels and learn high-level statistics. 

Many methods, including universal representation learners \citep{li2021universal}, have enhanced the generalization of their approaches by employing multiple backbones. \citep{dvornik2019diversity} used ensemble methods to mitigate the variance of few-shot learning classifiers, and jointly train their multiple backbones during supervised training with different penalizing terms for diversity and cooperation. Similarly, \citep{bendou2022easy} exploited the multiple backbones and improved the expressiveness of each backbone with self-supervised loss. In terms of leveraging other methods, \citep{ma2021partner} leveraged dual prototype networks, while one model is pre-trained to regularize the learning of the main encoder. Most recently, \citep{ma2022few} proposed a geometric ensemble approach by using Voronoi Diagrams to model class relationships on the latent space. 

Lastly, \cite{radford2021learning} proposed zero-shot learner training over 400 million (image, text) pairs. In this work, we focused on the few-shot ensemble learning, however, due to the flexibility of our algorithm, our method can be extended to ensembling any zero-shot learner by respecting their diversities with focal diversity. We leave this area to our future work.

\newpage

\begin{table}[hbt!]
    \small
    \centering
    \begin{adjustbox}{width=0.5\textwidth, center}
    \begin{tabular}{c c c c}
        \hline
    Ensemble Enumeration & Novel Accuracy & Ensemble Enumeration & Novel Accuracy  \\
    \hline
    C4-C6	& 51.91 &     RN10-WR-DN121 & 65.70 \\
    C4-RN10	& 61.40 &     C6-RN10-WR-DN121  & 65.72 \\
    C4-C6-RN10	& 61.78 &     C4-C6-RN10-WR-DN121  & 65.74 \\
    C6-RN10	& 61.83 &     RN10-RN18-RN34-DN121  & 65.74 \\
    C4-RN34	& 62.26 &     C4-C6-RN10-RN18-WR  & 65.78 \\
    C6-RN34	& 62.73 &     C6-RN10-RN18-RN34-WR  & 65.79 \\
    C4-C6-RN34	& 62.73 &     RN10-RN34-WR & 65.84 \\
    C4-RN18	& 63.09 &     C4-RN10-RN34-WR & 65.84 \\
    C6-RN18	& 63.11 &     C4-RN10-RN18-RN34-DN121 & 65.86 \\
    C4-C6-RN18	& 63.50 &     C4-C6-RN10-RN18-RN34-DN121 & 65.86 \\
    C4-DN121	& 63.69 &     C6-RN10-RN18-WR  & 65.87 \\
    RN10-RN18	& 63.75 &     C4-RN18-WR  & 65.87 \\
    C4-C6-RN10-RN34	& 63.83 &     C6-RN18-RN34-DN121  & 65.88 \\
    C6-RN10-RN34	& 63.90 &     C4-RN10-RN18-RN34-WR & 65.88 \\
    C4-C6-DN121	& 63.95 &     C4-RN10-RN18-RN34-WR-DN121  & 65.90 \\
    C6-DN121	& 63.96 &     C4-RN18-RN34-DN121 & 65.91 \\
    C4-RN10-RN34	& 64.01 &     C6-RN10-RN34-WR  & 65.91 \\
    RN10-RN34	& 64.05 &     C4-C6-RN10-RN18-RN34-WR & 65.95 \\
    C4-RN10-RN18	& 64.13 &     C6-RN10-RN18-RN34-DN121  & 65.96 \\
    RN10-DN121	& 64.18 &     C6-RN10-RN18-RN34-WR-DN121 & 65.98 \\
    C4-RN10-RN18-RN34	& 64.27 &     WR-DN121  & 65.98 \\
    C4-RN18-RN34	& 64.50 &     C4-RN34-WR & 65.99 \\
    C4-RN10-DN121	& 64.50 &     RN18-WR & 65.99 \\
    C4-WR	& 64.52 &     C4-C6-RN10-RN34-WR  & 66.00 \\
    RN10-RN18-RN34	& 64.54 &     C4-C6-WR-DN121 & 66.00 \\
    C6-RN10-RN18	& 64.58 &     C4-RN18-RN34-WR  & 66.01 \\
    C4-C6-RN10-RN18	& 64.58 &     RN18-RN34-DN121 & 66.02 \\
    C4-C6-RN18-RN34	& 64.60 &     C6-RN18-WR  & 66.02 \\
    C4-C6-RN10-DN121	& 64.63 &     \hl{RN34-WR}  & \hl{$66.04^3$} \\
    RN18-RN34	& 64.64 &     C4-C6-RN18-RN34-DN121 & 66.04 \\
    C6-RN18-RN34	& 64.66 &     C4-WR-DN121  & 66.04 \\
    RN10-WR	& 64.67 &     C4-C6-RN18-WR & 66.05 \\
    C6-WR	& 64.69 &     C4-C6-RN10-RN34-WR-DN121  & 66.08 \\
    C6-RN10-DN121	& 64.76 &     C6-WR-DN121 & 66.09 \\
    C4-C6-WR	& 64.78 &     C6-RN34-WR & 66.11 \\
    C4-RN10-WR	& 64.78 &     C4-C6-RN10-RN18-RN34-WR-DN121  & 66.13 \\
    C6-RN10-WR	& 64.80 &     RN10-RN18-WR-DN121 & 66.14 \\
    C4-C6-RN10-WR	& 64.94 &     C4-RN10-RN18-WR-DN121 & 66.14 \\
    C4-C6-RN10-RN18-RN34	& 64.97 &     C4-C6-RN34-WR  & 66.22 \\
    C4-C6-RN10-RN18-DN121	& 65.09 &     C6-RN10-RN34-WR-DN121 & 66.25 \\
    C6-RN10-RN18-RN34	& 65.10 &     C4-RN10-RN34-WR-DN121 & 66.32 \\
    C4-RN10-RN18-DN121	& 65.16 &     RN10-RN34-WR-DN121 & 66.33 \\
    C4-C6-RN34-DN121	& 65.17 &     C6-RN18-RN34-WR  & 66.37 \\
    C6-RN10-RN18-DN121	& 65.18 &     C6-RN10-RN18-WR-DN121  & 66.37 \\
    C4-RN34-DN121	& 65.18 &     RN18-RN34-WR  & 66.39 \\
    RN10-RN18-DN121	& 65.21 &     C4-C6-RN18-RN34-WR  & 66.39 \\
    C6-RN34-DN121	& 65.22 &     RN10-RN18-RN34-WR-DN121 & 66.42 \\
    C6-RN10-RN34-DN121	& 65.30 &     C4-C6-RN10-RN18-WR-DN121 & 66.47 \\
    C4-RN18-DN121	& 65.30 &     C4-RN18-WR-DN121  & 66.47 \\
    C4-C6-RN10-RN34-DN121	& 65.33 &     C4-RN34-WR-DN121  & 66.54 \\
    \hl{RN34-DN121}	& \hl{$65.36^4$} &     RN34-WR-DN121  & 66.58 \\
    C4-C6-RN18-DN121	& 65.37 &     C4-C6-RN18-WR-DN121  & 66.65 \\
    RN10-RN34-DN121	& 65.44 &     C6-RN34-WR-DN121  & 66.66 \\
    RN18-DN121	& 65.45 &     C6-RN18-WR-DN121 & 66.67 \\
    C4-RN10-WR-DN121	& 65.54 &     C6-RN18-RN34-WR-DN121  & 66.69 \\
    C4-RN10-RN18-WR	& 65.57 &     \hl{RN18-WR-DN121} & \hl{$66.70^2$} \\
    RN10-RN18-RN34-WR	& 65.57 &     C4-C6-RN18-RN34-WR-DN121  & 66.73 \\
    C6-RN18-DN121	& 65.58 &     RN18-RN34-WR-DN121  & 66.81 \\
    RN10-RN18-WR	& 65.59 &     C4-C6-RN34-WR-DN121  & 66.82 \\
    C4-RN10-RN34-DN121	& 65.67 &     \hl{C4-RN18-RN34-WR-DN121}  & \hl{$66.97^1$} \\
    \hline\\
    \end{tabular}
    \end{adjustbox}
    \caption{Comparison of FusionShot performance for each Few-shot Learning architecture combination for SimpleShot in 1-shot 5-way setting on \textit{mini}-Imagenet dataset. The highlighted data descriptions are in Appendix C.4}
    \label{table:bb_acc}
\end{table}

\begin{table}[hbt!]
    \centering
    \begin{adjustbox}{width=0.5\textwidth, center}
    \begin{tabular}{c c c c}
    \hline
    Ensemble Enumeration & Novel Accuracy & Ensemble Enumeration & Novel Accuracy \\
    \hline
    maml-rn & 49.06 & pn-maml-ss  & 64.02 \\
    mn-maml & 51.63 & mn-pn-EMD & 64.10 \\
    mn-maml-rn  & 51.82 & mn-pn-rn-EMD  & 64.12 \\
    mn-rn & 52.09 & mn-pn-ss  & 64.17 \\
    pn-rn & 52.81 & pn-rn-EMD & 64.20 \\
    pn-maml & 53.48 & pn-EMD  & 64.26 \\
    pn-maml-rn  & 53.98 & \hl{mn-pn-rn-ss} & \hl{$64.31^3$} \\
    mn-pn & 54.19 & rn-ss & 64.47 \\
    mn-pn-rn  & 54.37 & pn-ss & 64.48 \\
    mn-pn-maml  & 55.26 & mn-ss & 64.54 \\
    \hl{mn-pn-maml-rn} & \hl{$55.38^4$} & pn-rn-ss  & 64.56 \\
    mn-maml-rn-EMD  & 62.96 & mn-rn-ss  & 64.62 \\
    mn-maml-EMD & 62.97 & mn-pn-maml-rn-EMD-ss  & 64.97 \\
    maml-rn-EMD & 62.98 & mn-pn-maml-EMD-ss & 65.00 \\
    mn-maml-ss  & 63.14 & mn-maml-rn-EMD-ss & 65.06 \\
    mn-maml-rn-ss & 63.18 & mn-maml-EMD-ss  & 65.11 \\
    maml-EMD  & 63.22 & maml-rn-EMD-ss  & 65.18 \\
    mn-pn-maml-rn-ss  & 63.28 & maml-EMD-ss & 65.43 \\
    maml-rn-ss  & 63.34 & pn-maml-rn-EMD-ss & 65.43 \\
    mn-pn-maml-rn-EMD & 63.34 & pn-maml-EMD-ss  & 65.76 \\
    rn-EMD  & 63.35 & mn-pn-rn-EMD-ss & 66.10 \\
    maml-ss & 63.39 & rn-EMD-ss & 66.14 \\
    mn-pn-maml-ss & 63.56 & mn-EMD-ss & 66.16 \\
    mn-pn-maml-EMD  & 63.58 & \hl{EMD-ss}  & \hl{$66.20^2$} \\
    pn-maml-rn-EMD  & 63.75 & mn-rn-EMD-ss  & 66.22 \\
    mn-rn-EMD & 63.77 & pn-rn-EMD-ss  & 66.23 \\
    mn-EMD  & 63.83 & mn-pn-EMD-ss  & 66.24 \\
    pn-maml-rn-ss & 63.94 & \hl{pn-EMD-ss} & \hl{$66.38^1$} \\
    pn-maml-EMD & 63.94 & \\
    \hline
    \end{tabular}
    \end{adjustbox}
    \caption{{\small Comparison of FusionShot performance for each Few-shot Learning method combination for ResNet18 in 1-shot 5-way setting on \textit{mini}-Imagenet dataset. The highlighted data descriptions are in Appendix C.4}}
    \label{table:method_acc}
\end{table}

\end{document}